\documentclass{article}

\usepackage[main, final]{neurips_2026}
\usepackage[left]{lineno}

\usepackage{natbib}

\usepackage[utf8]{inputenc} %
\usepackage[T1]{fontenc}    %
\usepackage{hyperref}       %
\usepackage{url}            %
\usepackage{booktabs}       %
\usepackage{amsfonts}       %
\usepackage{nicefrac}       %
\usepackage{microtype}      %
\usepackage{xcolor}         %

\usepackage{float} %
\usepackage{amsmath}
\usepackage{amssymb}
\usepackage{CJKutf8}        %
\usepackage{wrapfig}
\usepackage{fontawesome5}
\usepackage{multirow}

\usepackage{todonotes}
\usepackage{tabularx}
\usepackage{caption}
\usepackage{subcaption}
\usepackage{etoc}       %
\usepackage{xcolor}
\definecolor{motoRedBg}{HTML}{D95C3E}      %
\definecolor{motoPurpleBg}{HTML}{7A52B8}   %
\definecolor{motoTealBg}{HTML}{2E96A8}     %
\definecolor{motoGreenBg}{HTML}{4EA868}    %
\definecolor{motoDarkPurpleBg}{HTML}{5C3AA8} %
\definecolor{motoYellowBg}{HTML}{D4A800}   %
\definecolor{motoGrayBg}{HTML}{888888}     %
\definecolor{motoPinkBg}{HTML}{C4527E}     %
\definecolor{motoRed}{HTML}{B03824}        %
\definecolor{motoPurple}{HTML}{6038A0}     %
\definecolor{motoTeal}{HTML}{1A6878}       %
\definecolor{motoGreen}{HTML}{2E7848}      %
\definecolor{motoDarkPurple}{HTML}{44289A} %
\definecolor{motoYellow}{HTML}{8C6400}     %
\definecolor{motoGray}{HTML}{505050}       %
\definecolor{motoPink}{HTML}{8C2858}       %
\definecolor{motoBlueBg}{HTML}{3E78C8}     %
\definecolor{motoOrangeBg}{HTML}{E08A1E}   %
\definecolor{motoBlue}{HTML}{1F4F96}       %
\definecolor{motoOrange}{HTML}{A85A0C}     %

\usepackage{xspace}
\usepackage{atkinson} 
\usepackage[T1]{fontenc}
\usepackage{atkinson}
\newcommand{\motofont}{lmtt}
\newcommand{\moto}{{\fontfamily{\motofont}\selectfont MoTo}\xspace}
\newcommand{\toksuite}{{\fontfamily{lmtt}\selectfont TokSuite}\xspace}

\usepackage{hyperref}
\usepackage{pgffor}

\hypersetup{
    citecolor=violet,
    linkcolor=motoPink,
    urlcolor=motoTeal,
    colorlinks=true,
}
\newcounter{question}

\usepackage{algpseudocode}
\usepackage{algorithm}

\makeatletter
\newcommand{\saveqcolor}[2]{%
  \global\@namedef{qcolor@#1}{#2}}
\newif\ifquestionleft

\let\origlabel\label
\renewcommand{\label}[1]{%
  \origlabel{#1}%
  \ifcsname qenvcolor\endcsname
    \protected@write\@auxout{}{\string\saveqcolor{#1}{\qenvcolor}}%
  \fi}

\newcommand{\qref}[1]{%
  \def\qref@count{0}%
  \foreach \lbl in {#1} {\xdef\qref@count{\the\numexpr\qref@count+1}}%
  Q.%
  \def\qref@i{0}%
  \foreach \lbl in {#1} {%
    \xdef\qref@i{\the\numexpr\qref@i+1}%
    \ifnum\qref@i=1\relax
    \else\ifnum\qref@i=\qref@count\relax
      \ifnum\qref@count>2 , \fi and %
    \else
      , %
    \fi\fi
    \begingroup\hypersetup{hidelinks}%
    \textcolor{\@nameuse{qcolor@\lbl}}{\hyperref[\lbl]{\ref*{\lbl}}}%
    \endgroup
  }%
}

\newcounter{observation}

\newenvironment{observation}[1][black]
  {\par\vspace{0.5ex}%
   \def\oenvcolor{#1}%
   \begingroup\centering
   \refstepcounter{observation}%
   \itshape\color{#1}%
   \textbf{O.\arabic{observation}}\quad\ignorespaces}
  {\par\endgroup\vspace{0.5ex}%
   \ignorespacesafterend}
\renewcommand{\label}[1]{%
  \origlabel{#1}%
  \ifcsname oenvcolor\endcsname
    \protected@write\@auxout{}{\string\saveqcolor{#1}{\oenvcolor}}%
  \fi
  \ifcsname qenvcolor\endcsname
    \protected@write\@auxout{}{\string\saveqcolor{#1}{\qenvcolor}}%
  \fi}

\newcommand{\oref}[1]{%
  \def\oref@count{0}%
  \foreach \lbl in {#1} {\xdef\oref@count{\the\numexpr\oref@count+1}}%
  Obs.%
  \def\oref@i{0}%
  \foreach \lbl in {#1} {%
    \xdef\oref@i{\the\numexpr\oref@i+1}%
    \ifnum\oref@i=1\relax
    \else\ifnum\oref@i=\oref@count\relax
      \ifnum\oref@count>2 , \fi and %
    \else
      , %
    \fi\fi
    \begingroup\hypersetup{hidelinks}%
    \textcolor{\@nameuse{qcolor@\lbl}}{\hyperref[\lbl]{\ref*{\lbl}}}%
    \endgroup
  }%
}
\makeatother

\usepackage[most]{tcolorbox}  %
\newcommand{\tokenbox}[2]{%
  \hspace{0.12em}%
  \tcbox[
    on line,
    colback=#1!25,
    colframe=#1!60!black,
    boxrule=0.4pt,
    arc=1.5pt,
    boxsep=0pt,
    left=3pt, right=3pt,
    top=1pt, bottom=1pt,   %
  ]{\small\texttt{\rule[-0.2em]{0pt}{0.8em}#2}}%
  \hspace{0.1em}%
}

\title{\moto: Mixture of Tokenizers\\Towards Fair Multilingual Language Modeling}

\author{%
G\"ul Sena Alt{\i}nta\c{s}\hspace{2cm}Colin Raffel\\
University of Toronto, Vector Institute\\
Canada\\
\texttt{\{gsaltintas,craffel\}@cs.toronto.edu} 
}

\newcommand{\multilingualaddition}{\href{https://huggingface.co/datasets/r-three-moto/multilingual-addition}{r-three-moto/multilingual-addition}}
\newcommand{\hflink}{\href{https://huggingface.co/r-three-moto}{r-three-moto}}
\newcommand{\githubmoto}{ \href{https://github.com/r-three/moto}{r-three/moto}}
\newcommand{\githublingua}{\href{https://github.com/r-three/lingua/tree/moto}{r-three/lingua}}

\begin{document}
\maketitle

\begin{abstract}
Large language models rely on a single tokenizer that is chosen at training time and fixed thereafter.
Most tokenizers are trained on English-dominant corpora, making them under-serve morphologically rich and non-Latin-script languages, where they can produce fragmented representations, longer 
sequence lengths, and reduced information density during training.
We propose \textit{Mixture of Tokenizers} (\moto), a modular 
tokenization framework that trains dedicated per-language BPE tokenizers and composes them into a unified superset vocabulary, where tokens shared across languages are deduplicated into a common ID space. Each tokenizer maintains its own normalization policy, pre-tokenization rules, and vocabulary budget.
Despite allocating significantly fewer entries to English, \moto retains English performance while delivering consistent gains across 18 out of 21 typologically diverse languages spanning six scripts, with improvements most pronounced on morphologically complex and non-Latin-script languages that are systematically underserved by English-dominant tokenizer design. Our results suggest that modular subword tokenization is a practical and extensible alternative to monolithic tokenization for multilingual parity.

\end{abstract}

\vspace{-1ex}
\begin{center}
\setlength{\tabcolsep}{4pt}
\begin{tabular}{@{}l@{\hspace{2em}}l@{}}
  \raisebox{-0.4ex}{\includegraphics[height=2.5ex]{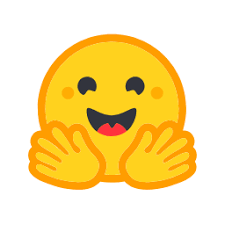}}~\textbf{Models \& Tokenizers:} \hflink
  &
  \raisebox{-0.4ex}{\includegraphics[height=2.5ex]{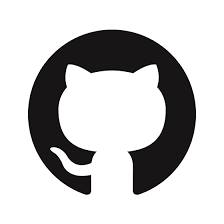}}~\textbf{Code:} \githubmoto,~\githublingua \\[0.5ex]
  \faGlobe~\textbf{Project page:} \href{https://gsaltintas.github.io/moto}{gsaltintas.github.io/moto}
  &
  \faCompass~\textbf{Explore tokenizers:} \href{https://tokenizers.gsaltintas.com/}{tokenizers.gsaltintas.com}
\end{tabular}
\end{center}

\begin{figure*}[tp]
    \centering
    \includegraphics[width=0.5\linewidth]{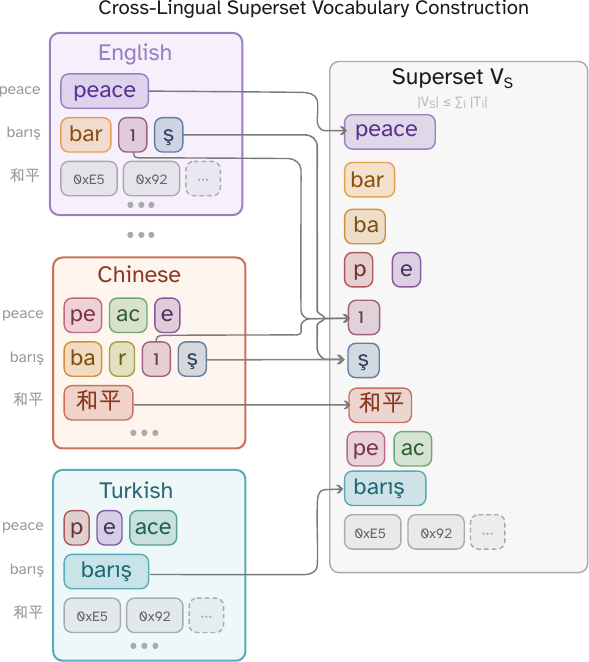}
    \hfill 
    \vrule
    \hfill
    \includegraphics[width=0.45\linewidth]{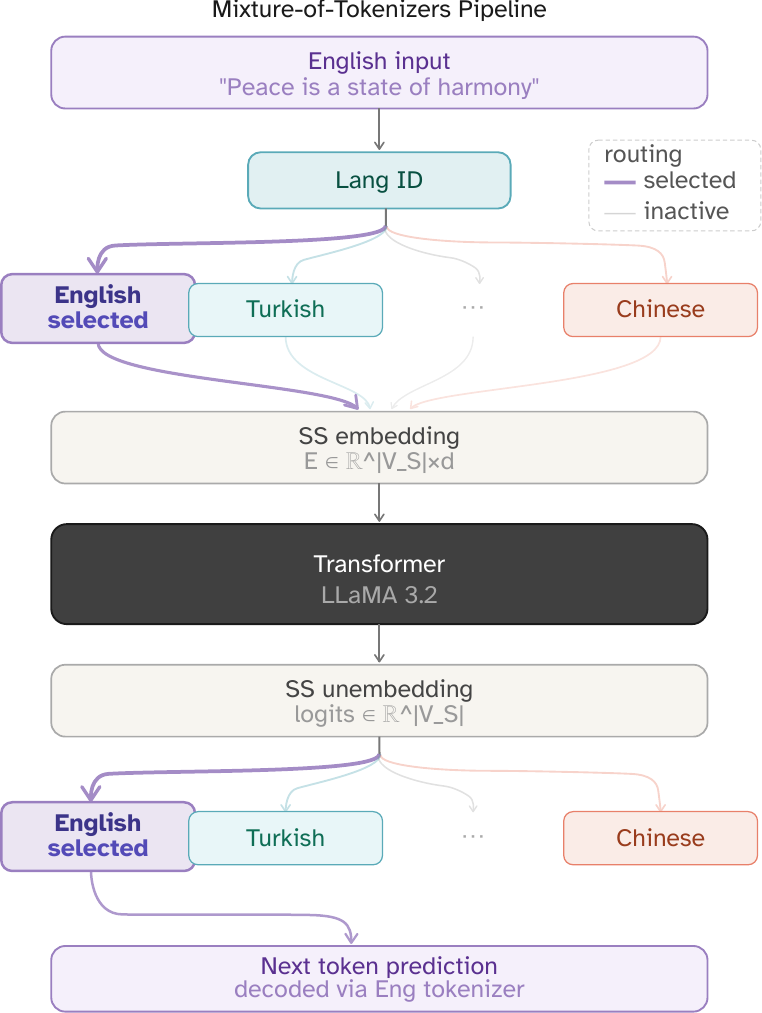}
    \caption{\textit{Left: Superset Vocabulary construction.} Language-specific subword tokenizers are trained on their corresponding corpus. Words native to that language tend to be represented with fewer tokens, while non-native words fall back to coarser subwords or byte-level tokens. The superset vocabulary $\mathcal{V}_S$ is constructed by taking the union of all tokenizer vocabularies, assigning shared tokens a single global index, with $\lvert \mathcal{V}_S\rvert \leq \sum_i \lvert T_i\rvert$. 
    \textit{Right: Mixture-of-Tokenizers (MoTo) pipeline.} 
    A tokenizer is selected based on the input language ID. The input is encoded into superset indices, 
    processed by the transformer, and decoded back with the tokens from the superset, additionally it can utilize rules from the input tokenizer.
  }
    \label{fig:superset:pipeline}
\end{figure*}

\section{Introduction}
Modern language models (LMs) are trained and deployed across a wide range of languages and scripts, yet rely on a single static tokenizer that is fixed early on. This is not an incidental limitation: tokenization is the first and least revisited design decision in language model training~\citep{alqahtani2026-stopGranted}, and its 
consequences propagate forward into  downstream representations. 
A base model is trained on general-purpose web corpora (e.g.,~\cite{penedo2024fineweb,grattafiori2024llama3}), and its fine-tuned, instruction-tuned, and RLHF-aligned variants inherit the same tokenizer.
However, many widespread tokenizers are trained primarily on English-dominant natural text, resulting in suboptimal performance when deployed across non-English languages~\citep{rust2021howGood,ali2024tokenizerChoice}.

A tokenizer consists of a vocabulary of subword units associated with learned embedding vectors and a segmentation algorithm that maps raw text into sequences over that vocabulary, both of which are  typically learned together based on a target vocabulary size and a choice of learning algorithm.
The vocabularies learned by two dominant learning algorithms, byte-pair encoding (BPE)~\citep{sennrich2016BPE, gage1994bpe} and UnigramLM~\citep{kudo2018sentencepiece}, are both primarily determined by the choice of training corpus and target vocabulary size, alongside discrete design decisions such as normalization rules and pre-tokenization patterns. 

However, designing a good tokenizer is far from straightforward. Byte-level compression, measured as the average number of tokens required to encode a given amount of text, is the dominant proxy~\citep{goldman2024unpacking}, as it directly affects how much information the model can consume during training and its inference efficiency, but it is an insufficient measure of quality~\citep{schmidt2024Compression}. 
Early tokenizers, largely designed for single languages or domains, incorporated strong domain-specific inductive biases through pre-processing choices that improved compression and simplified vocabulary. Both BERT~\citep{devlin2019BERT} and T5~\citep{raffel2020T5} collapse multiple whitespace characters into one; benign for natural language but destructive for Python, where indentation is syntactically significant.
Each of these decisions was locally motivated: it improved compression or reduced sparsity within the target domain. But as models, and therefore tokenizers, expanded to cover more languages and scripts, these assumptions became constraints. The field has since converged toward minimal pre-processing interventions, with Llama-3.2~\citep{grattafiori2024llama3}, Qwen-3~\citep{yang2025Qwen}, Tekken~\citep{mistral2024tekken}, Comma~\citep{kandpal2026comma}, and Aya~\citep{ustun2024aya} 
all adopting variants of the GPT-2 pre-tokenization 
regex~\citep{radford2019GPT2,openai2024gpt4technicalreport},  likely motivated by the following observation:
\begin{observation}[motoDarkPurple]\label{emph:every_domain}
     Preprocessing choices that improve efficiency for a target domain encode inductive biases that might not transfer across languages and scripts.
\end{observation}

Even so, minimizing pre-processing and treating tokenization as an end-to-end problem of finding co-occurring byte sequences does not resolve the issue that learned merges are driven by frequency, favoring languages that are better represented in the training corpus, leaving others with less vocabulary coverage.
A related but distinct concern is representability: to guarantee that \textit{any} input can be represented regardless of script or training coverage, most tokenizers adopt \textit{byte fallback} so that any character unseen during training can still be encoded in terms of its UTF-8 byte representation. 

Script-level asymmetries compound these disparities: languages without whitespace delimiters, like Chinese and Japanese, require explicit pre-tokenization decisions. Such differences in writing systems can lead to substantial differences in the number of tokens needed to encode equivalent content, a phenomenon known as the token premium~\citep{petrovLanguageModelTokenizers2023}.
Moreover, the byte premium~\citep{arnett2024bytePremium} of non-Latin scripts can translate into additional vocabulary overhead, as encoding a multi-byte character may require intermediate BPE merges, resulting in \emph{scaffold tokens} that serve primarily as building blocks for larger tokens~\citep{lian2024scaffoldbpe}.
Differences in byte-level representation can further affect vocabulary allocation, as frequent byte sequences and intermediate merges may occupy vocabulary slots without providing equally useful units across languages~\citep{arnett2024bytePremium,lian2024scaffoldbpe}. Some models address these imbalances by augmenting their vocabularies with frequent words from particular languages. For instance, Qwen~\citep{yang2025Qwen} and DeepSeek~\citep{deepseek2025v3} augment their vocabularies with frequent Chinese words, but such adjustments are ad hoc and do not generalize across 
languages or domains. 

More fundamentally, the learned vocabulary itself is rarely the object of deliberate design: it is treated as a byproduct of corpus frequency and vocabulary size, investigated only after the fact if at all. This has concrete consequences for multilingual fairness. For instance: despite its large vocabulary, Gemma-2~\citep{gemma2024gemma2} dedicates six separate entries to surface-level variations of the word \texttt{hello} (\tokenbox{motoPurpleBg}{hello}, \tokenbox{motoBlueBg}{\_hello}, \tokenbox{motoGreenBg}{Hello}, \tokenbox{motoYellowBg}{\_Hello}, \tokenbox{motoOrangeBg}{HELLO}, \tokenbox{motoRedBg}{\_HELLO}), resulting in embedding table entries allocated to English orthographic redundancy that could otherwise serve underrepresented languages. These limitations motivate a fundamental reexamination of how tokenizer vocabularies are constructed:

\begin{observation}[motoTeal]\label{emph:equity}
    A single jointly trained tokenizer cannot simultaneously serve all languages and domains well and a large vocabulary size does not guarantee multilingual or multi-domain coverage.
\end{observation} 

Recent work has sought to address these limitations through linguistically motivated relaxations~\citep{liu2025superbpe,schmidt2025boundlessBPE}, further refinement of vocabulary~\citep{chizhov2024PickyBPE,bauwens2024bpeKnockout} or by transferring and adapting existing tokenizers to new languages post-hoc~\citep{tran2020foreignTransfer,minixhofer2022wechsel,gee2022fastTransfer,dobler2023FOCUS,liu2024OFA,goddard2025tokenSurgeon,minixhofer2024xett}. In this paper, we question the fundamental assumption that a single jointly trained tokenizer should serve all languages and instead propose \moto: \textbf{Mixture of Tokenizers} as a modular-first design that treats the tokenizer as a composable artifact rather than a monolithic decision. 
Concretely, we make the following contributions:
\begin{itemize}
    \item \textbf{Modular superset tokenization. } We train small, dedicated BPE tokenizers per language across 21 languages with language-specific design choices, and compose them into a unified superset vocabulary without modifying the underlying model architecture (\S~\ref{sec:tokenizer_design} and Fig.~\ref{fig:superset:pipeline}).
    \item \textbf{Controlled multilingual  benchmark.} We introduce a synthetic multilingual addition task, and show that \moto with a superset of 2K tokens outperforms all variants while linguistic diversity prevents a jointly trained tokenizer from serving all languages equitably  (\S~\ref{sec:maddition}).
    \item \textbf{\moto models. } We train 1B non-embedding parameter decoder-only language models with \moto via language-ID-based routing, where each input is tokenized by its dedicated language tokenizer at training and inference time. \moto models improve compression and parity across all twenty non-English languages (\S~\ref{ap:sec:intrinsics}) and achieve consistent multilingual performance gains while preserving English 
    performance (O.~\ref{obs:eng_performace}), with the largest improvements on morphologically complex and non-Latin-script languages (O.~\ref{obs:moto_benefit}).
    \item \textbf{Ablations and analysis. } We ablate per-language vocabulary size, routing strategies (oracle, regularized, and random), and tokenizer granularity (per-language vs.\ per-script-family). We additionally train a model combining fourteen popular preexisting tokenizers into a single superset, providing a large-scale point of comparison (O.~\ref{obs:vocab_size},~\ref{obs:extending_toks}, and~\ref{obs:routing}).
\end{itemize}

\section{Background\label{sec:background}}

Modern language models operate on \tokenbox{motoPurpleBg}{tokens}, discrete units of text from a vocabulary~\citep{mielke2021survey}.
While tokenization can operate on individual bytes or whole words, subword tokenization segments raw text into units of varying length, capturing frequent words as single tokens while decomposing rare ones into smaller pieces. 

\paragraph{Learning algorithm.} Subword tokenization algorithms learn a vocabulary from a training corpus based on frequency. Byte-pair encoding (BPE)~\citep{gage1994bpe,sennrich2016BPE} starts from an initial symbol set that can tokenize any input text, such as  Unicode byte codes, and iteratively merges the most frequent adjacent pair until the target vocabulary size is reached.
UnigramLM~\citep{kudo2018sentencepiece} initializes from a large seed vocabulary and iteratively prunes the tokens whose removal increases loss on a corpus the least. Both are primarily determined by the training corpus and target vocabulary size. In this work we use BPE, though \moto's design is agnostic to this choice.

Beyond the algorithm choice, every tokenizer embodies a pipeline of interacting decisions: how to normalize Unicode codepoints, how to segment raw text into coarse units before 
subword learning begins, what corpus to train on, and how large the vocabulary should be. 
These decisions are fixed before training, making them among the most consequential and least revisited choices in the entire pipeline.
Each decision encodes assumptions about the intended usage of the model, but could break down when applied across typologically diverse languages and scripts. Below, we review the four axes along which these assumptions create challenges in multilingual settings, then describe how \moto addresses each as an explicit, per-language design choice.

\subparagraph{Pre-tokenization. } Before learning, most tokenizers split text into coarse segments based on predefined rules. This is applied identically during model training and inference and can substantially affect model performance~\citep{schmidt2024Compression}. The GPT-2 regex~\citep{radford2019GPT2} and its GPT-4 extension~\citep{openai2024gpt4technicalreport} have become de facto standards. Both encode English-specific assumptions: The contraction pattern \texttt{(?i:'s|'t|'re|'ve|'m|'ll|'d)} splits apostrophe-separated sequences as separate tokens, which is linguistically motivated for English (\textit{don't} 
$\rightarrow$ \tokenbox{motoYellowBg}{don}, \tokenbox{motoOrangeBg}{'t}. However, these rules can be problematic for other languages; for example, the \texttt{'d} rule would split the grammatical suffixes ``-de/-da'' in Turkish: \textit{Türkiye'de} (``in Turkey'') is 
split as \tokenbox{motoRedBg}{Türkiye}, \tokenbox{motoGreenBg}{'d}, \tokenbox{motoPurpleBg}{e}.
For Chinese and Japanese, the whitespace splitting assumption fails more fundamentally since these scripts do not use whitespace delimiters; without explicit character-level boundaries, BPE merges could then span entire sequences, which is intractable during tokenizer training.

Number handling is also not straightforward: digits can be grouped into fixed-size chunks (e.g., groups of three), segmented from left to right (\tokenbox{motoTealBg}{202}-\tokenbox{motoGreenBg}{6}), as in the Llama-3 tokenizer, from right to left (\tokenbox{motoOrangeBg}{2}-\tokenbox{motoYellowBg}{026}), as in Aya, or treated individually (\tokenbox{motoPurpleBg}{2}-\tokenbox{motoBlueBg}{0}-\tokenbox{motoPinkBg}{2}-\tokenbox{motoRedBg}{6}) as in Gemma-2.
These choices are reported to affect arithmetic performance~\citep{singhTokenizationCountsImpact2024}, while the appropriate segmentation also depends on whether numbers represent quantities, dates, or identifiers~\citep{lee2024digitstodecisionsHFBlog}.
Finally, Unicode boundary handling matters for scripts where a base character and its combining marks must be kept together as an atomic unit because splitting them would produce byte sequences that do not form a valid Unicode code point.

\subparagraph{Unicode normalization.} The same glyph can be encoded as multiple distinct codepoint sequences, introducing surface variations that are visually invisible but can change tokenization drastically. While Unicode normalization could resolve this ambiguity, this choice is not always lossless. Most popular tokenizers, (e.g., Llama-3, Gemma-2, and GPT-2) apply no normalization. Qwen-3 and Aya apply NFC, which resolves canonical codepoint variations while preserving semantic distinctions. NFKC, used by XGLM~\citep{lin2022XGLM-XStoryClose}, conflates visually similar but semantically distinct characters, destroying orthographic distinctions that are meaningful in Arabic, Farsi, and Japanese.
NFC is safer but with a nuance: combining marks such as shadda 
(\textsc{u+0651}) and kasra (\textsc{u+0650}) in Arabic interact under canonical ordering in ways that can reorder diacritic sequences, though this does not change meaning. We find NFC to be the best choice empirically across all 21 languages in early ablations.

\subparagraph{Frequency-driven vocabulary allocation. } 
Languages and constructions that appear more frequently in the tokenizer-training corpus receive dedicated vocabulary entries, while underrepresented languages are left closer to the byte level, requiring more tokens to express the same content.
This is particularly consequential for agglutinative languages such as Turkish and Hungarian, or root-pattern languages like Arabic
where the same root appears in many distinct surface forms across inflectional paradigms; BPE may merge across morpheme boundaries inconsistently depending on which surface forms are most frequent, producing different segmentations for semantically related words. 
Moreover, the most likely segmentation of a string varies with language, since token frequencies differ across languages~\citep{meister2026LID}.

Although morphologically informed tokenizers have been proposed~\citep{jabbar2024morphpiece,ehsaneddin2025morphbpe,bayram2026turkishTokenizer},  there is no strong evidence at scale that enforcing morphological boundaries improves downstream performance. Over-segmentation increases token count and can hurt rather than help. 
Ideally, tokens should carry roughly uniform information content across languages and morphological structures, rather than the arbitrary boundaries of any single linguistic theory.

\subparagraph{Vocabulary size.} Vocabulary size controls the embedding parameter budget and interacts with almost every other tokenizer design choice. A larger vocabulary reduces fertility (number of tokens produced for a given string) and allows more linguistically coherent tokens, but does not guarantee multilingual coverage if the tokenizer-training corpus lacks diversity. Moreover, scaffold tokens~\citep{lian2024scaffoldbpe}, undertrained entries or glitch tokens~\citep{li2024glitch,land2024magikarp}, vocabulary items that appear rarely during model training, can cause unpredictable model behavior at inference.
In multilingual settings a further confound arises: a language that dominates the tokenizer-training corpus will dominate the vocabulary regardless of size, crowding out underrepresented languages regardless of how large $|\mathcal{V}|$ is set.

\subsection{MoTo Tokenizer Design \label{sec:tokenizer_design}}

Instead of requiring a single monolithic tokenizer across languages, \moto treats tokenizer design choices on a per-language basis. This allows the tokenizer to better respect well-defined typographical conventions without imposing English-specific assumptions.
As a starting point, we train separate BPE tokenizers for 21 languages: English (\texttt{fw\_edu}) and those in FineWeb2-HQ~\citep{messmer2025fineweb2hq} chosen to span diverse scripts and morphological typologies. These include: Germanic (German~(\texttt{deu\_Latn}), Dutch~(\texttt{nld\_Latn}), Swedish~(\texttt{swe\_Latn}), Danish~(\texttt{dan\_Latn})), Romance (French~(\texttt{fra\_Latn}), Spanish~(\texttt{spa\_Latn}), Italian~(\texttt{ita\_Latn}), Portuguese~(\texttt{por\_Latn})), Slavic (Polish~(\texttt{pol\_Latn}), Czech~(\texttt{ces\_Latn}), Russian~(\texttt{rus\_Cyrl})), Austronesian (Indonesian~(\texttt{ind\_Latn})), and Austroasiatic (Vietnamese~(\texttt{vie\_Latn})) language families, alongside Hungarian~(\texttt{hun\_Latn}), Turkish~(\texttt{tur\_Latn}), Arabic~(\texttt{arb\_Arab}), Farsi~(\texttt{fas\_Arab}), Greek~(\texttt{ell\_Grek}), Chinese~(\texttt{cmn\_Hani}), and Japanese~(\texttt{jpn\_Jpan}). 
Together, this selection covers agglutinative, fusional, isolating, and root-pattern morphological typologies across Latin, Cyrillic, Arabic, Greek, and logographic scripts. 

We evaluate each setting with \texttt{TokEval}~\citep{meisterTokEvalTokenizerAnalysis2026} for cross-lingual parity for every language in Appendix~\ref{ap:sec:intrinsics}.
We discuss further design choices below:

\begin{wrapfigure}{r}{0.45\linewidth}
    \centering
    \includegraphics[width=\linewidth]{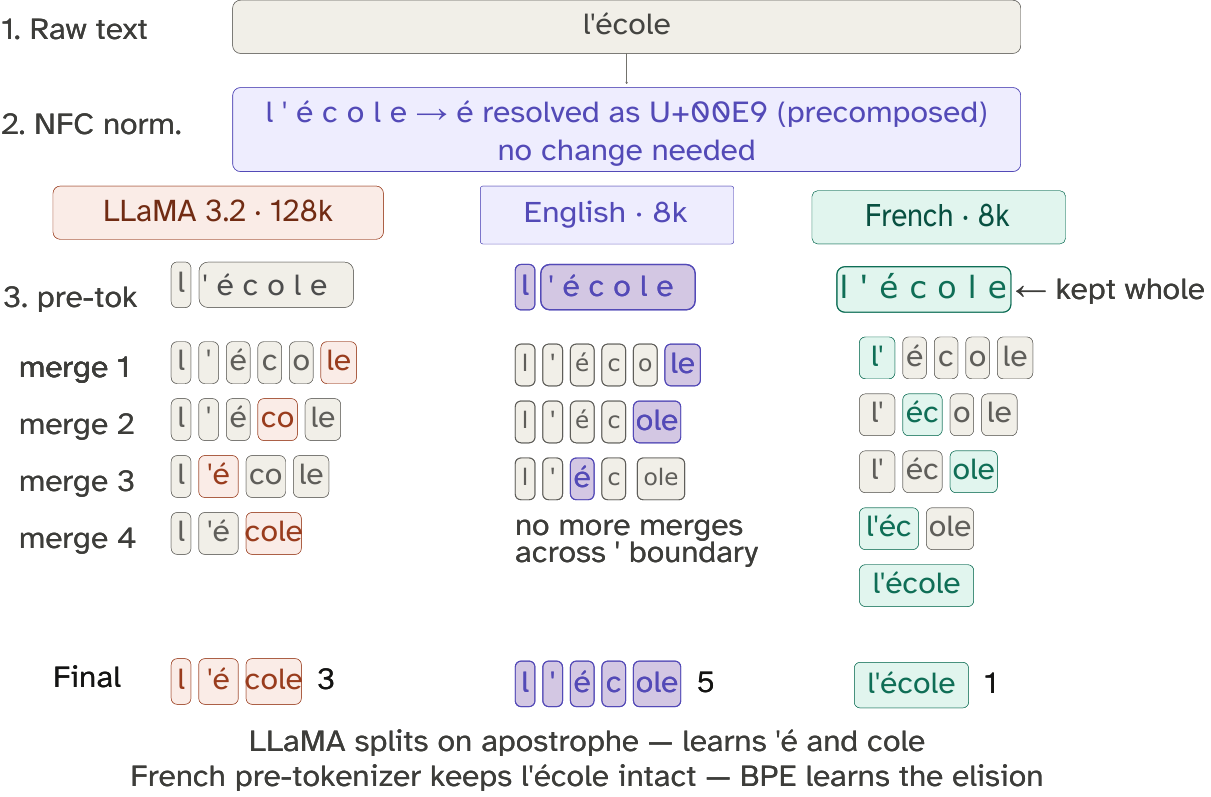}
    \vspace{-1.5ex}
    \caption{Tokenizing ``l'\'ecole'' with three BPE tokenizers with different pre-tokenization rules and vocabularies. Llama and the English tokenizer split on the apostrophe, preventing any merge across the elision boundary. The French tokenizer preserves \textit{l'école} as a single pre-token, allowing BPE to learn the elision \textit{l'} as a unit and compress the word to a single token. Highlighted tokens show the pair merged at each step.}
    \label{fig:bpe_merges_ecole}
\end{wrapfigure}
\paragraph{Language-aware pre-tokenization. } Rather than using a single global regex, we assign each language a custom pre-tokenization pattern inspired by Boundless BPE~\citep{schmidt2025boundlessBPE}, while preserving whitespace and punctuation splits. 
For example, Latin-script languages use a case-aware split that preserves apostrophe-joined forms as single pre-tokens important for French (\textit{j'ai}, \textit{l'\'ecole}), Italian (\textit{dell'a}), and others where elision is grammatically meaningful as demonstrated in Figure~\ref{fig:bpe_merges_ecole}. One can further extend this design to use Script-BPE~\citep{land2025scriptBPE}.
 \moto is agnostic to the choice of pre-tokenizer; the script-aware patterns are a default that can be replaced without modifying the rest of the pipeline.

\paragraph{Normalization. } We apply NFC normalization uniformly across all languages. 
Although Arabic and Farsi have combining mark interactions that require care under NFC, our experiments show that applying NFC uniformly outperforms omitting normalization for these languages. The reduction in codepoint variation outweighs the rare combining mark reordering edge cases, which does not change meaning.

\paragraph{Initial vocabulary and vocabulary size. } Beyond bytes, we seed Chinese and Japanese vocabularies with the most frequent Han, Hiragana, and Katakana characters covering 99.9\% of the respective tokenizer training corpus, ensuring single tokens for common characters rather than split byte sequences across Unicode code points.
We use left-to-right three digit number tokenization, and add all integers in $[0,\dots,999]$. 
Unless otherwise stated, target vocabulary size is 8,000 (8K) per language.

\paragraph{Superset composition. } The 21 individual tokenizers are combined into a single superset vocabulary by taking the union of all token strings based on their UTF-8 byte representations following \citet{toksuite}. 
The upper bound on superset size is $21 \times |\mathcal{V}|$, but languages sharing a script contribute overlapping subwords, reducing the effective cost per language. The resulting superset vocabulary sizes for each \moto setting are reported in Table~\ref{tab:moto_configs}. For the 21 8K-token vocabularies we train, $\lvert\mathcal{V_S}\rvert$ is below 120K.  
Each tokenizer maintains a mapping from its local token IDs to the global superset IDs (see Figure~\ref{fig:superset:pipeline}, left), we additionally reindex the tokenizers to eliminate additional lookup.

\paragraph{\moto routing. } During training, each input is routed to its dedicated tokenizer based on its language ID, tokenized locally, and then mapped to the shared superset vocabulary space before being passed to the model. 
At inference, the model predicts token IDs in the superset space. Our routing scheme requires no modification to the model architecture, only the embedding and unembedding matrices reflect the superset vocabulary size. When the language ID is unavailable, routing can fall back
to fastText-based language identification~\citep{joulin2016fasttext} or a base tokenizer.

\paragraph{\moto setting. } Unless otherwise stated, we refer to the combination of language-specific pre-tokenization, NFC normalization, and 8K per-language vocabulary as the default \moto configuration.
The full pipeline is illustrated in Figure~\ref{fig:superset:pipeline} right.

\section{A Controlled Study on Multilingual Addition\label{sec:maddition}}
We first study modular tokenization on a controlled synthetic task of multilingual addition across 21 languages\footnote{Available at \multilingualaddition.}.
Given two numbers expressed as words in a target language, the model must complete the equation, e.g. ``eight hundred and forty-five+five hundred and forty-five='' and produce ``one thousand, three hundred and ninety''. 
Generated in parallel across all 21 languages using the \citet{num2words} package and a digit-only split, the task is minimal and fast to iterate on with small scale models, yet exposes rich linguistic diversity in number systems from script variation (Arabic, Chinese, Japanese) to numeral compounding conventions (e.g. German, Italian).

We train a shallow decoder-only transformer ($d=256$, 4 layers, 4 attention heads, without weight-tying) with approximately 3.1M non-embedding parameters for 5000 steps ( Appendix~\ref{ap:sec:training}). 
Our \moto setting for this controlled experiment uses a lightweight superset of 1,999 tokens (256 byte tokens and at most 100 merges learned independently per language) with ID-based routing, a scaled-down instantiation of the design described in \S\ref{sec:tokenizer_design}.
We compare \moto against three baselines: \emph{mBPE}: a monolithic multilingual tokenizer trained jointly on all languages at 2K and 42K vocabulary sizes; 
\emph{Llama}, the Llama-3.2~\citep{grattafiori2024llama3} tokenizer with 128,256 tokens; and 
\emph{ByT5}~\citep{xue2022byt5} a byte-level baseline with 256 tokens.

\begin{figure}[!htbp]
    \centering
    \includegraphics[width=0.9\linewidth]{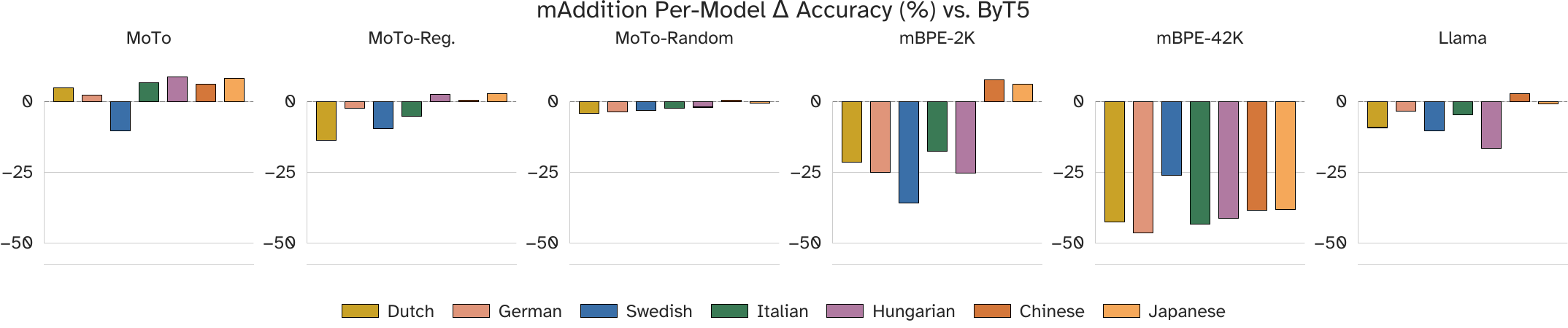}
    \vspace{-.75ex}
    \caption{
    Per-language accuracy delta relative to ByT5 on multilingual addition, shown for the seven languages with the largest cross-tokenizer variance. \moto and \moto-Reg. match or exceed ByT5, while jointly trained tokenizers (mBPE) degrade substantially for languages with complex numeral or morphological structure.
    }
    \label{fig:maddition_big_diff}
\end{figure}

\begin{observation}[motoTeal]\label{obs:maddition}
 Even on a perfectly parallel dataset, linguistic diversity prevents a jointly trained tokenizer from serving all languages equitably, motivating a modular design.
 \end{observation}
 
Despite a superset of only 1,999 tokens, \moto achieves the highest average accuracy with the largest gains on languages with complex numeral morphology (Figure~\ref{fig:maddition_big_diff}), while still performing best on others (Figure~\ref{fig:maddition_small_diff}). 
Both mBPE variants underperform on seven languages (Figure~\ref{fig:maddition_big_diff}): languages such as Hungarian and Italian form compound number words, and under a jointly trained tokenizer these long forms consume a disproportionate share of merge operations at the expense of general linguistic structure.

\textit{How sensitive is \moto to routing quality?} Since correct language-aware routing is central to \moto's design, we additionally evaluate two routing ablations to assess its importance: \textbf{\moto-Reg.}, which introduces noise by routing 10\% of inputs to a random tokenizer, and \textbf{\moto-Random}, which routes all inputs randomly and serves as a lower bound on routing quality. \moto-Reg and \moto-Random perform comparably with ByT5, confirming that gains are driven by language-aware routing (Figures~\ref{fig:maddition_big_diff} and~\ref{fig:maddition_small_diff}).

\section{Multilingual Language Model Pre-training on FineWeb2 with \moto\label{sec:mmoto}}
Having established \moto's effectiveness in a controlled setting, we now evaluate it at a realistic scale.
We train a Llama~3.2~\citep{grattafiori2024llama3} style decoder-only model with approximately one billion non-embedding parameters from scratch on a 100B-token multilingual corpus of approximately 40B English tokens from FineWeb-Edu~\citep{lozhkov2024fineweb-edu} and 60B tokens across 20 languages from FineWeb2-HQ~\citep{messmer2025fineweb2hq} with sampling rates set following the UniMax strategy~\citep{chung2023unimax}. Full training details and data composition are reported in Appendix~\ref{ap:sec:training}.

We evaluate the downstream performance on various benchmarks suitable for base models and report performance gains relative to the Llama-3 tokenizer-model baseline, trained on the same data and compute budget. 
As we train a single model per setting, error bars do not reflect seed variance instead capture test-set sampling uncertainty. For each task, we compute it from the per-example differences $d_i$ between the two models' outcomes on the same example, $\mathrm{SE}=\mathrm{std}(d)/\sqrt{n}$, using the per-sample logs of \texttt{lm-eval-harness}~\citep{lm-eval-harness}. We combine standard errors across tasks within a language, and across languages for the average, assuming independence.
Full results on a diverse set of benchmarks are presented in Appendix~\ref{ap:sec:full_results}. 

\paragraph{Tokenization.} The tokenizer is the only component that varies across settings. 
We use \moto with language-specific pre-tokenizer and 8K per-language vocabulary as our primary configuration, constructed as described in \S\ref{sec:tokenizer_design}, with inputs routed to their dedicated language tokenizer via document language ID and mapped to the shared superset vocabulary of 119,716. 

\emph{Baselines.} We compare \moto against three jointly trained multilingual tokenizers: \emph{mBPE-GPT4} and \emph{mBPE-Boundless}, which differ only in their pre-tokenization regex (GPT-4 and Boundless BPE~\citep{schmidt2025boundlessBPE}, respectively), and \emph{Parity-Aware BPE}~\citep{foroutan2025parityBPE}. We also include the \emph{Llama}~3.2 tokenizer with and without BPE-Dropout~\citep{provilkov2020bpeDropout}, and a byte-level baseline as in ByT5~\citep{xue2022byt5}. Table~\ref{tab:moto_configs} summarizes all configurations; further design ablations are introduced alongside their results.

\begin{table}[t]
\small
\centering
\caption{Summary of all \moto tokenizer configurations and baselines.}
\label{tab:moto_configs}
\begin{tabular}{llllr}
\toprule
\textbf{Configuration} & \textbf{Pre-tokenizer} & 
\textbf{Granularity} &
\textbf{Vocab size} \\
\midrule
\moto            & language-specific + NFC & per language, 8K  & 119,716 \\
\moto-GPT4       & GPT-4 + NFC        & per language, 8K  & 113,764 \\
\moto-16K        & language-specific + NFC & per language, 16K & 245,153 \\
\moto-Script     & language-specific + NFC & per script family & 165,022 \\
\moto-Llama-16K  & language-specific + NFC & per language, 16K, Llama for English & 323,711 \\
\midrule
Parity-Aware BPE & —  &  joint but parity-aware, 128K  &  128,000 \\ 
mBPE-GPT4        & GPT-4              & joint, 128K        & 128,000 \\
mBPE-Boundless   & Boundless BPE      & joint, 128K        & 128,000 \\
ByT5             & —                  & byte-level         & 259     \\
Llama            & GPT-4              &   —     & 128,256 \\
Llama-BPE-Drop.  & GPT-4              &  —    & 128,256 \\
\bottomrule
\end{tabular}
\end{table}

\subsection{Results}
We report per-benchmark accuracy differences (in percentage points) relative to our Llama-3 tokenizer-model. We choose Llama's tokenizer as baseline because it is of comparable vocabulary size to \moto and its multilingual coverage. All models are trained on identical data and compute.

\vspace{-.75ex}
\begin{observation}[motoTeal]\label{obs:eng_performace}
     \moto retains English performance despite allocating only 8K vocabulary entries, a fraction of Llama's vocabulary budget.
\end{observation}
\vspace{-.75ex}
Despite dedicating only 8K vocabulary entries to English, \moto matches Llama on English commonsense reasoning benchmarks (Table~\ref{ap:tab:english_perf}), suggesting that a compact language-specific vocabulary can be on par with a much larger monolithic one. This may also reflect a general 
model improvement from exposure to larger amounts of diverse multilingual data due to compression gains during training.

\begin{figure}[htbp]
    \centering
    \includegraphics[width=0.9\textwidth]{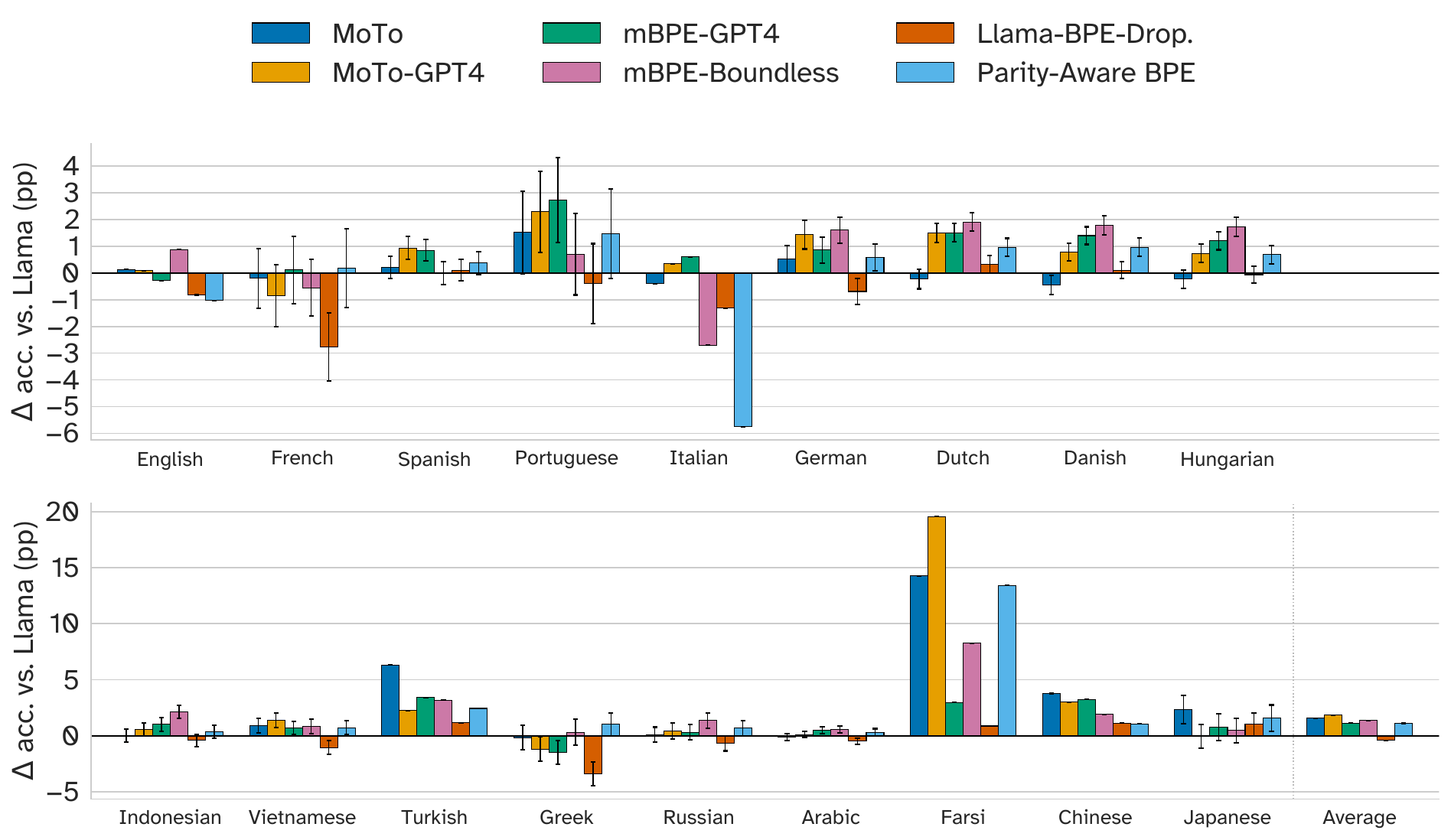}
      \vspace{-1.5ex} 
    \caption{\textbf{Main results per language.} Accuracy difference (pp) from the Llama-3 baseline, averaged over each language's tasks; the rightmost group is the average over languages. \moto and \moto-GPT4 use per-language tokenizers with oracle routing. mBPE-GPT4, mBPE-Boundless, and Parity-Aware BPE are single multilingual BPE tokenizers, and Llama-BPE-Drop.\ is the Llama tokenizer with BPE dropout.}
    \label{fig:main_delta}
\end{figure}

\vspace{-1.5ex}
\begin{observation}[motoDarkPurple]\label{obs:moto_benefit}
    Multilingual tokenization helps across the board, but non-Latin scripts and several smaller-share languages (Turkish, Vietnamese, Farsi) benefit the most.
\end{observation}
\vspace{-.75ex}
\moto and the mBPE baselines improve over the Llama baseline across most languages (Figure~\ref{fig:main_delta}).
This suggests that two things help: training the tokenizer on the same distribution as the language model, and segmenting each language with a vocabulary built for it. We observe the largest gains with \moto models on non-Latin scripts such as Farsi (Figure~\ref{fig:main_delta}), Chinese, and Japanese (Figures~\ref{fig:main_delta}, \ref{fig:paws}, \ref{fig:xwinograd}) or smaller-share languages like Turkish and Vietnamese (Figures~\ref{fig:main_delta},~\ref{fig:xcopa}). Improvements in high-resource Latin languages, like Italian, German, Spanish, and French remain modest.

\begin{figure}[htbp]
    \centering
    \includegraphics[width=0.870\textwidth]{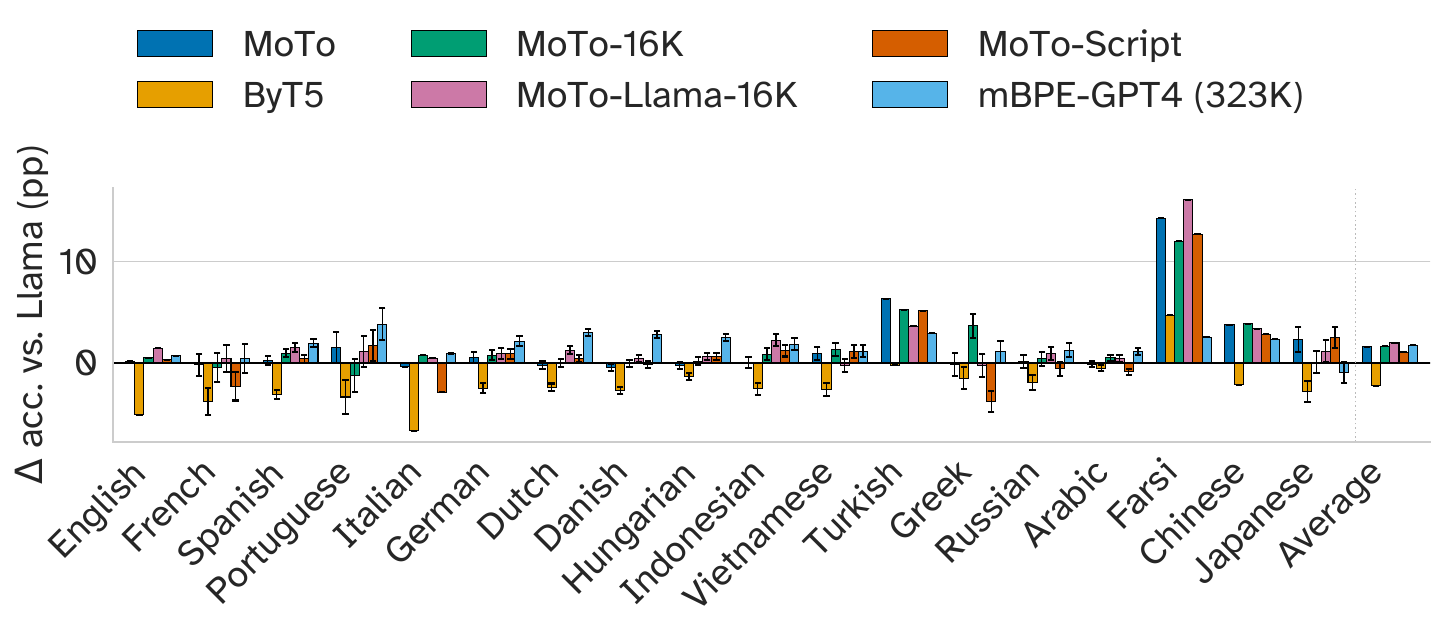}
  \vspace{-2ex} 
    \caption{\textbf{Vocabulary size and granularity.} Per-language accuracy difference (pp) from the Llama-3 baseline, averaged over each language's tasks; the rightmost group is the average over languages. \moto uses 21 per-language 8K BPE tokenizers; \moto-16K uses 16K per-language tokenizers; \moto-Llama-16K adds the Llama-3 tokenizer to the 16K per-language tokenizers as English tokenizer. ByT5 is byte-level (one token per byte), and mBPE-GPT4 (323K) is a single 323K BPE tokenizer trained jointly on all 21 languages. All \moto variants use oracle routing.}
    \label{fig:main_delta_vocab_size}
\vspace{-2ex}
\end{figure}

\vspace{-.75ex}
\begin{observation}[motoYellow]\label{obs:vocab_size}
    \moto's superset construction keeps embedding table growth sublinear in the number of languages while allowing for adaptive vocabulary expansion.
\end{observation}
\vspace{-1ex}
A natural concern with modular tokenization is the cost of maintaining multiple vocabularies: na\"ively, $K$ language-specific tokenizers of size $\lvert \mathcal{V}\rvert$ would require an embedding table of size $K \cdot \lvert \mathcal{V}\rvert$. However, MoTo's superset vocabulary construction amortizes this cost and the effective vocabulary grows sublinearly with the number of languages. As seen in Table~\ref{tab:moto_configs} combining 21 language-specific vocabularies of size 8K yields an effective superset of around 120K tokens. We observe non-trivial improvements with larger vocabulary variants (\moto-16K and \moto-Llama-16K) over \moto with 8K vocabularies in Figure~\ref{fig:main_delta_vocab_size}.

\begin{observation}[motoGreen]\label{obs:extending_toks}
    Modular tokenization enables targeted extension of existing tokenizers without retraining the full vocabulary.
\end{observation}
Beyond raw performance, our modular design offers a practical advantage over monolithic multilingual tokenizers. Training a single large multilingual vocabulary is notoriously difficult as shown in our multilingual addition example (O.~\ref{obs:maddition}). For example, while training early mBPE tokenizers, we observed that some languages were overrepresented in the vocabulary. Balancing token allocations across languages, avoiding fertility disparities, and auditing a 100K+ vocabulary for artifacts or script bleed requires significant engineering effort~\citep{rust2021howGood}. 
\moto mitigates this problem, as each tokenizer could be trained independently, making quality control substantially more tractable.

\vspace{-.75ex}
\begin{observation}[motoGray]\label{obs:routing}
    \moto could benefit from alternative routing strategies.
\end{observation}
\vspace{-.75ex}
\textit{Superset-Albert. }
A natural question is whether the benefits of modular composition extend to preexisting tokenizers.
We construct \textbf{Superset-Albert}, a superset of fourteen off-the-shelf widely used tokenizers studied in TokSuite~\citep{toksuite} with a total vocabulary of 851,586 tokens with ALBERT-style factorized embeddings~\citep{lan2020albert}. As there is no natural language-ID signal, we train this \moto variant with random routing.
We discuss the potential of routing and additional baselines for this setting in \S~\ref{ap:sec:routing}.

\section{Related Work}
\textit{Modular NLP \& multilinguality. }
Our work builds on the rich literature on modular architectures for multilingual generalization. Mixture-of-Expert architectures route inputs to specialized sub-networks~\citep{shazeer2017MOE}. This has been leveraged for multilingual adaptation through language-specific adapters that specialize pretrained models for individual languages without full fine-tuning~\citep{pfeiffer-etal-2020-mad,ansell-etal-2022-composable,
ustun-etal-2022-udapter}. Our framework (\moto) extends this principle to the tokenizer level.

\textit{Beyond subword tokenization. } While byte-level transformers~\citep{xue2022byt5,  minixhofer2025BOLMO} and pooling-based latent tokenization~\citep{yu2023megabyte, pagnoni2025BLT,hwang2025hnet} offer alternatives; they do not necessarily guarantee multilingual fairness due to the byte-premium problem~\citep{arnett2024bytePremium} that disproportionately burdens non-Latin-script languages. \citet{feher-etal-2025-retrofitting} propose dynamic tokenization with a hyper-network.
Vocabulary-free approaches~\citep{islam2022vocabularyFree}, and end-to-end tokenization via reinforcement learning~\citep{dauncey2026tokenizationRL} explore adjacent directions but lack evidence at scale.

Stochastic approaches such as subword regularization~\citep{kudo2018subword},  BPE-Dropout~\citep{provilkov2020bpeDropout},  StochasTok~\citep{sims2026stochastok} and its successor~\citep{steger2026stochasticityRobustness} introduce variability during training to improve robustness, and \citet{zheng2026brokenTokens} show that models can handle non-canonical tokenizations at inference time.

\textit{Tokenizer design. } Several works have relaxed individual constraints in tokenizer training. Boundless BPE~\citep{schmidt2025boundlessBPE} and SuperBPE~\citep{liu2025superbpe} show performance gains when tokenizers are allowed to learn merges across whitespace boundaries. Picky BPE~\citep{chizhov2024PickyBPE} and BPE-Knockout~\citep{bauwens2024bpeKnockout}  refine the vocabulary to eliminate redundant entries.

\textit{Tokenization and multilingual generalization. } Tokenization has been shown to have consequences for multilingual performance \citep{rust2021howGood,arnett2025Inequities,
arnett2024bytePremium,wegmann2025senstive,ali2024tokenizerChoice,toksuite}. 
\citet{ahia2023Cost} show that tokenization inequity manifests itself as higher fertility for morphologically rich and non-Latin-script languages, higher inference costs, and degraded downstream performance.
\citet{foroutan2025parityBPE} introduce parity-based regularization to BPE learning.
\citet{schmidt2024Compression} argue that compression is an insufficient proxy for tokenizer quality, and draw attention to pre-tokenization.
\citet{meister2026languagespecificeffectstokenizerchoice} show that tokenizer choice matters especially for low-resource languages, and that the tokenizer properties associated with downstream performance differ across languages.
\citet{abagyan2025tokenizerruleallemergent} show that a single multilingual tokenizer trained on many languages can achieve language plasticity in the downstream tasks.

\textit{Linguistic preprocessing tools.}
A rich tradition of language-specific NLP tools encodes deep linguistic knowledge for morphological analysis, normalization, and segmentation: Zemberek~\citep{zemberek} for Turkish, CamelTools~\citep{obeid2020camelArabic} 
and Farasa~\citep{abdelali2016farasaArabic} for Arabic, Hazm~\citep{hazmFarsi} and Parsivar~\citep{mohtaj2018parsivarFarsi} for Farsi, Morfeusz~\citep{wolinski2014morfeuszPolish} for Polish, MorphoDiTa~\citep{strakova2014MorphoditaCzech} for Czech, pymorphy3~\citep{pymorphy3} for Russian, HuSpaCy~\citep{HuSpaCy2021,HuSpaCy2023} for Hungarian, and spaCy~\citep{honnibal2020spacy} for a range of languages. 
Morphologically informed tokenizers have attempted to incorporate this knowledge directly into subword learning~\citep{jabbar2024morphpiece,ehsaneddin2025morphbpe,bayram2026turkishTokenizer,zhu2025morphTree}, though evidence for downstream gains remains mixed and limited to certain languages.

\section{Limitations \& Future Work\label{sec:limitations}}
We train one model per setting, so uncertainty reflects test-set sampling rather than seed variance.

\textit{Routing.} Our current routing operates at the sample level, assigning a single tokenizer to each sample based on its language ID. This coarse-grained approach may be suboptimal for code-switched or mixed-script inputs. While document-level language identification is well established~\citep{joulin2016fasttext}, several questions remain open: how to select a tokenizer for an unknown language, whether routing should operate at finer granularity, and whether routing decisions can be learned end-to-end during training. We discuss these challenges further in \S~\ref{ap:sec:routing}.

\textit{Subword sharing.} Currently, \moto allows modular design and segmentation, meaning that embeddings for subwords shared across languages are learned from signals from both. While sharing could benefit low-resource languages by allowing them to leverage representations learned from higher-resource languages, it could also introduce interference when shared subwords have different linguistic roles or distributions across languages. Understanding when cross-lingual sharing helps or hurts, and how it should be controlled, remains an open direction for future work.

\textit{Different allocations.} Currently, \moto allocates an equal share of the total vocabulary to each tokenizer. However, this allocation is a design choice rather than a constraint: vocabulary capacity could instead be distributed according to the language distribution in the model's training data, or based on other criteria such as language resource availability. Exploring alternative allocation strategies is an important direction for understanding the trade-offs between balanced multilingual coverage and overall model performance.

\textit{Instruction following.} We currently evaluate \moto on 1B base models. Extending \moto to instruction-tuned models and evaluating open-ended generation quality across languages remains an interesting and largely unexplored direction.

\section{Conclusion}
Multilingual tokenization is a largely overlooked component in language model development, where English-dominant design choices propagate into suboptimal downstream performance on multilingual tasks. In this paper we show that these shortcomings are inevitable even with multilingual tokenizers trained jointly on the target corpus. Instead, we propose \moto and show that treating the tokenizer as a modular, composable artifact yields consistent multilingual gains while preserving English performance, with no modification to the underlying model architecture.

\section*{Acknowledgments}
We are grateful to Chris J. Maddison and Marco Ciccone for fruitful discussions and constructive feedback throughout this project. We would like to thank the members of the r-three lab for feedback at several stages of development. 
We thank 
Sven Elflein and Noah Juravsky for feedback on this manuscript.
We thank Ahmadreza Jeddi and Malikeh Ehghaghi for clarifications on Farsi and Arabic, Bogl\'{a}rka Ecsedi on Hungarian, Haokun Liu on Chinese, Marco Ciccone on Italian, and Utku Ege Tuluk on Japanese.
Gül Sena would like to thank İdil Defne Çekin, Hazal Çintaş, Azmi Ozan Ateş, and Özgem Elif Acar for their continued support.

Resources used in preparing this research were provided, in part, by the Province of Ontario, the Government of Canada through CIFAR, the \href{https://www.alliancecan.ca}{Digital Research Alliance of Canada}, and companies sponsoring the \href{https://www.vectorinstitute.ai/partnerships/current-partners/}{Vector Institute}.
The authors do not have any financial conflict of interest.

\bibliographystyle{plainnat}  %
\bibliography{citations}

\begin{thebibliography}{110}
\providecommand{\natexlab}[1]{#1}
\providecommand{\url}[1]{\texttt{#1}}
\expandafter\ifx\csname urlstyle\endcsname\relax
  \providecommand{\doi}[1]{doi: #1}\else
  \providecommand{\doi}{doi: \begingroup \urlstyle{rm}\Url}\fi

\bibitem[zem(2024)]{zemberek}
Zemberek-nlp, 2024.
\newblock URL \url{https://github.com/ahmetaa/zemberek-nlp}.

\bibitem[Abagyan et~al.(2025)Abagyan, Salamanca, Cruz-Salinas, Cao, Lin,
  Locatelli, Fadaee, {\"U}st{\"u}n, and
  Hooker]{abagyan2025tokenizerruleallemergent}
Diana Abagyan, Alejandro~R. Salamanca, Andres~Felipe Cruz-Salinas, Kris Cao,
  Hangyu Lin, Acyr Locatelli, Marzieh Fadaee, Ahmet {\"U}st{\"u}n, and Sara
  Hooker.
\newblock One tokenizer to rule them all: Emergent language plasticity via
  multilingual tokenizers, 2025.
\newblock URL \url{https://arxiv.org/abs/2506.10766}.

\bibitem[Abdelali et~al.(2016)Abdelali, Darwish, Durrani, and
  Mubarak]{abdelali2016farasaArabic}
Ahmed Abdelali, Kareem Darwish, Nadir Durrani, and Hamdy Mubarak.
\newblock {F}arasa: A fast and furious segmenter for {A}rabic.
\newblock In John DeNero, Mark Finlayson, and Sravana Reddy, editors,
  \emph{Proceedings of the 2016 Conference of the North {A}merican Chapter of
  the Association for Computational Linguistics: Demonstrations}, pages 11--16,
  San Diego, California, June 2016. Association for Computational Linguistics.
\newblock \doi{10.18653/v1/N16-3003}.
\newblock URL \url{https://aclanthology.org/N16-3003/}.

\bibitem[Ahia et~al.(2023)Ahia, Kumar, Gonen, Kasai, Mortensen, Smith, and
  Tsvetkov]{ahia2023Cost}
Orevaoghene Ahia, Sachin Kumar, Hila Gonen, Jungo Kasai, David Mortensen, Noah
  Smith, and Yulia Tsvetkov.
\newblock Do all languages cost the same? tokenization in the era of commercial
  language models.
\newblock In Houda Bouamor, Juan Pino, and Kalika Bali, editors,
  \emph{Proceedings of the 2023 Conference on Empirical Methods in Natural
  Language Processing}, pages 9904--9923, Singapore, December 2023. Association
  for Computational Linguistics.
\newblock \doi{10.18653/v1/2023.emnlp-main.614}.
\newblock URL \url{https://aclanthology.org/2023.emnlp-main.614/}.

\bibitem[AI(2024)]{mistral2024tekken}
Mistral AI.
\newblock mistral-common.
\newblock \url{https://github.com/mistralai/mistral-common}, 2024.

\bibitem[Ali et~al.(2024)Ali, Fromm, Thellmann, Rutmann, L{\"u}bbering,
  Leveling, Klug, Ebert, Doll, Buschhoff, Jain, Weber, Jurkschat, Abdelwahab,
  John, Ortiz~Suarez, Ostendorff, Weinbach, Sifa, Kesselheim, and
  Flores-Herr]{ali2024tokenizerChoice}
Mehdi Ali, Michael Fromm, Klaudia Thellmann, Richard Rutmann, Max
  L{\"u}bbering, Johannes Leveling, Katrin Klug, Jan Ebert, Niclas Doll, Jasper
  Buschhoff, Charvi Jain, Alexander Weber, Lena Jurkschat, Hammam Abdelwahab,
  Chelsea John, Pedro Ortiz~Suarez, Malte Ostendorff, Samuel Weinbach, Rafet
  Sifa, Stefan Kesselheim, and Nicolas Flores-Herr.
\newblock Tokenizer choice for {LLM} training: Negligible or crucial?
\newblock In Kevin Duh, Helena Gomez, and Steven Bethard, editors,
  \emph{Findings of the Association for Computational Linguistics: NAACL 2024},
  pages 3907--3924, Mexico City, Mexico, June 2024. Association for
  Computational Linguistics.
\newblock \doi{10.18653/v1/2024.findings-naacl.247}.
\newblock URL \url{https://aclanthology.org/2024.findings-naacl.247/}.

\bibitem[Alqahtani et~al.(2026)Alqahtani, Nayeem, Laskar, Mohiuddin, and
  Bari]{alqahtani2026-stopGranted}
Sawsan Alqahtani, Mir~Tafseer Nayeem, Md~Tahmid~Rahman Laskar, Tasnim
  Mohiuddin, and M~Saiful Bari.
\newblock Stop taking tokenizers for granted: They are core design decisions in
  large language models.
\newblock In Vera Demberg, Kentaro Inui, and Llu{\'\i}s Marquez, editors,
  \emph{Proceedings of the 19th Conference of the {E}uropean Chapter of the
  {A}ssociation for {C}omputational {L}inguistics (Volume 1: Long Papers)},
  pages 8410--8432, Rabat, Morocco, March 2026. Association for Computational
  Linguistics.
\newblock ISBN 979-8-89176-380-7.
\newblock \doi{10.18653/v1/2026.eacl-long.394}.
\newblock URL \url{https://aclanthology.org/2026.eacl-long.394/}.

\bibitem[Alt{\i}nta{\c{s}} et~al.(2026)Alt{\i}nta{\c{s}}, Ehghaghi, Lester,
  Liu, Zhao, Ciccone, and Raffel]{toksuite}
G{\"u}l~Sena Alt{\i}nta{\c{s}}, Malikeh Ehghaghi, Brian Lester, Fengyuan Liu,
  Wanru Zhao, Marco Ciccone, and Colin Raffel.
\newblock {TokSuite}: Measuring the impact of tokenizer choice on language
  model behavior.
\newblock In \emph{Forty-third International Conference on Machine Learning},
  2026.
\newblock URL \url{https://openreview.net/forum?id=vIZz7LvObC}.

\bibitem[Ansell et~al.(2022)Ansell, Ponti, Korhonen, and
  Vuli{\'c}]{ansell-etal-2022-composable}
Alan Ansell, Edoardo Ponti, Anna Korhonen, and Ivan Vuli{\'c}.
\newblock Composable sparse fine-tuning for cross-lingual transfer.
\newblock In Smaranda Muresan, Preslav Nakov, and Aline Villavicencio, editors,
  \emph{Proceedings of the 60th Annual Meeting of the Association for
  Computational Linguistics (Volume 1: Long Papers)}, pages 1778--1796, Dublin,
  Ireland, May 2022. Association for Computational Linguistics.
\newblock \doi{10.18653/v1/2022.acl-long.125}.
\newblock URL \url{https://aclanthology.org/2022.acl-long.125/}.

\bibitem[Arnett et~al.(2024)Arnett, Chang, and Bergen]{arnett2024bytePremium}
Catherine Arnett, Tyler~A. Chang, and Benjamin Bergen.
\newblock A bit of a problem: Measurement disparities in dataset sizes across
  languages.
\newblock In Maite Melero, Sakriani Sakti, and Claudia Soria, editors,
  \emph{Proceedings of the 3rd Annual Meeting of the Special Interest Group on
  Under-resourced Languages @ LREC-COLING 2024}, pages 1--9, Torino, Italia,
  May 2024. ELRA and ICCL.
\newblock URL \url{https://aclanthology.org/2024.sigul-1.1/}.

\bibitem[Arnett et~al.(2025)Arnett, Chang, Biderman, and
  Bergen]{arnett2025Inequities}
Catherine Arnett, Tyler~A. Chang, Stella Biderman, and Ben Bergen.
\newblock Explaining and mitigating crosslingual tokenizer inequities.
\newblock In \emph{The Thirty-ninth Annual Conference on Neural Information
  Processing Systems}, 2025.
\newblock URL \url{https://openreview.net/forum?id=0Y7AxxNCYh}.

\bibitem[Asgari et~al.(2025)Asgari, Kheir, and
  Javaheri]{ehsaneddin2025morphbpe}
Ehsaneddin Asgari, Yassine~El Kheir, and Mohammad Ali~Sadraei Javaheri.
\newblock Morphbpe: A morpho-aware tokenizer bridging linguistic complexity for
  efficient llm training across morphologies.
\newblock \emph{CoRR}, abs/2502.00894, February 2025.
\newblock URL \url{https://doi.org/10.48550/arXiv.2502.00894}.

\bibitem[Bandarkar et~al.(2024)Bandarkar, Liang, Muller, Artetxe, Shukla, Husa,
  Goyal, Krishnan, Zettlemoyer, and Khabsa]{bandarkar2024belebele}
Lucas Bandarkar, Davis Liang, Benjamin Muller, Mikel Artetxe, Satya~Narayan
  Shukla, Donald Husa, Naman Goyal, Abhinandan Krishnan, Luke Zettlemoyer, and
  Madian Khabsa.
\newblock The belebele benchmark: a parallel reading comprehension dataset in
  122 language variants.
\newblock In Lun-Wei Ku, Andre Martins, and Vivek Srikumar, editors,
  \emph{Proceedings of the 62nd Annual Meeting of the Association for
  Computational Linguistics (Volume 1: Long Papers)}, pages 749--775, Bangkok,
  Thailand, August 2024. Association for Computational Linguistics.
\newblock \doi{10.18653/v1/2024.acl-long.44}.
\newblock URL \url{https://aclanthology.org/2024.acl-long.44/}.

\bibitem[Bauwens and Delobelle(2024)]{bauwens2024bpeKnockout}
Thomas Bauwens and Pieter Delobelle.
\newblock {BPE}-knockout: Pruning pre-existing {BPE} tokenisers with
  backwards-compatible morphological semi-supervision.
\newblock In Kevin Duh, Helena Gomez, and Steven Bethard, editors,
  \emph{Proceedings of the 2024 Conference of the North American Chapter of the
  Association for Computational Linguistics: Human Language Technologies
  (Volume 1: Long Papers)}, pages 5810--5832, Mexico City, Mexico, June 2024.
  Association for Computational Linguistics.
\newblock \doi{10.18653/v1/2024.naacl-long.324}.
\newblock URL \url{https://aclanthology.org/2024.naacl-long.324/}.

\bibitem[Bayram et~al.(2026)Bayram, Fincan, G{\"u}m{\"u}{\c s}, Karaka{\c s},
  Diri, Yıldırım, and {\c C}elik]{bayram2026turkishTokenizer}
M.~Ali Bayram, Ali~Arda Fincan, Ahmet~Semih G{\"u}m{\"u}{\c s}, Sercan
  Karaka{\c s}, Banu Diri, Sava{\c s} Yıldırım, and Demircan {\c C}elik.
\newblock Tokens with meaning: A hybrid tokenization approach for turkish,
  2026.
\newblock URL \url{https://arxiv.org/abs/2508.14292}.

\bibitem[Bisk et~al.(2020)Bisk, Zellers, Bras, Gao, and Choi]{bisk2020piqa}
Yonatan Bisk, Rowan Zellers, Ronan~Le Bras, Jianfeng Gao, and Yejin Choi.
\newblock {PIQA:} reasoning about physical commonsense in natural language.
\newblock In \emph{The Thirty-Fourth {AAAI} Conference on Artificial
  Intelligence, {AAAI} 2020, The Thirty-Second Innovative Applications of
  Artificial Intelligence Conference, {IAAI} 2020, The Tenth {AAAI} Symposium
  on Educational Advances in Artificial Intelligence, {EAAI} 2020, New York,
  NY, USA, February 7-12, 2020}, pages 7432--7439. {AAAI} Press, 2020.
\newblock \doi{10.1609/AAAI.V34I05.6239}.
\newblock URL \url{https://doi.org/10.1609/aaai.v34i05.6239}.

\bibitem[Chizhov et~al.(2024)Chizhov, Arnett, Korotkova, and
  Yamshchikov]{chizhov2024PickyBPE}
Pavel Chizhov, Catherine Arnett, Elizaveta Korotkova, and Ivan~P. Yamshchikov.
\newblock {BPE} gets picky: Efficient vocabulary refinement during tokenizer
  training.
\newblock In Yaser Al-Onaizan, Mohit Bansal, and Yun-Nung Chen, editors,
  \emph{Proceedings of the 2024 Conference on Empirical Methods in Natural
  Language Processing}, pages 16587--16604, Miami, Florida, USA, November 2024.
  Association for Computational Linguistics.
\newblock \doi{10.18653/v1/2024.emnlp-main.925}.
\newblock URL \url{https://aclanthology.org/2024.emnlp-main.925/}.

\bibitem[Chung et~al.(2023)Chung, Garcia, Roberts, Tay, Firat, Narang, and
  Constant]{chung2023unimax}
Hyung~Won Chung, Xavier Garcia, Adam Roberts, Yi~Tay, Orhan Firat, Sharan
  Narang, and Noah Constant.
\newblock Unimax: Fairer and more effective language sampling for large-scale
  multilingual pretraining.
\newblock In \emph{The Eleventh International Conference on Learning
  Representations}, 2023.
\newblock URL \url{https://openreview.net/forum?id=kXwdL1cWOAi}.

\bibitem[Clark et~al.(2019)Clark, Lee, Chang, Kwiatkowski, Collins, and
  Toutanova]{clark2019boolq}
Christopher Clark, Kenton Lee, Ming-Wei Chang, Tom Kwiatkowski, Michael
  Collins, and Kristina Toutanova.
\newblock {B}ool{Q}: Exploring the surprising difficulty of natural yes/no
  questions.
\newblock In Jill Burstein, Christy Doran, and Thamar Solorio, editors,
  \emph{Proceedings of the 2019 Conference of the North {A}merican Chapter of
  the Association for Computational Linguistics: Human Language Technologies,
  Volume 1 (Long and Short Papers)}, pages 2924--2936, Minneapolis, Minnesota,
  June 2019. Association for Computational Linguistics.
\newblock \doi{10.18653/v1/N19-1300}.
\newblock URL \url{https://aclanthology.org/N19-1300/}.

\bibitem[Clark et~al.(2018)Clark, Cowhey, Etzioni, Khot, Sabharwal, Schoenick,
  and Tafjord]{clark2018arc}
Peter Clark, Isaac Cowhey, Oren Etzioni, Tushar Khot, Ashish Sabharwal, Carissa
  Schoenick, and Oyvind Tafjord.
\newblock Think you have solved question answering? try arc, the {AI2}
  reasoning challenge.
\newblock \emph{CoRR}, abs/1803.05457, 2018.
\newblock URL \url{http://arxiv.org/abs/1803.05457}.

\bibitem[Conneau et~al.(2018)Conneau, Rinott, Lample, Williams, Bowman,
  Schwenk, and Stoyanov]{conneau2018xnli}
Alexis Conneau, Ruty Rinott, Guillaume Lample, Adina Williams, Samuel Bowman,
  Holger Schwenk, and Veselin Stoyanov.
\newblock {XNLI}: Evaluating cross-lingual sentence representations.
\newblock In Ellen Riloff, David Chiang, Julia Hockenmaier, and Jun{'}ichi
  Tsujii, editors, \emph{Proceedings of the 2018 Conference on Empirical
  Methods in Natural Language Processing}, pages 2475--2485, Brussels, Belgium,
  October-November 2018. Association for Computational Linguistics.
\newblock \doi{10.18653/v1/D18-1269}.
\newblock URL \url{https://aclanthology.org/D18-1269/}.

\bibitem[Dauncey and Wattenhofer(2026)]{dauncey2026tokenizationRL}
Sam Dauncey and Roger Wattenhofer.
\newblock You can learn tokenization end-to-end with reinforcement learning,
  2026.
\newblock URL \url{https://arxiv.org/abs/2602.13940}.

\bibitem[DeepSeek-AI et~al.(2025)DeepSeek-AI, Liu, Feng, Xue, Wang, Wu, Lu,
  Zhao, Deng, Zhang, Ruan, Dai, Guo, Yang, Chen, Ji, Li, Lin, Dai, Luo, Hao,
  Chen, Li, Zhang, Bao, Xu, Wang, Zhang, Ding, Xin, Gao, Li, Qu, Cai, Liang,
  Guo, Ni, Li, Wang, Chen, Chen, Yuan, Qiu, Li, Song, Dong, Hu, Gao, Guan,
  Huang, Yu, Wang, Zhang, Xu, Xia, Zhao, Wang, Zhang, Li, Wang, Zhang, Zhang,
  Tang, Li, Tian, Huang, Wang, Zhang, Wang, Zhu, Chen, Du, Chen, Jin, Ge,
  Zhang, Pan, Wang, Xu, Zhang, Chen, Li, Lu, Zhou, Chen, Wu, Ye, Ye, Ma, Wang,
  Zhou, Yu, Zhou, Pan, Wang, Yun, Pei, Sun, Xiao, Zeng, Zhao, An, Liu, Liang,
  Gao, Yu, Zhang, Li, Jin, Wang, Bi, Liu, Wang, Shen, Chen, Zhang, Chen, Nie,
  Sun, Wang, Cheng, Liu, Xie, Liu, Yu, Song, Shan, Zhou, Yang, Li, Su, Lin, Li,
  Wang, Wei, Zhu, Zhang, Xu, Xu, Huang, Li, Zhao, Sun, Li, Wang, Yu, Zheng,
  Zhang, Shi, Xiong, He, Tang, Piao, Wang, Tan, Ma, Liu, Guo, Wu, Ou, Zhu,
  Wang, Gong, Zou, He, Zha, Xiong, Ma, Yan, Luo, You, Liu, Zhou, Wu, Ren, Ren,
  Sha, Fu, Xu, Huang, Zhang, Xie, Zhang, Hao, Gou, Ma, Yan, Shao, Xu, Wu,
  Zhang, Li, Gu, Zhu, Liu, Li, Xie, Song, Gao, and Pan]{deepseek2025v3}
DeepSeek-AI, Aixin Liu, Bei Feng, Bing Xue, Bingxuan Wang, Bochao Wu, Chengda
  Lu, Chenggang Zhao, Chengqi Deng, Chenyu Zhang, Chong Ruan, Damai Dai, Daya
  Guo, Dejian Yang, Deli Chen, Dongjie Ji, Erhang Li, Fangyun Lin, Fucong Dai,
  Fuli Luo, Guangbo Hao, Guanting Chen, Guowei Li, H.~Zhang, Han Bao, Hanwei
  Xu, Haocheng Wang, Haowei Zhang, Honghui Ding, Huajian Xin, Huazuo Gao, Hui
  Li, Hui Qu, J.~L. Cai, Jian Liang, Jianzhong Guo, Jiaqi Ni, Jiashi Li, Jiawei
  Wang, Jin Chen, Jingchang Chen, Jingyang Yuan, Junjie Qiu, Junlong Li,
  Junxiao Song, Kai Dong, Kai Hu, Kaige Gao, Kang Guan, Kexin Huang, Kuai Yu,
  Lean Wang, Lecong Zhang, Lei Xu, Leyi Xia, Liang Zhao, Litong Wang, Liyue
  Zhang, Meng Li, Miaojun Wang, Mingchuan Zhang, Minghua Zhang, Minghui Tang,
  Mingming Li, Ning Tian, Panpan Huang, Peiyi Wang, Peng Zhang, Qiancheng Wang,
  Qihao Zhu, Qinyu Chen, Qiushi Du, R.~J. Chen, R.~L. Jin, Ruiqi Ge, Ruisong
  Zhang, Ruizhe Pan, Runji Wang, Runxin Xu, Ruoyu Zhang, Ruyi Chen, S.~S. Li,
  Shanghao Lu, Shangyan Zhou, Shanhuang Chen, Shaoqing Wu, Shengfeng Ye,
  Shengfeng Ye, Shirong Ma, Shiyu Wang, Shuang Zhou, Shuiping Yu, Shunfeng
  Zhou, Shuting Pan, T.~Wang, Tao Yun, Tian Pei, Tianyu Sun, W.~L. Xiao,
  Wangding Zeng, Wanjia Zhao, Wei An, Wen Liu, Wenfeng Liang, Wenjun Gao,
  Wenqin Yu, Wentao Zhang, X.~Q. Li, Xiangyue Jin, Xianzu Wang, Xiao Bi,
  Xiaodong Liu, Xiaohan Wang, Xiaojin Shen, Xiaokang Chen, Xiaokang Zhang,
  Xiaosha Chen, Xiaotao Nie, Xiaowen Sun, Xiaoxiang Wang, Xin Cheng, Xin Liu,
  Xin Xie, Xingchao Liu, Xingkai Yu, Xinnan Song, Xinxia Shan, Xinyi Zhou,
  Xinyu Yang, Xinyuan Li, Xuecheng Su, Xuheng Lin, Y.~K. Li, Y.~Q. Wang, Y.~X.
  Wei, Y.~X. Zhu, Yang Zhang, Yanhong Xu, Yanhong Xu, Yanping Huang, Yao Li,
  Yao Zhao, Yaofeng Sun, Yaohui Li, Yaohui Wang, Yi~Yu, Yi~Zheng, Yichao Zhang,
  Yifan Shi, Yiliang Xiong, Ying He, Ying Tang, Yishi Piao, Yisong Wang, Yixuan
  Tan, Yiyang Ma, Yiyuan Liu, Yongqiang Guo, Yu~Wu, Yuan Ou, Yuchen Zhu, Yuduan
  Wang, Yue Gong, Yuheng Zou, Yujia He, Yukun Zha, Yunfan Xiong, Yunxian Ma,
  Yuting Yan, Yuxiang Luo, Yuxiang You, Yuxuan Liu, Yuyang Zhou, Z.~F. Wu,
  Z.~Z. Ren, Zehui Ren, Zhangli Sha, Zhe Fu, Zhean Xu, Zhen Huang, Zhen Zhang,
  Zhenda Xie, Zhengyan Zhang, Zhewen Hao, Zhibin Gou, Zhicheng Ma, Zhigang Yan,
  Zhihong Shao, Zhipeng Xu, Zhiyu Wu, Zhongyu Zhang, Zhuoshu Li, Zihui Gu,
  Zijia Zhu, Zijun Liu, Zilin Li, Ziwei Xie, Ziyang Song, Ziyi Gao, and Zizheng
  Pan.
\newblock Deepseek-v3 technical report, 2025.
\newblock URL \url{https://arxiv.org/abs/2412.19437}.

\bibitem[Devlin et~al.(2019)Devlin, Chang, Lee, and Toutanova]{devlin2019BERT}
Jacob Devlin, Ming-Wei Chang, Kenton Lee, and Kristina Toutanova.
\newblock {BERT}: Pre-training of deep bidirectional transformers for language
  understanding.
\newblock In Jill Burstein, Christy Doran, and Thamar Solorio, editors,
  \emph{Proceedings of the 2019 Conference of the North {A}merican Chapter of
  the Association for Computational Linguistics: Human Language Technologies,
  Volume 1 (Long and Short Papers)}, pages 4171--4186, Minneapolis, Minnesota,
  June 2019. Association for Computational Linguistics.
\newblock \doi{10.18653/v1/N19-1423}.
\newblock URL \url{https://aclanthology.org/N19-1423/}.

\bibitem[Dobler and de~Melo(2023)]{dobler2023FOCUS}
Konstantin Dobler and Gerard de~Melo.
\newblock {FOCUS}: Effective embedding initialization for monolingual
  specialization of multilingual models.
\newblock In Houda Bouamor, Juan Pino, and Kalika Bali, editors,
  \emph{Proceedings of the 2023 Conference on Empirical Methods in Natural
  Language Processing}, pages 13440--13454, Singapore, December 2023.
  Association for Computational Linguistics.
\newblock \doi{10.18653/v1/2023.emnlp-main.829}.
\newblock URL \url{https://aclanthology.org/2023.emnlp-main.829/}.

\bibitem[Feher et~al.(2025)Feher, Vuli{\'c}, and
  Minixhofer]{feher-etal-2025-retrofitting}
Darius Feher, Ivan Vuli{\'c}, and Benjamin Minixhofer.
\newblock Retrofitting large language models with dynamic tokenization.
\newblock In Wanxiang Che, Joyce Nabende, Ekaterina Shutova, and Mohammad~Taher
  Pilehvar, editors, \emph{Proceedings of the 63rd Annual Meeting of the
  Association for Computational Linguistics (Volume 1: Long Papers)}, pages
  29866--29883, Vienna, Austria, July 2025. Association for Computational
  Linguistics.
\newblock ISBN 979-8-89176-251-0.
\newblock \doi{10.18653/v1/2025.acl-long.1444}.
\newblock URL \url{https://aclanthology.org/2025.acl-long.1444/}.

\bibitem[Foroutan et~al.(2025)Foroutan, Meister, Paul, Niklaus, Ahmadi,
  Bosselut, and Sennrich]{foroutan2025parityBPE}
Negar Foroutan, Clara Meister, Debjit Paul, Joel Niklaus, Sina Ahmadi, Antoine
  Bosselut, and Rico Sennrich.
\newblock Parity-aware byte-pair encoding: Improving cross-lingual fairness in
  tokenization, 2025.
\newblock URL \url{https://arxiv.org/abs/2508.04796}.

\bibitem[Gage(1994)]{gage1994bpe}
Philip Gage.
\newblock A new algorithm for data compression.
\newblock \emph{C Users Journal}, 12\penalty0 (2):\penalty0 23--38, 1994.
\newblock URL
  \url{http://www.pennelynn.com/Documents/CUJ/HTML/94HTML/19940045.HTM}.

\bibitem[Gao et~al.(2023)Gao, Tow, Abbasi, Biderman, Black, DiPofi, Foster,
  Golding, Hsu, Le~Noac'h, Li, McDonell, Muennighoff, Ociepa, Phang, Reynolds,
  Schoelkopf, Skowron, Sutawika, Tang, Thite, Wang, Wang, and
  Zou]{lm-eval-harness}
Leo Gao, Jonathan Tow, Baber Abbasi, Stella Biderman, Sid Black, Anthony
  DiPofi, Charles Foster, Laurence Golding, Jeffrey Hsu, Alain Le~Noac'h,
  Haonan Li, Kyle McDonell, Niklas Muennighoff, Chris Ociepa, Jason Phang,
  Laria Reynolds, Hailey Schoelkopf, Aviya Skowron, Lintang Sutawika, Eric
  Tang, Anish Thite, Ben Wang, Kevin Wang, and Andy Zou.
\newblock A framework for few-shot language model evaluation, 12 2023.
\newblock URL \url{https://zenodo.org/records/10256836}.

\bibitem[Garreth~Lee and Wolf(2024)]{lee2024digitstodecisionsHFBlog}
Leandro von~Werra Garreth~Lee, Guilherme~Penedo and Thomas Wolf.
\newblock From digits to decisions: How tokenization impacts arithmetic in
  llms, 2024.
\newblock URL
  \url{https://huggingface.co/spaces/huggingface/number-tokenization-blog}.

\bibitem[Gee et~al.(2022)Gee, Zugarini, Rigutini, and
  Torroni]{gee2022fastTransfer}
Leonidas Gee, Andrea Zugarini, Leonardo Rigutini, and Paolo Torroni.
\newblock Fast vocabulary transfer for language model compression.
\newblock In Yunyao Li and Angeliki Lazaridou, editors, \emph{Proceedings of
  the 2022 Conference on Empirical Methods in Natural Language Processing:
  Industry Track}, pages 409--416, Abu Dhabi, UAE, December 2022. Association
  for Computational Linguistics.
\newblock \doi{10.18653/v1/2022.emnlp-industry.41}.
\newblock URL \url{https://aclanthology.org/2022.emnlp-industry.41/}.

\bibitem[Goddard and Neto(2025)]{goddard2025tokenSurgeon}
Charles Goddard and Fernando~Fernandes Neto.
\newblock Training-free tokenizer transplantation via orthogonal matching
  pursuit, 2025.
\newblock URL \url{https://arxiv.org/abs/2506.06607}.

\bibitem[Goldman et~al.(2024)Goldman, Caciularu, Eyal, Cao, Szpektor, and
  Tsarfaty]{goldman2024unpacking}
Omer Goldman, Avi Caciularu, Matan Eyal, Kris Cao, Idan Szpektor, and Reut
  Tsarfaty.
\newblock Unpacking tokenization: Evaluating text compression and its
  correlation with model performance.
\newblock In Lun-Wei Ku, Andre Martins, and Vivek Srikumar, editors,
  \emph{Findings of the Association for Computational Linguistics: ACL 2024},
  pages 2274--2286, Bangkok, Thailand, August 2024. Association for
  Computational Linguistics.
\newblock \doi{10.18653/v1/2024.findings-acl.134}.
\newblock URL \url{https://aclanthology.org/2024.findings-acl.134/}.

\bibitem[Grattafiori et~al.(2024)Grattafiori, Dubey, Jauhri, Pandey, Kadian,
  Al-Dahle, Letman, Mathur, Schelten, Vaughan, Yang, Fan, Goyal, Hartshorn,
  Yang, Mitra, Sravankumar, Korenev, Hinsvark, Rao, Zhang, Rodriguez,
  Gregerson, Spataru, Roziere, Biron, Tang, Chern, Caucheteux, Nayak, Bi,
  Marra, McConnell, Keller, Touret, Wu, Wong, Ferrer, Nikolaidis, Allonsius,
  Song, Pintz, Livshits, Wyatt, Esiobu, Choudhary, Mahajan, Garcia-Olano,
  Perino, Hupkes, Lakomkin, AlBadawy, Lobanova, Dinan, Smith, Radenovic,
  Guzm{\'a}n, Zhang, Synnaeve, Lee, Anderson, Thattai, Nail, Mialon, Pang,
  Cucurell, Nguyen, Korevaar, Xu, Touvron, Zarov, Ibarra, Kloumann, Misra,
  Evtimov, Zhang, Copet, Lee, Geffert, Vranes, Park, Mahadeokar, Shah, van~der
  Linde, Billock, Hong, Lee, Fu, Chi, Huang, Liu, Wang, Yu, Bitton, Spisak,
  Park, Rocca, Johnstun, Saxe, Jia, Alwala, Prasad, Upasani, Plawiak, Li,
  Heafield, Stone, El-Arini, Iyer, Malik, Chiu, Bhalla, Lakhotia,
  Rantala-Yeary, van~der Maaten, Chen, Tan, Jenkins, Martin, Madaan, Malo,
  Blecher, Landzaat, de~Oliveira, Muzzi, Pasupuleti, Singh, Paluri, Kardas,
  Tsimpoukelli, Oldham, Rita, Pavlova, Kambadur, Lewis, Si, Singh, Hassan,
  Goyal, Torabi, Bashlykov, Bogoychev, Chatterji, Zhang, Duchenne, {\c C}elebi,
  Alrassy, Zhang, Li, Vasic, Weng, Bhargava, Dubal, Krishnan, Koura, Xu, He,
  Dong, Srinivasan, Ganapathy, Calderer, Cabral, Stojnic, Raileanu, Maheswari,
  Girdhar, Patel, Sauvestre, Polidoro, Sumbaly, Taylor, Silva, Hou, Wang,
  Hosseini, Chennabasappa, Singh, Bell, Kim, Edunov, Nie, Narang, Raparthy,
  Shen, Wan, Bhosale, Zhang, Vandenhende, Batra, Whitman, Sootla, Collot,
  Gururangan, Borodinsky, Herman, Fowler, Sheasha, Georgiou, Scialom,
  Speckbacher, Mihaylov, Xiao, Karn, Goswami, Gupta, Ramanathan, Kerkez,
  Gonguet, Do, Vogeti, Albiero, Petrovic, Chu, Xiong, Fu, Meers, Martinet,
  Wang, Wang, Tan, Xia, Xie, Jia, Wang, Goldschlag, Gaur, Babaei, Wen, Song,
  Zhang, Li, Mao, Coudert, Yan, Chen, Papakipos, Singh, Srivastava, Jain,
  Kelsey, Shajnfeld, Gangidi, Victoria, Goldstand, Menon, Sharma, Boesenberg,
  Baevski, Feinstein, Kallet, Sangani, Teo, Yunus, Lupu, Alvarado, Caples, Gu,
  Ho, Poulton, Ryan, Ramchandani, Dong, Franco, Goyal, Saraf, Chowdhury,
  Gabriel, Bharambe, Eisenman, Yazdan, James, Maurer, Leonhardi, Huang, Loyd,
  Paola, Paranjape, Liu, Wu, Ni, Hancock, Wasti, Spence, Stojkovic, Gamido,
  Montalvo, Parker, Burton, Mejia, Liu, Wang, Kim, Zhou, Hu, Chu, Cai, Tindal,
  Feichtenhofer, Gao, Civin, Beaty, Kreymer, Li, Adkins, Xu, Testuggine, David,
  Parikh, Liskovich, Foss, Wang, Le, Holland, Dowling, Jamil, Montgomery,
  Presani, Hahn, Wood, Le, Brinkman, Arcaute, Dunbar, Smothers, Sun, Kreuk,
  Tian, Kokkinos, Ozgenel, Caggioni, Kanayet, Seide, Florez, Schwarz, Badeer,
  Swee, Halpern, Herman, Sizov, Guangyi, Zhang, Lakshminarayanan, Inan,
  Shojanazeri, Zou, Wang, Zha, Habeeb, Rudolph, Suk, Aspegren, Goldman, Zhan,
  Damlaj, Molybog, Tufanov, Leontiadis, Veliche, Gat, Weissman, Geboski, Kohli,
  Lam, Asher, Gaya, Marcus, Tang, Chan, Zhen, Reizenstein, Teboul, Zhong, Jin,
  Yang, Cummings, Carvill, Shepard, McPhie, Torres, Ginsburg, Wang, Wu, U,
  Saxena, Khandelwal, Zand, Matosich, Veeraraghavan, Michelena, Li, Jagadeesh,
  Huang, Chawla, Huang, Chen, Garg, A, Silva, Bell, Zhang, Guo, Yu, Moshkovich,
  Wehrstedt, Khabsa, Avalani, Bhatt, Mankus, Hasson, Lennie, Reso, Groshev,
  Naumov, Lathi, Keneally, Liu, Seltzer, Valko, Restrepo, Patel, Vyatskov,
  Samvelyan, Clark, Macey, Wang, Hermoso, Metanat, Rastegari, Bansal,
  Santhanam, Parks, White, Bawa, Singhal, Egebo, Usunier, Mehta, Laptev, Dong,
  Cheng, Chernoguz, Hart, Salpekar, Kalinli, Kent, Parekh, Saab, Balaji,
  Rittner, Bontrager, Roux, Dollar, Zvyagina, Ratanchandani, Yuvraj, Liang,
  Alao, Rodriguez, Ayub, Murthy, Nayani, Mitra, Parthasarathy, Li, Hogan,
  Battey, Wang, Howes, Rinott, Mehta, Siby, Bondu, Datta, Chugh, Hunt, Dhillon,
  Sidorov, Pan, Mahajan, Verma, Yamamoto, Ramaswamy, Lindsay, Lindsay, Feng,
  Lin, Zha, Patil, Shankar, Zhang, Zhang, Wang, Agarwal, Sajuyigbe, Chintala,
  Max, Chen, Kehoe, Satterfield, Govindaprasad, Gupta, Deng, Cho, Virk,
  Subramanian, Choudhury, Goldman, Remez, Glaser, Best, Koehler, Robinson, Li,
  Zhang, Matthews, Chou, Shaked, Vontimitta, Ajayi, Montanez, Mohan, Kumar,
  Mangla, Ionescu, Poenaru, Mihailescu, Ivanov, Li, Wang, Jiang, Bouaziz,
  Constable, Tang, Wu, Wang, Wu, Gao, Kleinman, Chen, Hu, Jia, Qi, Li, Zhang,
  Zhang, Adi, Nam, Yu, Wang, Zhao, Hao, Qian, Li, He, Rait, DeVito, Rosnbrick,
  Wen, Yang, Zhao, and Ma]{grattafiori2024llama3}
Aaron Grattafiori, Abhimanyu Dubey, Abhinav Jauhri, Abhinav Pandey, Abhishek
  Kadian, Ahmad Al-Dahle, Aiesha Letman, Akhil Mathur, Alan Schelten, Alex
  Vaughan, Amy Yang, Angela Fan, Anirudh Goyal, Anthony Hartshorn, Aobo Yang,
  Archi Mitra, Archie Sravankumar, Artem Korenev, Arthur Hinsvark, Arun Rao,
  Aston Zhang, Aurelien Rodriguez, Austen Gregerson, Ava Spataru, Baptiste
  Roziere, Bethany Biron, Binh Tang, Bobbie Chern, Charlotte Caucheteux, Chaya
  Nayak, Chloe Bi, Chris Marra, Chris McConnell, Christian Keller, Christophe
  Touret, Chunyang Wu, Corinne Wong, Cristian~Canton Ferrer, Cyrus Nikolaidis,
  Damien Allonsius, Daniel Song, Danielle Pintz, Danny Livshits, Danny Wyatt,
  David Esiobu, Dhruv Choudhary, Dhruv Mahajan, Diego Garcia-Olano, Diego
  Perino, Dieuwke Hupkes, Egor Lakomkin, Ehab AlBadawy, Elina Lobanova, Emily
  Dinan, Eric~Michael Smith, Filip Radenovic, Francisco Guzm{\'a}n, Frank
  Zhang, Gabriel Synnaeve, Gabrielle Lee, Georgia~Lewis Anderson, Govind
  Thattai, Graeme Nail, Gregoire Mialon, Guan Pang, Guillem Cucurell, Hailey
  Nguyen, Hannah Korevaar, Hu~Xu, Hugo Touvron, Iliyan Zarov, Imanol~Arrieta
  Ibarra, Isabel Kloumann, Ishan Misra, Ivan Evtimov, Jack Zhang, Jade Copet,
  Jaewon Lee, Jan Geffert, Jana Vranes, Jason Park, Jay Mahadeokar, Jeet Shah,
  Jelmer van~der Linde, Jennifer Billock, Jenny Hong, Jenya Lee, Jeremy Fu,
  Jianfeng Chi, Jianyu Huang, Jiawen Liu, Jie Wang, Jiecao Yu, Joanna Bitton,
  Joe Spisak, Jongsoo Park, Joseph Rocca, Joshua Johnstun, Joshua Saxe, Junteng
  Jia, Kalyan~Vasuden Alwala, Karthik Prasad, Kartikeya Upasani, Kate Plawiak,
  Ke~Li, Kenneth Heafield, Kevin Stone, Khalid El-Arini, Krithika Iyer, Kshitiz
  Malik, Kuenley Chiu, Kunal Bhalla, Kushal Lakhotia, Lauren Rantala-Yeary,
  Laurens van~der Maaten, Lawrence Chen, Liang Tan, Liz Jenkins, Louis Martin,
  Lovish Madaan, Lubo Malo, Lukas Blecher, Lukas Landzaat, Luke de~Oliveira,
  Madeline Muzzi, Mahesh Pasupuleti, Mannat Singh, Manohar Paluri, Marcin
  Kardas, Maria Tsimpoukelli, Mathew Oldham, Mathieu Rita, Maya Pavlova,
  Melanie Kambadur, Mike Lewis, Min Si, Mitesh~Kumar Singh, Mona Hassan, Naman
  Goyal, Narjes Torabi, Nikolay Bashlykov, Nikolay Bogoychev, Niladri
  Chatterji, Ning Zhang, Olivier Duchenne, Onur {\c C}elebi, Patrick Alrassy,
  Pengchuan Zhang, Pengwei Li, Petar Vasic, Peter Weng, Prajjwal Bhargava,
  Pratik Dubal, Praveen Krishnan, Punit~Singh Koura, Puxin Xu, Qing He,
  Qingxiao Dong, Ragavan Srinivasan, Raj Ganapathy, Ramon Calderer,
  Ricardo~Silveira Cabral, Robert Stojnic, Roberta Raileanu, Rohan Maheswari,
  Rohit Girdhar, Rohit Patel, Romain Sauvestre, Ronnie Polidoro, Roshan
  Sumbaly, Ross Taylor, Ruan Silva, Rui Hou, Rui Wang, Saghar Hosseini, Sahana
  Chennabasappa, Sanjay Singh, Sean Bell, Seohyun~Sonia Kim, Sergey Edunov,
  Shaoliang Nie, Sharan Narang, Sharath Raparthy, Sheng Shen, Shengye Wan,
  Shruti Bhosale, Shun Zhang, Simon Vandenhende, Soumya Batra, Spencer Whitman,
  Sten Sootla, Stephane Collot, Suchin Gururangan, Sydney Borodinsky, Tamar
  Herman, Tara Fowler, Tarek Sheasha, Thomas Georgiou, Thomas Scialom, Tobias
  Speckbacher, Todor Mihaylov, Tong Xiao, Ujjwal Karn, Vedanuj Goswami, Vibhor
  Gupta, Vignesh Ramanathan, Viktor Kerkez, Vincent Gonguet, Virginie Do, Vish
  Vogeti, V{\'\i}tor Albiero, Vladan Petrovic, Weiwei Chu, Wenhan Xiong, Wenyin
  Fu, Whitney Meers, Xavier Martinet, Xiaodong Wang, Xiaofang Wang,
  Xiaoqing~Ellen Tan, Xide Xia, Xinfeng Xie, Xuchao Jia, Xuewei Wang, Yaelle
  Goldschlag, Yashesh Gaur, Yasmine Babaei, Yi~Wen, Yiwen Song, Yuchen Zhang,
  Yue Li, Yuning Mao, Zacharie~Delpierre Coudert, Zheng Yan, Zhengxing Chen,
  Zoe Papakipos, Aaditya Singh, Aayushi Srivastava, Abha Jain, Adam Kelsey,
  Adam Shajnfeld, Adithya Gangidi, Adolfo Victoria, Ahuva Goldstand, Ajay
  Menon, Ajay Sharma, Alex Boesenberg, Alexei Baevski, Allie Feinstein, Amanda
  Kallet, Amit Sangani, Amos Teo, Anam Yunus, Andrei Lupu, Andres Alvarado,
  Andrew Caples, Andrew Gu, Andrew Ho, Andrew Poulton, Andrew Ryan, Ankit
  Ramchandani, Annie Dong, Annie Franco, Anuj Goyal, Aparajita Saraf,
  Arkabandhu Chowdhury, Ashley Gabriel, Ashwin Bharambe, Assaf Eisenman, Azadeh
  Yazdan, Beau James, Ben Maurer, Benjamin Leonhardi, Bernie Huang, Beth Loyd,
  Beto~De Paola, Bhargavi Paranjape, Bing Liu, Bo~Wu, Boyu Ni, Braden Hancock,
  Bram Wasti, Brandon Spence, Brani Stojkovic, Brian Gamido, Britt Montalvo,
  Carl Parker, Carly Burton, Catalina Mejia, Ce~Liu, Changhan Wang, Changkyu
  Kim, Chao Zhou, Chester Hu, Ching-Hsiang Chu, Chris Cai, Chris Tindal,
  Christoph Feichtenhofer, Cynthia Gao, Damon Civin, Dana Beaty, Daniel
  Kreymer, Daniel Li, David Adkins, David Xu, Davide Testuggine, Delia David,
  Devi Parikh, Diana Liskovich, Didem Foss, Dingkang Wang, Duc Le, Dustin
  Holland, Edward Dowling, Eissa Jamil, Elaine Montgomery, Eleonora Presani,
  Emily Hahn, Emily Wood, Eric-Tuan Le, Erik Brinkman, Esteban Arcaute, Evan
  Dunbar, Evan Smothers, Fei Sun, Felix Kreuk, Feng Tian, Filippos Kokkinos,
  Firat Ozgenel, Francesco Caggioni, Frank Kanayet, Frank Seide,
  Gabriela~Medina Florez, Gabriella Schwarz, Gada Badeer, Georgia Swee, Gil
  Halpern, Grant Herman, Grigory Sizov, Guangyi, Zhang, Guna Lakshminarayanan,
  Hakan Inan, Hamid Shojanazeri, Han Zou, Hannah Wang, Hanwen Zha, Haroun
  Habeeb, Harrison Rudolph, Helen Suk, Henry Aspegren, Hunter Goldman, Hongyuan
  Zhan, Ibrahim Damlaj, Igor Molybog, Igor Tufanov, Ilias Leontiadis,
  Irina-Elena Veliche, Itai Gat, Jake Weissman, James Geboski, James Kohli,
  Janice Lam, Japhet Asher, Jean-Baptiste Gaya, Jeff Marcus, Jeff Tang,
  Jennifer Chan, Jenny Zhen, Jeremy Reizenstein, Jeremy Teboul, Jessica Zhong,
  Jian Jin, Jingyi Yang, Joe Cummings, Jon Carvill, Jon Shepard, Jonathan
  McPhie, Jonathan Torres, Josh Ginsburg, Junjie Wang, Kai Wu, Kam~Hou U, Karan
  Saxena, Kartikay Khandelwal, Katayoun Zand, Kathy Matosich, Kaushik
  Veeraraghavan, Kelly Michelena, Keqian Li, Kiran Jagadeesh, Kun Huang, Kunal
  Chawla, Kyle Huang, Lailin Chen, Lakshya Garg, Lavender A, Leandro Silva, Lee
  Bell, Lei Zhang, Liangpeng Guo, Licheng Yu, Liron Moshkovich, Luca Wehrstedt,
  Madian Khabsa, Manav Avalani, Manish Bhatt, Martynas Mankus, Matan Hasson,
  Matthew Lennie, Matthias Reso, Maxim Groshev, Maxim Naumov, Maya Lathi,
  Meghan Keneally, Miao Liu, Michael~L. Seltzer, Michal Valko, Michelle
  Restrepo, Mihir Patel, Mik Vyatskov, Mikayel Samvelyan, Mike Clark, Mike
  Macey, Mike Wang, Miquel~Jubert Hermoso, Mo~Metanat, Mohammad Rastegari,
  Munish Bansal, Nandhini Santhanam, Natascha Parks, Natasha White, Navyata
  Bawa, Nayan Singhal, Nick Egebo, Nicolas Usunier, Nikhil Mehta,
  Nikolay~Pavlovich Laptev, Ning Dong, Norman Cheng, Oleg Chernoguz, Olivia
  Hart, Omkar Salpekar, Ozlem Kalinli, Parkin Kent, Parth Parekh, Paul Saab,
  Pavan Balaji, Pedro Rittner, Philip Bontrager, Pierre Roux, Piotr Dollar,
  Polina Zvyagina, Prashant Ratanchandani, Pritish Yuvraj, Qian Liang, Rachad
  Alao, Rachel Rodriguez, Rafi Ayub, Raghotham Murthy, Raghu Nayani, Rahul
  Mitra, Rangaprabhu Parthasarathy, Raymond Li, Rebekkah Hogan, Robin Battey,
  Rocky Wang, Russ Howes, Ruty Rinott, Sachin Mehta, Sachin Siby, Sai~Jayesh
  Bondu, Samyak Datta, Sara Chugh, Sara Hunt, Sargun Dhillon, Sasha Sidorov,
  Satadru Pan, Saurabh Mahajan, Saurabh Verma, Seiji Yamamoto, Sharadh
  Ramaswamy, Shaun Lindsay, Shaun Lindsay, Sheng Feng, Shenghao Lin,
  Shengxin~Cindy Zha, Shishir Patil, Shiva Shankar, Shuqiang Zhang, Shuqiang
  Zhang, Sinong Wang, Sneha Agarwal, Soji Sajuyigbe, Soumith Chintala,
  Stephanie Max, Stephen Chen, Steve Kehoe, Steve Satterfield, Sudarshan
  Govindaprasad, Sumit Gupta, Summer Deng, Sungmin Cho, Sunny Virk, Suraj
  Subramanian, Sy~Choudhury, Sydney Goldman, Tal Remez, Tamar Glaser, Tamara
  Best, Thilo Koehler, Thomas Robinson, Tianhe Li, Tianjun Zhang, Tim Matthews,
  Timothy Chou, Tzook Shaked, Varun Vontimitta, Victoria Ajayi, Victoria
  Montanez, Vijai Mohan, Vinay~Satish Kumar, Vishal Mangla, Vlad Ionescu, Vlad
  Poenaru, Vlad~Tiberiu Mihailescu, Vladimir Ivanov, Wei Li, Wenchen Wang,
  Wenwen Jiang, Wes Bouaziz, Will Constable, Xiaocheng Tang, Xiaojian Wu,
  Xiaolan Wang, Xilun Wu, Xinbo Gao, Yaniv Kleinman, Yanjun Chen, Ye~Hu,
  Ye~Jia, Ye~Qi, Yenda Li, Yilin Zhang, Ying Zhang, Yossi Adi, Youngjin Nam,
  Yu, Wang, Yu~Zhao, Yuchen Hao, Yundi Qian, Yunlu Li, Yuzi He, Zach Rait,
  Zachary DeVito, Zef Rosnbrick, Zhaoduo Wen, Zhenyu Yang, Zhiwei Zhao, and
  Zhiyu Ma.
\newblock The llama 3 herd of models, 2024.
\newblock URL \url{https://arxiv.org/abs/2407.21783}.

\bibitem[Honnibal et~al.(2020)Honnibal, Montani, Van~Landeghem, and
  Boyd]{honnibal2020spacy}
Matthew Honnibal, Ines Montani, Sofie Van~Landeghem, and Adriane Boyd.
\newblock {spaCy: Industrial-strength Natural Language Processing in Python}.
\newblock 2020.
\newblock \doi{10.5281/zenodo.1212303}.

\bibitem[Hwang et~al.(2025)Hwang, Wang, and Gu]{hwang2025hnet}
Sukjun Hwang, Brandon Wang, and Albert Gu.
\newblock Dynamic chunking for end-to-end hierarchical sequence modeling, 2025.
\newblock URL \url{https://arxiv.org/abs/2507.07955}.

\bibitem[Jabbar(2024)]{jabbar2024morphpiece}
Haris Jabbar.
\newblock Morphpiece : A linguistic tokenizer for large language models, 2024.
\newblock URL \url{https://arxiv.org/abs/2307.07262}.

\bibitem[Joulin et~al.(2016)Joulin, Grave, Bojanowski, Douze, J{\'e}gou, and
  Mikolov]{joulin2016fasttext}
Armand Joulin, Edouard Grave, Piotr Bojanowski, Matthijs Douze, H{\'e}rve
  J{\'e}gou, and Tomas Mikolov.
\newblock Fasttext.zip: Compressing text classification models.
\newblock \emph{arXiv preprint arXiv:1612.03651}, 2016.

\bibitem[Joulin et~al.(2017)Joulin, Grave, Bojanowski, and
  Mikolov]{joulin2017bag}
Armand Joulin, Edouard Grave, Piotr Bojanowski, and Tomas Mikolov.
\newblock Bag of tricks for efficient text classification.
\newblock In Mirella Lapata, Phil Blunsom, and Alexander Koller, editors,
  \emph{Proceedings of the 15th Conference of the {E}uropean Chapter of the
  Association for Computational Linguistics: Volume 2, Short Papers}, pages
  427--431, Valencia, Spain, April 2017. Association for Computational
  Linguistics.
\newblock URL \url{https://aclanthology.org/E17-2068/}.

\bibitem[Jumelet et~al.(2026)Jumelet, Weissweiler, Nivre, and
  Bisazza]{jumelet2026multiblimp}
Jaap Jumelet, Leonie Weissweiler, Joakim Nivre, and Arianna Bisazza.
\newblock {M}ulti{BL}i{MP} 1.0: A massively multilingual benchmark of
  linguistic minimal pairs.
\newblock \emph{Transactions of the Association for Computational Linguistics},
  14:\penalty0 193--216, 2026.
\newblock \doi{10.1162/tacl.a.600}.
\newblock URL \url{https://aclanthology.org/2026.tacl-1.10/}.

\bibitem[Kandpal et~al.(2026)Kandpal, Lester, Raffel, Majstorovic, Biderman,
  Abbasi, Soldaini, Shippole, Cooper, Skowron, Longpre, Sutawika, Albalak, Xu,
  Penedo, allal, Bakouch, Pressman, Fan, Stander, Song, Gokaslan, Kirchenbauer,
  Goldstein, Bartoldson, Kailkhura, and Murray]{kandpal2026comma}
Nikhil Kandpal, Brian Lester, Colin Raffel, Sebastian Majstorovic, Stella
  Biderman, Baber Abbasi, Luca Soldaini, Enrico Shippole, A.~Feder Cooper,
  Aviya Skowron, Shayne Longpre, Lintang Sutawika, Alon Albalak, Zhenlin Xu,
  Guilherme Penedo, Loubna~Ben allal, Elie Bakouch, John~David Pressman, Honglu
  Fan, Dashiell Stander, Guangyu Song, Aaron Gokaslan, John Kirchenbauer, Tom
  Goldstein, Brian~R. Bartoldson, Bhavya Kailkhura, and Tyler Murray.
\newblock The common pile v0.1: An 8{TB} dataset of public domain and openly
  licensed text.
\newblock In \emph{The Thirty-ninth Annual Conference on Neural Information
  Processing Systems Datasets and Benchmarks Track}, 2026.
\newblock URL \url{https://openreview.net/forum?id=DIELgiqdvJ}.

\bibitem[Kargaran et~al.(2023)Kargaran, Imani, Yvon, and
  Schuetze]{kargaran2023glotlid}
Amir~Hossein Kargaran, Ayyoob Imani, Fran{\c{c}}ois Yvon, and Hinrich Schuetze.
\newblock {G}lot{LID}: Language identification for low-resource languages.
\newblock In Houda Bouamor, Juan Pino, and Kalika Bali, editors, \emph{Findings
  of the Association for Computational Linguistics: EMNLP 2023}, pages
  6155--6218, Singapore, December 2023. Association for Computational
  Linguistics.
\newblock \doi{10.18653/v1/2023.findings-emnlp.410}.
\newblock URL \url{https://aclanthology.org/2023.findings-emnlp.410/}.

\bibitem[Kudo(2018)]{kudo2018subword}
Taku Kudo.
\newblock Subword regularization: Improving neural network translation models
  with multiple subword candidates.
\newblock In Iryna Gurevych and Yusuke Miyao, editors, \emph{Proceedings of the
  56th Annual Meeting of the Association for Computational Linguistics (Volume
  1: Long Papers)}, pages 66--75, Melbourne, Australia, July 2018. Association
  for Computational Linguistics.
\newblock \doi{10.18653/v1/P18-1007}.
\newblock URL \url{https://aclanthology.org/P18-1007/}.

\bibitem[Kudo and Richardson(2018)]{kudo2018sentencepiece}
Taku Kudo and John Richardson.
\newblock {S}entence{P}iece: A simple and language independent subword
  tokenizer and detokenizer for neural text processing.
\newblock In Eduardo Blanco and Wei Lu, editors, \emph{Proceedings of the 2018
  Conference on Empirical Methods in Natural Language Processing: System
  Demonstrations}, pages 66--71, Brussels, Belgium, November 2018. Association
  for Computational Linguistics.
\newblock \doi{10.18653/v1/D18-2012}.
\newblock URL \url{https://aclanthology.org/D18-2012/}.

\bibitem[Lai et~al.(2023)Lai, Nguyen, Ngo, Nguyen, Dernoncourt, Rossi, and
  Nguyen]{lai2023okapi}
Viet Lai, Chien Nguyen, Nghia Ngo, Thuat Nguyen, Franck Dernoncourt, Ryan
  Rossi, and Thien Nguyen.
\newblock Okapi: Instruction-tuned large language models in multiple languages
  with reinforcement learning from human feedback.
\newblock In Yansong Feng and Els Lefever, editors, \emph{Proceedings of the
  2023 Conference on Empirical Methods in Natural Language Processing: System
  Demonstrations}, pages 318--327, Singapore, December 2023. Association for
  Computational Linguistics.
\newblock \doi{10.18653/v1/2023.emnlp-demo.28}.
\newblock URL \url{https://aclanthology.org/2023.emnlp-demo.28/}.

\bibitem[Lan et~al.(2020)Lan, Chen, Goodman, Gimpel, Sharma, and
  Soricut]{lan2020albert}
Zhenzhong Lan, Mingda Chen, Sebastian Goodman, Kevin Gimpel, Piyush Sharma, and
  Radu Soricut.
\newblock Albert: A lite bert for self-supervised learning of language
  representations, 2020.
\newblock URL \url{https://arxiv.org/abs/1909.11942}.

\bibitem[Land and Arnett(2025)]{land2025scriptBPE}
Sander Land and Catherine Arnett.
\newblock {BPE} stays on {SCRIPT}: Structured encoding for robust multilingual
  pretokenization.
\newblock In \emph{Tokenization Workshop}, 2025.

\bibitem[Land and Bartolo(2024)]{land2024magikarp}
Sander Land and Max Bartolo.
\newblock Fishing for magikarp: Automatically detecting under-trained tokens in
  large language models.
\newblock In Yaser Al-Onaizan, Mohit Bansal, and Yun-Nung Chen, editors,
  \emph{Proceedings of the 2024 Conference on Empirical Methods in Natural
  Language Processing}, pages 11631--11646, Miami, Florida, USA, November 2024.
  Association for Computational Linguistics.
\newblock \doi{10.18653/v1/2024.emnlp-main.649}.
\newblock URL \url{https://aclanthology.org/2024.emnlp-main.649/}.

\bibitem[Li et~al.(2024)Li, Liu, Deng, Zhang, Song, Shi, Wang, Li, Liu, and
  Wang]{li2024glitch}
Yuxi Li, Yi~Liu, Gelei Deng, Ying Zhang, Wenjia Song, Ling Shi, Kailong Wang,
  Yuekang Li, Yang Liu, and Haoyu Wang.
\newblock Glitch tokens in large language models: Categorization taxonomy and
  effective detection.
\newblock \emph{Proc. ACM Softw. Eng.}, 1\penalty0 (FSE), July 2024.
\newblock \doi{10.1145/3660799}.
\newblock URL \url{https://doi.org/10.1145/3660799}.

\bibitem[Lian et~al.(2024)Lian, Xiong, Niu, Mo, Su, Lin, Chen, Liu, Han, and
  Ding]{lian2024scaffoldbpe}
Haoran Lian, Yizhe Xiong, Jianwei Niu, Shasha Mo, Zhenpeng Su, Zijia Lin, Hui
  Chen, Peng Liu, Jungong Han, and Guiguang Ding.
\newblock Scaffold-bpe: Enhancing byte pair encoding for large language models
  with simple and effective scaffold token removal, 2024.
\newblock URL \url{https://arxiv.org/abs/2404.17808}.

\bibitem[Lin et~al.(2022)Lin, Mihaylov, Artetxe, Wang, Chen, Simig, Ott, Goyal,
  Bhosale, Du, Pasunuru, Shleifer, Koura, Chaudhary, O{'}Horo, Wang,
  Zettlemoyer, Kozareva, Diab, Stoyanov, and Li]{lin2022XGLM-XStoryClose}
Xi~Victoria Lin, Todor Mihaylov, Mikel Artetxe, Tianlu Wang, Shuohui Chen,
  Daniel Simig, Myle Ott, Naman Goyal, Shruti Bhosale, Jingfei Du, Ramakanth
  Pasunuru, Sam Shleifer, Punit~Singh Koura, Vishrav Chaudhary, Brian O{'}Horo,
  Jeff Wang, Luke Zettlemoyer, Zornitsa Kozareva, Mona Diab, Veselin Stoyanov,
  and Xian Li.
\newblock Few-shot learning with multilingual generative language models.
\newblock In Yoav Goldberg, Zornitsa Kozareva, and Yue Zhang, editors,
  \emph{Proceedings of the 2022 Conference on Empirical Methods in Natural
  Language Processing}, pages 9019--9052, Abu Dhabi, United Arab Emirates,
  December 2022. Association for Computational Linguistics.
\newblock \doi{10.18653/v1/2022.emnlp-main.616}.
\newblock URL \url{https://aclanthology.org/2022.emnlp-main.616/}.

\bibitem[Liu et~al.(2025)Liu, Hayase, Hofmann, Oh, Smith, and
  Choi]{liu2025superbpe}
Alisa Liu, Jonathan Hayase, Valentin Hofmann, Sewoong Oh, Noah~A. Smith, and
  Yejin Choi.
\newblock Super{BPE}: Space travel for language models.
\newblock In \emph{Second Conference on Language Modeling}, 2025.
\newblock URL \url{https://openreview.net/forum?id=lcDRvffeNP}.

\bibitem[Liu et~al.(2024)Liu, Lin, Wang, and Schuetze]{liu2024OFA}
Yihong Liu, Peiqin Lin, Mingyang Wang, and Hinrich Schuetze.
\newblock {OFA}: A framework of initializing unseen subword embeddings for
  efficient large-scale multilingual continued pretraining.
\newblock In Kevin Duh, Helena Gomez, and Steven Bethard, editors,
  \emph{Findings of the Association for Computational Linguistics: NAACL 2024},
  pages 1067--1097, Mexico City, Mexico, June 2024. Association for
  Computational Linguistics.
\newblock \doi{10.18653/v1/2024.findings-naacl.68}.
\newblock URL \url{https://aclanthology.org/2024.findings-naacl.68/}.

\bibitem[Loshchilov and Hutter(2019)]{loshchilov2018AdamW}
Ilya Loshchilov and Frank Hutter.
\newblock Decoupled weight decay regularization.
\newblock In \emph{International Conference on Learning Representations}, 2019.
\newblock URL \url{https://openreview.net/forum?id=Bkg6RiCqY7}.

\bibitem[Lozhkov et~al.(2024)Lozhkov, Ben~Allal, von Werra, and
  Wolf]{lozhkov2024fineweb-edu}
Anton Lozhkov, Loubna Ben~Allal, Leandro von Werra, and Thomas Wolf.
\newblock Fineweb-edu: the finest collection of educational content, 2024.
\newblock URL \url{https://huggingface.co/datasets/HuggingFaceFW/fineweb-edu}.

\bibitem[Meister(2026)]{meisterTokEvalTokenizerAnalysis2026}
Clara Meister.
\newblock Tokeval: A tokenizer analysis suite.
\newblock In \emph{Third Conference on Language Modeling}, 2026.
\newblock URL \url{https://openreview.net/forum?id=RpcJS02Wcb}.

\bibitem[Meister et~al.(2026{\natexlab{a}})Meister, Altıntaş, and
  Bosselut]{meister2026languagespecificeffectstokenizerchoice}
Clara Meister, Gül~Sena Altıntaş, and Antoine Bosselut.
\newblock Language-specific effects of tokenizer choice in multilingual
  language models, 2026{\natexlab{a}}.
\newblock URL \url{https://arxiv.org/abs/2610.12144}.

\bibitem[Meister et~al.(2026{\natexlab{b}})Meister, Yavuz, Lesci, and
  Pimentel]{meister2026LID}
Clara Meister, Ahmetcan Yavuz, Pietro Lesci, and Tiago Pimentel.
\newblock What language is this? ask your tokenizer, 2026{\natexlab{b}}.
\newblock URL \url{https://arxiv.org/abs/2602.17655}.

\bibitem[Messmer et~al.(2025)Messmer, Sabol{\v{c}}ec, and
  Jaggi]{messmer2025fineweb2hq}
Bettina Messmer, Vinko Sabol{\v{c}}ec, and Martin Jaggi.
\newblock Enhancing multilingual {LLM} pretraining with model-based data
  selection.
\newblock In Jonathan Gerber, Mark Cieliebak, Don Tuggener, and Manuela
  H{\"u}rlimann, editors, \emph{Proceedings of the 10th edition of the Swiss
  Text Analytics Conference}, pages 31--56, Winterthur, Switzerland, May 2025.
  Association for Computational Linguistics.
\newblock URL \url{https://aclanthology.org/2025.swisstext-1.4/}.

\bibitem[Mielke et~al.(2021)Mielke, Alyafeai, Salesky, Raffel, Dey,
  Gall{\'{e}}, Raja, Si, Lee, Sagot, and Tan]{mielke2021survey}
Sabrina~J. Mielke, Zaid Alyafeai, Elizabeth Salesky, Colin Raffel, Manan Dey,
  Matthias Gall{\'{e}}, Arun Raja, Chenglei Si, Wilson~Y. Lee, Beno{\^{\i}}t
  Sagot, and Samson Tan.
\newblock Between words and characters: {A} brief history of open-vocabulary
  modeling and tokenization in {NLP}.
\newblock \emph{CoRR}, abs/2112.10508, 2021.
\newblock URL \url{https://arxiv.org/abs/2112.10508}.

\bibitem[Minixhofer et~al.(2022)Minixhofer, Paischer, and
  Rekabsaz]{minixhofer2022wechsel}
Benjamin Minixhofer, Fabian Paischer, and Navid Rekabsaz.
\newblock {WECHSEL}: Effective initialization of subword embeddings for
  cross-lingual transfer of monolingual language models.
\newblock In Marine Carpuat, Marie-Catherine de~Marneffe, and Ivan~Vladimir
  Meza~Ruiz, editors, \emph{Proceedings of the 2022 Conference of the North
  American Chapter of the Association for Computational Linguistics: Human
  Language Technologies}, pages 3992--4006, Seattle, United States, July 2022.
  Association for Computational Linguistics.
\newblock \doi{10.18653/v1/2022.naacl-main.293}.
\newblock URL \url{https://aclanthology.org/2022.naacl-main.293/}.

\bibitem[Minixhofer et~al.(2024)Minixhofer, Ponti, and
  Vuli{\'c}]{minixhofer2024xett}
Benjamin Minixhofer, Edoardo Ponti, and Ivan Vuli{\'c}.
\newblock Zero-shot tokenizer transfer.
\newblock In \emph{The Thirty-eighth Annual Conference on Neural Information
  Processing Systems}, 2024.
\newblock URL \url{https://openreview.net/forum?id=RwBObRsIzC}.

\bibitem[Minixhofer et~al.(2025)Minixhofer, Murray, Limisiewicz, Korhonen,
  Zettlemoyer, Smith, Ponti, Soldaini, and Hofmann]{minixhofer2025BOLMO}
Benjamin Minixhofer, Tyler Murray, Tomasz Limisiewicz, Anna Korhonen, Luke
  Zettlemoyer, Noah~A. Smith, Edoardo~M. Ponti, Luca Soldaini, and Valentin
  Hofmann.
\newblock Bolmo: Byteifying the next generation of language models.
\newblock \emph{CoRR}, abs/2512.15586, 2025.
\newblock \doi{10.48550/ARXIV.2512.15586}.
\newblock URL \url{https://doi.org/10.48550/arXiv.2512.15586}.

\bibitem[Mofijul~Islam et~al.(2022)Mofijul~Islam, Aguilar, Ponnusamy,
  Solomon~Mathialagan, Ma, and Guo]{islam2022vocabularyFree}
Md~Mofijul~Islam, Gustavo Aguilar, Pragaash Ponnusamy, Clint
  Solomon~Mathialagan, Chengyuan Ma, and Chenlei Guo.
\newblock A vocabulary-free multilingual neural tokenizer for end-to-end task
  learning.
\newblock In Spandana Gella, He~He, Bodhisattwa~Prasad Majumder, Burcu Can,
  Eleonora Giunchiglia, Samuel Cahyawijaya, Sewon Min, Maximilian Mozes,
  Xiang~Lorraine Li, Isabelle Augenstein, Anna Rogers, Kyunghyun Cho, Edward
  Grefenstette, Laura Rimell, and Chris Dyer, editors, \emph{Proceedings of the
  7th Workshop on Representation Learning for NLP}, pages 91--99, Dublin,
  Ireland, May 2022. Association for Computational Linguistics.
\newblock \doi{10.18653/v1/2022.repl4nlp-1.10}.
\newblock URL \url{https://aclanthology.org/2022.repl4nlp-1.10/}.

\bibitem[Mohtaj et~al.(2018)Mohtaj, Roshanfekr, Zafarian, and
  Asghari]{mohtaj2018parsivarFarsi}
Salar Mohtaj, Behnam Roshanfekr, Atefeh Zafarian, and Habibollah Asghari.
\newblock {P}arsivar: A language processing toolkit for {P}ersian.
\newblock In Nicoletta Calzolari, Khalid Choukri, Christopher Cieri, Thierry
  Declerck, Sara Goggi, Koiti Hasida, Hitoshi Isahara, Bente Maegaard, Joseph
  Mariani, H{\'e}l{\`e}ne Mazo, Asuncion Moreno, Jan Odijk, Stelios Piperidis,
  and Takenobu Tokunaga, editors, \emph{Proceedings of the Eleventh
  International Conference on Language Resources and Evaluation ({LREC} 2018)},
  Miyazaki, Japan, May 2018. European Language Resources Association (ELRA).
\newblock URL \url{https://aclanthology.org/L18-1179/}.

\bibitem[Muennighoff et~al.(2022)Muennighoff, Wang, Sutawika, Roberts,
  Biderman, Scao, Bari, Shen, Yong, Schoelkopf, Tang, Radev, Aji, Almubarak,
  Albanie, Alyafeai, Webson, Raff, and Raffel]{muennighoff2022crosslingual}
Niklas Muennighoff, Thomas Wang, Lintang Sutawika, Adam Roberts, Stella
  Biderman, Teven~Le Scao, M~Saiful Bari, Sheng Shen, Zheng-Xin Yong, Hailey
  Schoelkopf, Xiangru Tang, Dragomir Radev, Alham~Fikri Aji, Khalid Almubarak,
  Samuel Albanie, Zaid Alyafeai, Albert Webson, Edward Raff, and Colin Raffel.
\newblock Crosslingual generalization through multitask finetuning, 2022.

\bibitem[{no-plagiarism}(2025)]{pymorphy3}
{no-plagiarism}.
\newblock pymorphy3: Morphological analyzer for russian and ukrainian.
\newblock \url{https://github.com/no-plagiarism/pymorphy3}, 2025.
\newblock Continuation of pymorphy2.

\bibitem[num2words()]{num2words}
num2words.
\newblock num2words library - convert numbers to words in multiple languages.

\bibitem[Obeid et~al.(2020)Obeid, Zalmout, Khalifa, Taji, Oudah, Alhafni,
  Inoue, Eryani, Erdmann, and Habash]{obeid2020camelArabic}
Ossama Obeid, Nasser Zalmout, Salam Khalifa, Dima Taji, Mai Oudah, Bashar
  Alhafni, Go~Inoue, Fadhl Eryani, Alexander Erdmann, and Nizar Habash.
\newblock {CAM}e{L} tools: An open source python toolkit for {A}rabic natural
  language processing.
\newblock In Nicoletta Calzolari, Fr{\'e}d{\'e}ric B{\'e}chet, Philippe Blache,
  Khalid Choukri, Christopher Cieri, Thierry Declerck, Sara Goggi, Hitoshi
  Isahara, Bente Maegaard, Joseph Mariani, H{\'e}l{\`e}ne Mazo, Asuncion
  Moreno, Jan Odijk, and Stelios Piperidis, editors, \emph{Proceedings of the
  Twelfth Language Resources and Evaluation Conference}, pages 7022--7032,
  Marseille, France, May 2020. European Language Resources Association.
\newblock ISBN 979-10-95546-34-4.
\newblock URL \url{https://aclanthology.org/2020.lrec-1.868/}.

\bibitem[OpenAI et~al.(2024)OpenAI, Achiam, Adler, Agarwal, Ahmad, Akkaya,
  Aleman, Almeida, Altenschmidt, Altman, Anadkat, Avila, Babuschkin, Balaji,
  Balcom, Baltescu, Bao, Bavarian, Belgum, Bello, Berdine, Bernadett-Shapiro,
  Berner, Bogdonoff, Boiko, Boyd, Brakman, Brockman, Brooks, Brundage, Button,
  Cai, Campbell, Cann, Carey, Carlson, Carmichael, Chan, Chang, Chantzis, Chen,
  Chen, Chen, Chen, Chen, Chess, Cho, Chu, Chung, Cummings, Currier, Dai,
  Decareaux, Degry, Deutsch, Deville, Dhar, Dohan, Dowling, Dunning, Ecoffet,
  Eleti, Eloundou, Farhi, Fedus, Felix, Fishman, Forte, Fulford, Gao, Georges,
  Gibson, Goel, Gogineni, Goh, Gontijo-Lopes, Gordon, Grafstein, Gray, Greene,
  Gross, Gu, Guo, Hallacy, Han, Harris, He, Heaton, Heidecke, Hesse, Hickey,
  Hickey, Hoeschele, Houghton, Hsu, Hu, Hu, Huizinga, Jain, Jain, Jang, Jiang,
  Jiang, Jin, Jin, Jomoto, Jonn, Jun, Kaftan, Kaiser, Kamali, Kanitscheider,
  Keskar, Khan, Kilpatrick, Kim, Kim, Kim, Kirchner, Kiros, Knight, Kokotajlo,
  Kondraciuk, Kondrich, Konstantinidis, Kosic, Krueger, Kuo, Lampe, Lan, Lee,
  Leike, Leung, Levy, Li, Lim, Lin, Lin, Litwin, Lopez, Lowe, Lue, Makanju,
  Malfacini, Manning, Markov, Markovski, Martin, Mayer, Mayne, McGrew,
  McKinney, McLeavey, McMillan, McNeil, Medina, Mehta, Menick, Metz,
  Mishchenko, Mishkin, Monaco, Morikawa, Mossing, Mu, Murati, Murk, M{\'e}ly,
  Nair, Nakano, Nayak, Neelakantan, Ngo, Noh, Ouyang, O'Keefe, Pachocki, Paino,
  Palermo, Pantuliano, Parascandolo, Parish, Parparita, Passos, Pavlov, Peng,
  Perelman, de~Avila Belbute~Peres, Petrov, de~Oliveira~Pinto, Michael,
  Pokorny, Pokrass, Pong, Powell, Power, Power, Proehl, Puri, Radford, Rae,
  Ramesh, Raymond, Real, Rimbach, Ross, Rotsted, Roussez, Ryder, Saltarelli,
  Sanders, Santurkar, Sastry, Schmidt, Schnurr, Schulman, Selsam, Sheppard,
  Sherbakov, Shieh, Shoker, Shyam, Sidor, Sigler, Simens, Sitkin, Slama, Sohl,
  Sokolowsky, Song, Staudacher, Such, Summers, Sutskever, Tang, Tezak,
  Thompson, Tillet, Tootoonchian, Tseng, Tuggle, Turley, Tworek, Uribe,
  Vallone, Vijayvergiya, Voss, Wainwright, Wang, Wang, Wang, Ward, Wei,
  Weinmann, Welihinda, Welinder, Weng, Weng, Wiethoff, Willner, Winter,
  Wolrich, Wong, Workman, Wu, Wu, Wu, Xiao, Xu, Yoo, Yu, Yuan, Zaremba,
  Zellers, Zhang, Zhang, Zhao, Zheng, Zhuang, Zhuk, and
  Zoph]{openai2024gpt4technicalreport}
OpenAI, Josh Achiam, Steven Adler, Sandhini Agarwal, Lama Ahmad, Ilge Akkaya,
  Florencia~Leoni Aleman, Diogo Almeida, Janko Altenschmidt, Sam Altman,
  Shyamal Anadkat, Red Avila, Igor Babuschkin, Suchir Balaji, Valerie Balcom,
  Paul Baltescu, Haiming Bao, Mohammad Bavarian, Jeff Belgum, Irwan Bello, Jake
  Berdine, Gabriel Bernadett-Shapiro, Christopher Berner, Lenny Bogdonoff, Oleg
  Boiko, Madelaine Boyd, Anna-Luisa Brakman, Greg Brockman, Tim Brooks, Miles
  Brundage, Kevin Button, Trevor Cai, Rosie Campbell, Andrew Cann, Brittany
  Carey, Chelsea Carlson, Rory Carmichael, Brooke Chan, Che Chang, Fotis
  Chantzis, Derek Chen, Sully Chen, Ruby Chen, Jason Chen, Mark Chen, Ben
  Chess, Chester Cho, Casey Chu, Hyung~Won Chung, Dave Cummings, Jeremiah
  Currier, Yunxing Dai, Cory Decareaux, Thomas Degry, Noah Deutsch, Damien
  Deville, Arka Dhar, David Dohan, Steve Dowling, Sheila Dunning, Adrien
  Ecoffet, Atty Eleti, Tyna Eloundou, David Farhi, Liam Fedus, Niko Felix,
  Sim{\'o}n~Posada Fishman, Juston Forte, Isabella Fulford, Leo Gao, Elie
  Georges, Christian Gibson, Vik Goel, Tarun Gogineni, Gabriel Goh, Rapha
  Gontijo-Lopes, Jonathan Gordon, Morgan Grafstein, Scott Gray, Ryan Greene,
  Joshua Gross, Shixiang~Shane Gu, Yufei Guo, Chris Hallacy, Jesse Han, Jeff
  Harris, Yuchen He, Mike Heaton, Johannes Heidecke, Chris Hesse, Alan Hickey,
  Wade Hickey, Peter Hoeschele, Brandon Houghton, Kenny Hsu, Shengli Hu, Xin
  Hu, Joost Huizinga, Shantanu Jain, Shawn Jain, Joanne Jang, Angela Jiang,
  Roger Jiang, Haozhun Jin, Denny Jin, Shino Jomoto, Billie Jonn, Heewoo Jun,
  Tomer Kaftan, {\L}ukasz Kaiser, Ali Kamali, Ingmar Kanitscheider,
  Nitish~Shirish Keskar, Tabarak Khan, Logan Kilpatrick, Jong~Wook Kim,
  Christina Kim, Yongjik Kim, Jan~Hendrik Kirchner, Jamie Kiros, Matt Knight,
  Daniel Kokotajlo, {\L}ukasz Kondraciuk, Andrew Kondrich, Aris Konstantinidis,
  Kyle Kosic, Gretchen Krueger, Vishal Kuo, Michael Lampe, Ikai Lan, Teddy Lee,
  Jan Leike, Jade Leung, Daniel Levy, Chak~Ming Li, Rachel Lim, Molly Lin,
  Stephanie Lin, Mateusz Litwin, Theresa Lopez, Ryan Lowe, Patricia Lue, Anna
  Makanju, Kim Malfacini, Sam Manning, Todor Markov, Yaniv Markovski, Bianca
  Martin, Katie Mayer, Andrew Mayne, Bob McGrew, Scott~Mayer McKinney,
  Christine McLeavey, Paul McMillan, Jake McNeil, David Medina, Aalok Mehta,
  Jacob Menick, Luke Metz, Andrey Mishchenko, Pamela Mishkin, Vinnie Monaco,
  Evan Morikawa, Daniel Mossing, Tong Mu, Mira Murati, Oleg Murk, David
  M{\'e}ly, Ashvin Nair, Reiichiro Nakano, Rajeev Nayak, Arvind Neelakantan,
  Richard Ngo, Hyeonwoo Noh, Long Ouyang, Cullen O'Keefe, Jakub Pachocki, Alex
  Paino, Joe Palermo, Ashley Pantuliano, Giambattista Parascandolo, Joel
  Parish, Emy Parparita, Alex Passos, Mikhail Pavlov, Andrew Peng, Adam
  Perelman, Filipe de~Avila Belbute~Peres, Michael Petrov, Henrique~Ponde
  de~Oliveira~Pinto, Michael, Pokorny, Michelle Pokrass, Vitchyr~H. Pong, Tolly
  Powell, Alethea Power, Boris Power, Elizabeth Proehl, Raul Puri, Alec
  Radford, Jack Rae, Aditya Ramesh, Cameron Raymond, Francis Real, Kendra
  Rimbach, Carl Ross, Bob Rotsted, Henri Roussez, Nick Ryder, Mario Saltarelli,
  Ted Sanders, Shibani Santurkar, Girish Sastry, Heather Schmidt, David
  Schnurr, John Schulman, Daniel Selsam, Kyla Sheppard, Toki Sherbakov, Jessica
  Shieh, Sarah Shoker, Pranav Shyam, Szymon Sidor, Eric Sigler, Maddie Simens,
  Jordan Sitkin, Katarina Slama, Ian Sohl, Benjamin Sokolowsky, Yang Song,
  Natalie Staudacher, Felipe~Petroski Such, Natalie Summers, Ilya Sutskever,
  Jie Tang, Nikolas Tezak, Madeleine~B. Thompson, Phil Tillet, Amin
  Tootoonchian, Elizabeth Tseng, Preston Tuggle, Nick Turley, Jerry Tworek,
  Juan Felipe~Cer{\'o}n Uribe, Andrea Vallone, Arun Vijayvergiya, Chelsea Voss,
  Carroll Wainwright, Justin~Jay Wang, Alvin Wang, Ben Wang, Jonathan Ward,
  Jason Wei, CJ~Weinmann, Akila Welihinda, Peter Welinder, Jiayi Weng, Lilian
  Weng, Matt Wiethoff, Dave Willner, Clemens Winter, Samuel Wolrich, Hannah
  Wong, Lauren Workman, Sherwin Wu, Jeff Wu, Michael Wu, Kai Xiao, Tao Xu,
  Sarah Yoo, Kevin Yu, Qiming Yuan, Wojciech Zaremba, Rowan Zellers, Chong
  Zhang, Marvin Zhang, Shengjia Zhao, Tianhao Zheng, Juntang Zhuang, William
  Zhuk, and Barret Zoph.
\newblock Gpt-4 technical report, 2024.
\newblock URL \url{https://arxiv.org/abs/2303.08774}.

\bibitem[Orosz et~al.(2022)Orosz, Sz{\'a}nt{\'o}, Berkecz, Szab{\'o}, and
  Farkas]{HuSpaCy2021}
Gy{\"o}rgy Orosz, Zsolt Sz{\'a}nt{\'o}, P{\'e}ter Berkecz, Gerg{\H o}
  Szab{\'o}, and Rich{\'a}rd Farkas.
\newblock {HuSpaCy: an industrial-strength Hungarian natural language
  processing toolkit}.
\newblock In \emph{{XVIII. Magyar Sz{\'a}m{\'\i}t{\'o}g{\'e}pes Nyelv{\'e}szeti
  Konferencia}}, pages 59--73, 2022.

\bibitem[Orosz et~al.(2023)Orosz, Szab{\'o}, Berkecz, Sz{\'a}nt{\'o}, and
  Farkas]{HuSpaCy2023}
Gy{\"o}rgy Orosz, Gerg{\H{o}} Szab{\'o}, P{\'e}ter Berkecz, Zsolt
  Sz{\'a}nt{\'o}, and Rich{\'a}rd Farkas.
\newblock {"Advancing Hungarian Text Processing with HuSpaCy: Efficient and
  Accurate NLP Pipelines"}.
\newblock In Kamil "Ek{\v{s}}tein, Franti{\v{s}}ek P{\'a}rtl, and Miloslav"
  Konop{\'\i}k, editors, \emph{{"Text, Speech, and Dialogue"}}, pages 58--69,
  {"Cham"}, 2023. {"Springer Nature Switzerland"}.
\newblock ISBN 978-3-031-40498-6.

\bibitem[Pagnoni et~al.(2025)Pagnoni, Pasunuru, Rodriguez, Nguyen, Muller, Li,
  Zhou, Yu, Weston, Zettlemoyer, Ghosh, Lewis, Holtzman, and
  Iyer]{pagnoni2025BLT}
Artidoro Pagnoni, Ramakanth Pasunuru, Pedro Rodriguez, John Nguyen, Benjamin
  Muller, Margaret Li, Chunting Zhou, Lili Yu, Jason~E Weston, Luke
  Zettlemoyer, Gargi Ghosh, Mike Lewis, Ari Holtzman, and Srini Iyer.
\newblock Byte latent transformer: Patches scale better than tokens.
\newblock In Wanxiang Che, Joyce Nabende, Ekaterina Shutova, and Mohammad~Taher
  Pilehvar, editors, \emph{Proceedings of the 63rd Annual Meeting of the
  Association for Computational Linguistics (Volume 1: Long Papers)}, pages
  9238--9258, Vienna, Austria, July 2025. Association for Computational
  Linguistics.
\newblock ISBN 979-8-89176-251-0.
\newblock \doi{10.18653/v1/2025.acl-long.453}.
\newblock URL \url{https://aclanthology.org/2025.acl-long.453/}.

\bibitem[Penedo et~al.(2024)Penedo, Kydl{\'\i}{\v{c}}ek, allal, Lozhkov,
  Mitchell, Raffel, Werra, and Wolf]{penedo2024fineweb}
Guilherme Penedo, Hynek Kydl{\'\i}{\v{c}}ek, Loubna~Ben allal, Anton Lozhkov,
  Margaret Mitchell, Colin Raffel, Leandro~Von Werra, and Thomas Wolf.
\newblock The fineweb datasets: Decanting the web for the finest text data at
  scale.
\newblock In \emph{The Thirty-eight Conference on Neural Information Processing
  Systems Datasets and Benchmarks Track}, 2024.
\newblock URL \url{https://openreview.net/forum?id=n6SCkn2QaG}.

\bibitem[Petrov et~al.(2023)Petrov, Malfa, Torr, and
  Bibi]{petrovLanguageModelTokenizers2023}
Aleksandar Petrov, Emanuele~La Malfa, Philip Torr, and Adel Bibi.
\newblock Language {Model} {Tokenizers} {Introduce} {Unfairness} {Between}
  {Languages}.
\newblock November 2023.
\newblock URL \url{https://openreview.net/forum?id=78yDLKi95p}.

\bibitem[Pfeiffer et~al.(2020)Pfeiffer, Vuli{\'c}, Gurevych, and
  Ruder]{pfeiffer-etal-2020-mad}
Jonas Pfeiffer, Ivan Vuli{\'c}, Iryna Gurevych, and Sebastian Ruder.
\newblock {MAD-X}: {A}n {A}dapter-{B}ased {F}ramework for {M}ulti-{T}ask
  {C}ross-{L}ingual {T}ransfer.
\newblock In Bonnie Webber, Trevor Cohn, Yulan He, and Yang Liu, editors,
  \emph{Proceedings of the 2020 Conference on Empirical Methods in Natural
  Language Processing (EMNLP)}, pages 7654--7673, Online, November 2020.
  Association for Computational Linguistics.
\newblock \doi{10.18653/v1/2020.emnlp-main.617}.
\newblock URL \url{https://aclanthology.org/2020.emnlp-main.617/}.

\bibitem[Ponti et~al.(2020)Ponti, Glava{\v{s}}, Majewska, Liu, Vuli{\'c}, and
  Korhonen]{ponti2020xcopa}
Edoardo~Maria Ponti, Goran Glava{\v{s}}, Olga Majewska, Qianchu Liu, Ivan
  Vuli{\'c}, and Anna Korhonen.
\newblock {XCOPA}: A multilingual dataset for causal commonsense reasoning.
\newblock In Bonnie Webber, Trevor Cohn, Yulan He, and Yang Liu, editors,
  \emph{Proceedings of the 2020 Conference on Empirical Methods in Natural
  Language Processing (EMNLP)}, pages 2362--2376, Online, November 2020.
  Association for Computational Linguistics.
\newblock \doi{10.18653/v1/2020.emnlp-main.185}.
\newblock URL \url{https://aclanthology.org/2020.emnlp-main.185/}.

\bibitem[Provilkov et~al.(2020)Provilkov, Emelianenko, and
  Voita]{provilkov2020bpeDropout}
Ivan Provilkov, Dmitrii Emelianenko, and Elena Voita.
\newblock {BPE}-dropout: Simple and effective subword regularization.
\newblock In Dan Jurafsky, Joyce Chai, Natalie Schluter, and Joel Tetreault,
  editors, \emph{Proceedings of the 58th Annual Meeting of the Association for
  Computational Linguistics}, pages 1882--1892, Online, July 2020. Association
  for Computational Linguistics.
\newblock \doi{10.18653/v1/2020.acl-main.170}.
\newblock URL \url{https://aclanthology.org/2020.acl-main.170/}.

\bibitem[Radford et~al.(2019)Radford, Wu, Child, Luan, Amodei, and
  Sutskever]{radford2019GPT2}
Alec Radford, Jeff Wu, Rewon Child, David Luan, Dario Amodei, and Ilya
  Sutskever.
\newblock Language models are unsupervised multitask learners.
\newblock 2019.

\bibitem[Raffel et~al.(2020)Raffel, Shazeer, Roberts, Lee, Narang, Matena,
  Zhou, Li, and Liu]{raffel2020T5}
Colin Raffel, Noam Shazeer, Adam Roberts, Katherine Lee, Sharan Narang, Michael
  Matena, Yanqi Zhou, Wei Li, and Peter~J. Liu.
\newblock Exploring the limits of transfer learning with a unified text-to-text
  transformer.
\newblock \emph{J. Mach. Learn. Res.}, 21\penalty0 (1), January 2020.
\newblock ISSN 1532-4435.

\bibitem[Research()]{hazmFarsi}
Roshan Research.
\newblock Hazm: Persian text processing library.
\newblock \url{https://github.com/roshan-research/hazm}.

\bibitem[Roemmele et~al.(2011)Roemmele, Bejan, and Gordon]{roemmele2011copa1}
Melissa Roemmele, Cosmin~Adrian Bejan, and Andrew~S Gordon.
\newblock Choice of plausible alternatives: An evaluation of commonsense causal
  reasoning.
\newblock In \emph{2011 AAAI Spring Symposium Series}, 2011.
\newblock URL
  \url{https://people.ict.usc.edu/~gordon/publications/AAAI-SPRING11A.PDF}.

\bibitem[Romanou et~al.(2024)Romanou, Foroutan, Sotnikova, Chen, Nelaturu,
  Singh, Maheshwary, Altomare, Haggag, Amayuelas, et~al.]{romanou2024include}
Angelika Romanou, Negar Foroutan, Anna Sotnikova, Zeming Chen, Sree~Harsha
  Nelaturu, Shivalika Singh, Rishabh Maheshwary, Micol Altomare, Mohamed~A
  Haggag, Alfonso Amayuelas, et~al.
\newblock Include: Evaluating multilingual language understanding with regional
  knowledge.
\newblock \emph{arXiv preprint arXiv:2411.19799}, 2024.

\bibitem[Rust et~al.(2021)Rust, Pfeiffer, Vuli{\'c}, Ruder, and
  Gurevych]{rust2021howGood}
Phillip Rust, Jonas Pfeiffer, Ivan Vuli{\'c}, Sebastian Ruder, and Iryna
  Gurevych.
\newblock How good is your tokenizer? on the monolingual performance of
  multilingual language models.
\newblock In Chengqing Zong, Fei Xia, Wenjie Li, and Roberto Navigli, editors,
  \emph{Proceedings of the 59th Annual Meeting of the Association for
  Computational Linguistics and the 11th International Joint Conference on
  Natural Language Processing (Volume 1: Long Papers)}, pages 3118--3135,
  Online, August 2021. Association for Computational Linguistics.
\newblock \doi{10.18653/v1/2021.acl-long.243}.
\newblock URL \url{https://aclanthology.org/2021.acl-long.243/}.

\bibitem[Schmidt et~al.(2024)Schmidt, Reddy, Zhang, Alameddine, Uzan, Pinter,
  and Tanner]{schmidt2024Compression}
Craig~W Schmidt, Varshini Reddy, Haoran Zhang, Alec Alameddine, Omri Uzan,
  Yuval Pinter, and Chris Tanner.
\newblock Tokenization is more than compression.
\newblock In Yaser Al-Onaizan, Mohit Bansal, and Yun-Nung Chen, editors,
  \emph{Proceedings of the 2024 Conference on Empirical Methods in Natural
  Language Processing}, pages 678--702, Miami, Florida, USA, November 2024.
  Association for Computational Linguistics.
\newblock \doi{10.18653/v1/2024.emnlp-main.40}.
\newblock URL \url{https://aclanthology.org/2024.emnlp-main.40/}.

\bibitem[Schmidt et~al.(2025)Schmidt, Reddy, Tanner, and
  Pinter]{schmidt2025boundlessBPE}
Craig~W Schmidt, Varshini Reddy, Chris Tanner, and Yuval Pinter.
\newblock Boundless byte pair encoding: Breaking the pre-tokenization barrier.
\newblock In \emph{Second Conference on Language Modeling}, 2025.
\newblock URL \url{https://openreview.net/forum?id=oPAjXGV8qQ}.

\bibitem[Sennrich et~al.(2016)Sennrich, Haddow, and Birch]{sennrich2016BPE}
Rico Sennrich, Barry Haddow, and Alexandra Birch.
\newblock Neural machine translation of rare words with subword units.
\newblock In Katrin Erk and Noah~A. Smith, editors, \emph{Proceedings of the
  54th Annual Meeting of the Association for Computational Linguistics (Volume
  1: Long Papers)}, pages 1715--1725, Berlin, Germany, August 2016. Association
  for Computational Linguistics.
\newblock \doi{10.18653/v1/P16-1162}.
\newblock URL \url{https://aclanthology.org/P16-1162/}.

\bibitem[Shazeer et~al.(2017)Shazeer, Mirhoseini, Maziarz, Davis, Le, Hinton,
  and Dean]{shazeer2017MOE}
Noam Shazeer, *Azalia Mirhoseini, *Krzysztof Maziarz, Andy Davis, Quoc Le,
  Geoffrey Hinton, and Jeff Dean.
\newblock Outrageously large neural networks: The sparsely-gated
  mixture-of-experts layer.
\newblock In \emph{International Conference on Learning Representations}, 2017.
\newblock URL \url{https://openreview.net/forum?id=B1ckMDqlg}.

\bibitem[Sims et~al.(2026)Sims, Foster, Kaleb, Nguyen, Lee, Foerster, Teh, and
  Lu]{sims2026stochastok}
Anya Sims, Thom Foster, Klara Kaleb, Tuan-Duy~H. Nguyen, Joseph Lee, Jakob~N.
  Foerster, Yee~Whye Teh, and Cong Lu.
\newblock Stochastok: Improving fine-grained subword understanding in llms,
  2026.
\newblock URL \url{https://arxiv.org/abs/2506.01687}.

\bibitem[Singh and Strouse(2024)]{singhTokenizationCountsImpact2024}
Aaditya~K. Singh and D.~J. Strouse.
\newblock Tokenization counts: the impact of tokenization on arithmetic in
  frontier {LLMs}, February 2024.
\newblock URL \url{http://arxiv.org/abs/2402.14903}.
\newblock arXiv:2402.14903 [cs.CL].

\bibitem[Steger et~al.(2026)Steger, Li, Ennadir, Sims, Solin, Pernkopf, and
  Trapp]{steger2026stochasticityRobustness}
Sophie Steger, Rui Li, Sofiane Ennadir, Anya Sims, Arno Solin, Franz Pernkopf,
  and Martin Trapp.
\newblock Stochasticity in tokenisation improves robustness, 2026.
\newblock URL \url{https://arxiv.org/abs/2604.16037}.

\bibitem[Strakov\'{a} et~al.(2014)Strakov\'{a}, Straka, and
  Haji\v{c}]{strakova2014MorphoditaCzech}
Jana Strakov\'{a}, Milan Straka, and Jan Haji\v{c}.
\newblock Open-{S}ource {T}ools for {M}orphology, {L}emmatization, {POS}
  {T}agging and {N}amed {E}ntity {R}ecognition.
\newblock In \emph{Proceedings of 52nd Annual Meeting of the Association for
  Computational Linguistics: System Demonstrations}, pages 13--18, Baltimore,
  Maryland, June 2014. Association for Computational Linguistics.
\newblock URL \url{http://www.aclweb.org/anthology/P/P14/P14-5003.pdf}.

\bibitem[Talmor et~al.(2019)Talmor, Herzig, Lourie, and
  Berant]{talmor2019commonsenseqa}
Alon Talmor, Jonathan Herzig, Nicholas Lourie, and Jonathan Berant.
\newblock {C}ommonsense{QA}: A question answering challenge targeting
  commonsense knowledge.
\newblock In Jill Burstein, Christy Doran, and Thamar Solorio, editors,
  \emph{Proceedings of the 2019 Conference of the North {A}merican Chapter of
  the Association for Computational Linguistics: Human Language Technologies,
  Volume 1 (Long and Short Papers)}, pages 4149--4158, Minneapolis, Minnesota,
  June 2019. Association for Computational Linguistics.
\newblock \doi{10.18653/v1/N19-1421}.
\newblock URL \url{https://aclanthology.org/N19-1421/}.

\bibitem[Team et~al.(2024)Team, Riviere, Pathak, Sessa, Hardin, Bhupatiraju,
  Hussenot, Mesnard, Shahriari, Ram{\'e}, Ferret, Liu, Tafti, Friesen, Casbon,
  Ramos, Kumar, Lan, Jerome, Tsitsulin, Vieillard, Stanczyk, Girgin, Momchev,
  Hoffman, Thakoor, Grill, Neyshabur, Bachem, Walton, Severyn, Parrish, Ahmad,
  Hutchison, Abdagic, Carl, Shen, Brock, Coenen, Laforge, Paterson, Bastian,
  Piot, Wu, Royal, Chen, Kumar, Perry, Welty, Choquette-Choo, Sinopalnikov,
  Weinberger, Vijaykumar, Rogozi{\'n}ska, Herbison, Bandy, Wang, Noland,
  Moreira, Senter, Eltyshev, Visin, Rasskin, Wei, Cameron, Martins, Hashemi,
  Klimczak-Pluci{\'n}ska, Batra, Dhand, Nardini, Mein, Zhou, Svensson, Stanway,
  Chan, Zhou, Carrasqueira, Iljazi, Becker, Fernandez, van Amersfoort, Gordon,
  Lipschultz, Newlan, yeong Ji, Mohamed, Badola, Black, Millican, McDonell,
  Nguyen, Sodhia, Greene, Sjoesund, Usui, Sifre, Heuermann, Lago, McNealus,
  Soares, Kilpatrick, Dixon, Martins, Reid, Singh, Iverson, G{\"o}rner,
  Velloso, Wirth, Davidow, Miller, Rahtz, Watson, Risdal, Kazemi, Moynihan,
  Zhang, Kahng, Park, Rahman, Khatwani, Dao, Bardoliwalla, Devanathan, Dumai,
  Chauhan, Wahltinez, Botarda, Barnes, Barham, Michel, Jin, Georgiev, Culliton,
  Kuppala, Comanescu, Merhej, Jana, Rokni, Agarwal, Mullins, Saadat, Carthy,
  Cogan, Perrin, Arnold, Krause, Dai, Garg, Sheth, Ronstrom, Chan, Jordan, Yu,
  Eccles, Hennigan, Kocisky, Doshi, Jain, Yadav, Meshram, Dharmadhikari,
  Barkley, Wei, Ye, Han, Kwon, Xu, Shen, Gong, Wei, Cotruta, Kirk, Rao, Giang,
  Peran, Warkentin, Collins, Barral, Ghahramani, Hadsell, Sculley, Banks,
  Dragan, Petrov, Vinyals, Dean, Hassabis, Kavukcuoglu, Farabet, Buchatskaya,
  Borgeaud, Fiedel, Joulin, Kenealy, Dadashi, and Andreev]{gemma2024gemma2}
Gemma Team, Morgane Riviere, Shreya Pathak, Pier~Giuseppe Sessa, Cassidy
  Hardin, Surya Bhupatiraju, L{\'e}onard Hussenot, Thomas Mesnard, Bobak
  Shahriari, Alexandre Ram{\'e}, Johan Ferret, Peter Liu, Pouya Tafti, Abe
  Friesen, Michelle Casbon, Sabela Ramos, Ravin Kumar, Charline~Le Lan, Sammy
  Jerome, Anton Tsitsulin, Nino Vieillard, Piotr Stanczyk, Sertan Girgin,
  Nikola Momchev, Matt Hoffman, Shantanu Thakoor, Jean-Bastien Grill, Behnam
  Neyshabur, Olivier Bachem, Alanna Walton, Aliaksei Severyn, Alicia Parrish,
  Aliya Ahmad, Allen Hutchison, Alvin Abdagic, Amanda Carl, Amy Shen, Andy
  Brock, Andy Coenen, Anthony Laforge, Antonia Paterson, Ben Bastian, Bilal
  Piot, Bo~Wu, Brandon Royal, Charlie Chen, Chintu Kumar, Chris Perry, Chris
  Welty, Christopher~A. Choquette-Choo, Danila Sinopalnikov, David Weinberger,
  Dimple Vijaykumar, Dominika Rogozi{\'n}ska, Dustin Herbison, Elisa Bandy,
  Emma Wang, Eric Noland, Erica Moreira, Evan Senter, Evgenii Eltyshev,
  Francesco Visin, Gabriel Rasskin, Gary Wei, Glenn Cameron, Gus Martins, Hadi
  Hashemi, Hanna Klimczak-Pluci{\'n}ska, Harleen Batra, Harsh Dhand, Ivan
  Nardini, Jacinda Mein, Jack Zhou, James Svensson, Jeff Stanway, Jetha Chan,
  Jin~Peng Zhou, Joana Carrasqueira, Joana Iljazi, Jocelyn Becker, Joe
  Fernandez, Joost van Amersfoort, Josh Gordon, Josh Lipschultz, Josh Newlan,
  Ju~yeong Ji, Kareem Mohamed, Kartikeya Badola, Kat Black, Katie Millican,
  Keelin McDonell, Kelvin Nguyen, Kiranbir Sodhia, Kish Greene, Lars~Lowe
  Sjoesund, Lauren Usui, Laurent Sifre, Lena Heuermann, Leticia Lago, Lilly
  McNealus, Livio~Baldini Soares, Logan Kilpatrick, Lucas Dixon, Luciano
  Martins, Machel Reid, Manvinder Singh, Mark Iverson, Martin G{\"o}rner, Mat
  Velloso, Mateo Wirth, Matt Davidow, Matt Miller, Matthew Rahtz, Matthew
  Watson, Meg Risdal, Mehran Kazemi, Michael Moynihan, Ming Zhang, Minsuk
  Kahng, Minwoo Park, Mofi Rahman, Mohit Khatwani, Natalie Dao, Nenshad
  Bardoliwalla, Nesh Devanathan, Neta Dumai, Nilay Chauhan, Oscar Wahltinez,
  Pankil Botarda, Parker Barnes, Paul Barham, Paul Michel, Pengchong Jin, Petko
  Georgiev, Phil Culliton, Pradeep Kuppala, Ramona Comanescu, Ramona Merhej,
  Reena Jana, Reza~Ardeshir Rokni, Rishabh Agarwal, Ryan Mullins, Samaneh
  Saadat, Sara~Mc Carthy, Sarah Cogan, Sarah Perrin, S{\'e}bastien M.~R.
  Arnold, Sebastian Krause, Shengyang Dai, Shruti Garg, Shruti Sheth, Sue
  Ronstrom, Susan Chan, Timothy Jordan, Ting Yu, Tom Eccles, Tom Hennigan,
  Tomas Kocisky, Tulsee Doshi, Vihan Jain, Vikas Yadav, Vilobh Meshram, Vishal
  Dharmadhikari, Warren Barkley, Wei Wei, Wenming Ye, Woohyun Han, Woosuk Kwon,
  Xiang Xu, Zhe Shen, Zhitao Gong, Zichuan Wei, Victor Cotruta, Phoebe Kirk,
  Anand Rao, Minh Giang, Ludovic Peran, Tris Warkentin, Eli Collins, Joelle
  Barral, Zoubin Ghahramani, Raia Hadsell, D.~Sculley, Jeanine Banks, Anca
  Dragan, Slav Petrov, Oriol Vinyals, Jeff Dean, Demis Hassabis, Koray
  Kavukcuoglu, Clement Farabet, Elena Buchatskaya, Sebastian Borgeaud, Noah
  Fiedel, Armand Joulin, Kathleen Kenealy, Robert Dadashi, and Alek Andreev.
\newblock Gemma 2: Improving open language models at a practical size, 2024.
\newblock URL \url{https://arxiv.org/abs/2408.00118}.

\bibitem[Team et~al.(2022)Team, Costa-juss{\`a}, Cross, {\c C}elebi, Elbayad,
  Heafield, Heffernan, Kalbassi, Lam, Licht, Maillard, Sun, Wang, Wenzek,
  Youngblood, Akula, Barrault, Gonzalez, Hansanti, Hoffman, Jarrett, Sadagopan,
  Rowe, Spruit, Tran, Andrews, Ayan, Bhosale, Edunov, Fan, Gao, Goswami,
  Guzm{\'a}n, Koehn, Mourachko, Ropers, Saleem, Schwenk, and
  Wang]{flores1-nllb2022}
NLLB Team, Marta~R. Costa-juss{\`a}, James Cross, Onur {\c C}elebi, Maha
  Elbayad, Kenneth Heafield, Kevin Heffernan, Elahe Kalbassi, Janice Lam,
  Daniel Licht, Jean Maillard, Anna Sun, Skyler Wang, Guillaume Wenzek,
  Al~Youngblood, Bapi Akula, Loic Barrault, Gabriel~Mejia Gonzalez, Prangthip
  Hansanti, John Hoffman, Semarley Jarrett, Kaushik~Ram Sadagopan, Dirk Rowe,
  Shannon Spruit, Chau Tran, Pierre Andrews, Necip~Fazil Ayan, Shruti Bhosale,
  Sergey Edunov, Angela Fan, Cynthia Gao, Vedanuj Goswami, Francisco
  Guzm{\'a}n, Philipp Koehn, Alexandre Mourachko, Christophe Ropers, Safiyyah
  Saleem, Holger Schwenk, and Jeff Wang.
\newblock No language left behind: Scaling human-centered machine translation,
  2022.
\newblock URL \url{https://arxiv.org/abs/2207.04672}.

\bibitem[Tikhonov and Ryabinin(2021)]{tikhonov2021heads}
Alexey Tikhonov and Max Ryabinin.
\newblock It's all in the heads: Using attention heads as a baseline for
  cross-lingual transfer in commonsense reasoning, 2021.

\bibitem[Tran(2020)]{tran2020foreignTransfer}
Ke~Tran.
\newblock From english to foreign languages: Transferring pre-trained language
  models, 2020.
\newblock URL \url{https://arxiv.org/abs/2002.07306}.

\bibitem[{\"U}st{\"u}n et~al.(2022){\"U}st{\"u}n, Bisazza, Bouma, and van
  Noord]{ustun-etal-2022-udapter}
Ahmet {\"U}st{\"u}n, Arianna Bisazza, Gosse Bouma, and Gertjan van Noord.
\newblock {UD}apter: Typology-based language adapters for multilingual
  dependency parsing and sequence labeling.
\newblock \emph{Computational Linguistics}, 48\penalty0 (3):\penalty0 555--592,
  September 2022.
\newblock \doi{10.1162/coli_a_00443}.
\newblock URL \url{https://aclanthology.org/2022.cl-3.3/}.

\bibitem[{\"U}st{\"u}n et~al.(2024){\"U}st{\"u}n, Aryabumi, Yong, Ko,
  D{'}souza, Onilude, Bhandari, Singh, Ooi, Kayid, Vargus, Blunsom, Longpre,
  Muennighoff, Fadaee, Kreutzer, and Hooker]{ustun2024aya}
Ahmet {\"U}st{\"u}n, Viraat Aryabumi, Zheng Yong, Wei-Yin Ko, Daniel D{'}souza,
  Gbemileke Onilude, Neel Bhandari, Shivalika Singh, Hui-Lee Ooi, Amr Kayid,
  Freddie Vargus, Phil Blunsom, Shayne Longpre, Niklas Muennighoff, Marzieh
  Fadaee, Julia Kreutzer, and Sara Hooker.
\newblock Aya model: An instruction finetuned open-access multilingual language
  model.
\newblock In Lun-Wei Ku, Andre Martins, and Vivek Srikumar, editors,
  \emph{Proceedings of the 62nd Annual Meeting of the Association for
  Computational Linguistics (Volume 1: Long Papers)}, pages 15894--15939,
  Bangkok, Thailand, August 2024. Association for Computational Linguistics.
\newblock \doi{10.18653/v1/2024.acl-long.845}.
\newblock URL \url{https://aclanthology.org/2024.acl-long.845/}.

\bibitem[Videau et~al.(2024)Videau, Idrissi, Haziza, Wehrstedt, Copet, Teytaud,
  and Lopez-Paz]{meta_lingua}
Mathurin Videau, Badr~Youbi Idrissi, Daniel Haziza, Luca Wehrstedt, Jade Copet,
  Olivier Teytaud, and David Lopez-Paz.
\newblock {Meta Lingua}: A minimal {PyTorch LLM} training library, 2024.
\newblock URL \url{https://github.com/facebookresearch/lingua}.

\bibitem[Wegmann et~al.(2025)Wegmann, Nguyen, and Jurgens]{wegmann2025senstive}
Anna Wegmann, Dong Nguyen, and David Jurgens.
\newblock Tokenization is sensitive to language variation.
\newblock In Wanxiang Che, Joyce Nabende, Ekaterina Shutova, and Mohammad~Taher
  Pilehvar, editors, \emph{Findings of the Association for Computational
  Linguistics: ACL 2025}, pages 10958--10983, Vienna, Austria, July 2025.
  Association for Computational Linguistics.
\newblock ISBN 979-8-89176-256-5.
\newblock \doi{10.18653/v1/2025.findings-acl.572}.
\newblock URL \url{https://aclanthology.org/2025.findings-acl.572/}.

\bibitem[Woli{\'n}ski(2014)]{wolinski2014morfeuszPolish}
Marcin Woli{\'n}ski.
\newblock Morfeusz reloaded.
\newblock In Nicoletta Calzolari, Khalid Choukri, Thierry Declerck, Hrafn
  Loftsson, Bente Maegaard, Joseph Mariani, Asuncion Moreno, Jan Odijk, and
  Stelios Piperidis, editors, \emph{Proceedings of the Ninth International
  Conference on Language Resources and Evaluation ({LREC}'14)}, pages
  1106--1111, Reykjavik, Iceland, May 2014. European Language Resources
  Association (ELRA).
\newblock URL \url{https://aclanthology.org/L14-1597/}.

\bibitem[Xue et~al.(2022)Xue, Barua, Constant, Al-Rfou, Narang, Kale, Roberts,
  and Raffel]{xue2022byt5}
Linting Xue, Aditya Barua, Noah Constant, Rami Al-Rfou, Sharan Narang, Mihir
  Kale, Adam Roberts, and Colin Raffel.
\newblock {B}y{T}5: Towards a token-free future with pre-trained byte-to-byte
  models.
\newblock \emph{Transactions of the Association for Computational Linguistics},
  10:\penalty0 291--306, 2022.
\newblock \doi{10.1162/tacl_a_00461}.
\newblock URL \url{https://aclanthology.org/2022.tacl-1.17/}.

\bibitem[Yang et~al.(2025)Yang, Li, Yang, Zhang, Hui, Zheng, Yu, Gao, Huang,
  Lv, Zheng, Liu, Zhou, Huang, Hu, Ge, Wei, Lin, Tang, Yang, Tu, Zhang, Yang,
  Yang, Zhou, Zhou, Lin, Dang, Bao, Yang, Yu, Deng, Li, Xue, Li, Zhang, Wang,
  Zhu, Men, Gao, Liu, Luo, Li, Tang, Yin, Ren, Wang, Zhang, Ren, Fan, Su,
  Zhang, Zhang, Wan, Liu, Wang, Cui, Zhang, Zhou, and Qiu]{yang2025Qwen}
An~Yang, Anfeng Li, Baosong Yang, Beichen Zhang, Binyuan Hui, Bo~Zheng, Bowen
  Yu, Chang Gao, Chengen Huang, Chenxu Lv, Chujie Zheng, Dayiheng Liu, Fan
  Zhou, Fei Huang, Feng Hu, Hao Ge, Haoran Wei, Huan Lin, Jialong Tang, Jian
  Yang, Jianhong Tu, Jianwei Zhang, Jianxin Yang, Jiaxi Yang, Jing Zhou,
  Jingren Zhou, Junyang Lin, Kai Dang, Keqin Bao, Kexin Yang, Le~Yu, Lianghao
  Deng, Mei Li, Mingfeng Xue, Mingze Li, Pei Zhang, Peng Wang, Qin Zhu, Rui
  Men, Ruize Gao, Shixuan Liu, Shuang Luo, Tianhao Li, Tianyi Tang, Wenbiao
  Yin, Xingzhang Ren, Xinyu Wang, Xinyu Zhang, Xuancheng Ren, Yang Fan, Yang
  Su, Yichang Zhang, Yinger Zhang, Yu~Wan, Yuqiong Liu, Zekun Wang, Zeyu Cui,
  Zhenru Zhang, Zhipeng Zhou, and Zihan Qiu.
\newblock Qwen3 technical report, 2025.
\newblock URL \url{https://arxiv.org/abs/2505.09388}.

\bibitem[Yang and Chai(2025)]{yang2025codemixbench}
Yilun Yang and Yekun Chai.
\newblock {C}ode{M}ix{B}ench: Evaluating code-mixing capabilities of {LLM}s
  across 18 languages.
\newblock In Christos Christodoulopoulos, Tanmoy Chakraborty, Carolyn Rose, and
  Violet Peng, editors, \emph{Proceedings of the 2025 Conference on Empirical
  Methods in Natural Language Processing}, pages 2139--2169, Suzhou, China,
  November 2025. Association for Computational Linguistics.
\newblock ISBN 979-8-89176-332-6.
\newblock \doi{10.18653/v1/2025.emnlp-main.109}.
\newblock URL \url{https://aclanthology.org/2025.emnlp-main.109/}.

\bibitem[Yang et~al.(2019)Yang, Zhang, Tar, and Baldridge]{yang2019pawsx}
Yinfei Yang, Yuan Zhang, Chris Tar, and Jason Baldridge.
\newblock {PAWS}-{X}: A cross-lingual adversarial dataset for paraphrase
  identification.
\newblock In Kentaro Inui, Jing Jiang, Vincent Ng, and Xiaojun Wan, editors,
  \emph{Proceedings of the 2019 Conference on Empirical Methods in Natural
  Language Processing and the 9th International Joint Conference on Natural
  Language Processing (EMNLP-IJCNLP)}, pages 3687--3692, Hong Kong, China,
  November 2019. Association for Computational Linguistics.
\newblock \doi{10.18653/v1/D19-1382}.
\newblock URL \url{https://aclanthology.org/D19-1382/}.

\bibitem[Yu et~al.(2023)Yu, Simig, Flaherty, Aghajanyan, Zettlemoyer, and
  Lewis]{yu2023megabyte}
Lili Yu, Daniel Simig, Colin Flaherty, Armen Aghajanyan, Luke Zettlemoyer, and
  Mike Lewis.
\newblock {MEGABYTE}: Predicting million-byte sequences with multiscale
  transformers.
\newblock In \emph{Thirty-seventh Conference on Neural Information Processing
  Systems}, 2023.
\newblock URL \url{https://openreview.net/forum?id=JTmO2V9Xpz}.

\bibitem[Zellers et~al.(2019)Zellers, Holtzman, Bisk, Farhadi, and
  Choi]{zellers2019hellaswag}
Rowan Zellers, Ari Holtzman, Yonatan Bisk, Ali Farhadi, and Yejin Choi.
\newblock {H}ella{S}wag: Can a machine really finish your sentence?
\newblock In Anna Korhonen, David Traum, and Llu{\'\i}s M{\`a}rquez, editors,
  \emph{Proceedings of the 57th Annual Meeting of the Association for
  Computational Linguistics}, pages 4791--4800, Florence, Italy, July 2019.
  Association for Computational Linguistics.
\newblock \doi{10.18653/v1/P19-1472}.
\newblock URL \url{https://aclanthology.org/P19-1472/}.

\bibitem[Zheng et~al.(2025)Zheng, Liu, Ahia, Hayase, Choi, and
  Smith]{zheng2026brokenTokens}
Brian~Siyuan Zheng, Alisa Liu, Orevaoghene Ahia, Jonathan Hayase, Yejin Choi,
  and Noah~A. Smith.
\newblock Broken tokens? your language model can secretly handle non-canonical
  tokenizations.
\newblock In \emph{The Thirty-ninth Annual Conference on Neural Information
  Processing Systems}, 2025.
\newblock URL \url{https://openreview.net/forum?id=WrYWolqKh3}.

\bibitem[Zhu et~al.(2025)Zhu, Hu, Ji, Wu, and Tu]{zhu2025morphTree}
Qingyang Zhu, Xiang Hu, Pengyu Ji, Wei Wu, and Kewei Tu.
\newblock Unsupervised morphological tree tokenizer.
\newblock In Wanxiang Che, Joyce Nabende, Ekaterina Shutova, and Mohammad~Taher
  Pilehvar, editors, \emph{Findings of the Association for Computational
  Linguistics: ACL 2025}, pages 22299--22312, Vienna, Austria, July 2025.
  Association for Computational Linguistics.
\newblock ISBN 979-8-89176-256-5.
\newblock \doi{10.18653/v1/2025.findings-acl.1146}.
\newblock URL \url{https://aclanthology.org/2025.findings-acl.1146/}.

\end{thebibliography}

\newpage
\appendix
\section*{Appendix Contents}
\noindent\ref{ap:sec:ethics}~~Ethics Statement \dotfill \pageref{ap:sec:ethics}\\[0.4em]
\noindent\ref{ap:sec:llms}~~Declaration of LLM Usage \dotfill \pageref{ap:sec:llms}\\[0.4em]
\noindent\ref{ap:sec:training}~~Multilingual Language Model Pre-training with \moto \dotfill \pageref{ap:sec:training}\\[0.4em]
\noindent\ref{ap:sec:full_results}~~Full Results \dotfill \pageref{ap:sec:full_results}\\
\noindent\hspace*{1.5em}\ref{ap:sec:full_results:maddition}~~Multilingual Addition \dotfill \pageref{ap:sec:full_results:maddition}\\
\noindent\hspace*{1.5em}\ref{ap:sec:full_results:pretraining}~~Multilingual Pre-training \dotfill \pageref{ap:sec:full_results:pretraining}\\[0.4em]
\noindent\ref{ap:sec:routing}~~Routing \dotfill \pageref{ap:sec:routing}\\[0.4em]
\noindent\ref{app:moto-merge}~~Merging Per-Language Tokenizers into a Monolithic BPE Tokenizer \dotfill \pageref{app:moto-merge}\\[0.4em]
\noindent\ref{ap:sec:intrinsics}~~Tokenizer Quality \& Compression \dotfill \pageref{ap:sec:intrinsics}

\section{Ethics Statement\label{ap:sec:ethics}}
This work addresses tokenization inequities in multilingual language models, a problem with direct consequences for speakers of underrepresented languages, who may experience higher costs and degraded model performance due to design choices made primarily for high-resource languages.
We acknowledge that our study currently covers 21 languages among many of the world's languages.

\section{Declaration of LLM Usage\label{ap:sec:llms}}
Large Language Models (LLMs) were used for grammar checking and feedback on clarity during the writing of this paper. Throughout the development of the project LLMs were also used for assisting in ideation, exploration, coding, diagram, and figure creation.

\section{Multilingual Language Model Pre-training with \moto\label{ap:sec:training}}

\paragraph{Model architecture.}
All experiments use Llama 3.2~\citep{grattafiori2024llama3} architecture with approximately one billion non-embedding parameters as the base transformer architecture ($d=2048$, 25 layers, 16 attention heads, no weight-tying). 

\paragraph{Model training.} All models are trained from scratch using the Meta Lingua framework~\citep{meta_lingua} with full parameter updates in \texttt{bfloat16} precision for 100,000 steps with AdamW~\citep{loshchilov2018AdamW} optimizer. 
We use a cosine learning rate schedule with a peak learning rate of $10^{-3}$, 2,000 warmup steps, gradient clipping at 1.0, and weight decay of 0.1. The effective batch size is 256 with a sequence length of 4,096 tokens.

Each model is trained on a single 8$\times$H100 node. Models with effective (superset) vocabulary size around 128K (e.g., Llama, \moto, \moto-GPT4, mBPE) take approximately 4 days to train, while larger ablations (e.g., \moto-16K or \moto-Script) take approximately 5.5 days.

\paragraph{Training data.} We train all models on a multilingual corpus combining English and 20 additional languages. English data is sampled from the 100 billion subset of FineWeb-Edu~\citep{lozhkov2024fineweb-edu}, a high-quality educational data filtered from FineWeb~\citep{penedo2024fineweb}; while non-English languages are sampled from FineWeb2-HQ~\citep{messmer2025fineweb2hq}. Sampling weights are set proportional to corpus availability following the UniMax strategy~\citep{chung2023unimax}, with English receiving 40\% of the training budget; full per-language weights are reported in Table~\ref{tab:data_mixture}.

\begin{table}[!htbp]
    \caption{Data mixture rates for pre-training following UniMax~\citep{chung2023unimax} sampling.}
    \label{tab:data_mixture}
    \centering
\begin{tabular}{@{}rlllr@{}}

\toprule
FLORES code & Language   & Script   & Mixture Rate  & Approx. \# Tokens (B/1e9) \\ \midrule
arb\_Arab   & Arabic     & Arabic   & 0.02343237066 & 2.46                    \\
cmn\_Hani   & Chinese    & Hani     & 0.04540675214 & 4.76                    \\
ces\_Latn   & Czech      & Latin    & 0.02480244553 & 2.60                    \\
dan\_Latn   & Danish     & Latin    & 0.02165828697 & 2.27                    \\
nld\_Latn   & Dutch      & Latin    & 0.02777963266 & 2.91                    \\
eng\_Latn     & English    & Latin    & 0.4           & 41.94                   \\
fra\_Latn   & French     & Latin    & 0.03818355607 & 4.00                    \\
deu\_Latn   & German     & Latin    & 0.04199253405 & 4.40                    \\
ell\_Grek   & Greek      & Greek    & 0.02336708869 & 2.45                    \\
hun\_Latn   & Hungarian  & Latin    & 0.02471945736 & 2.59                    \\
ind\_Latn   & Indonesian & Latin    & 0.02693220546 & 2.82                    \\
ita\_Latn   & Italian    & Latin    & 0.0324056371  & 3.40                    \\
jpn\_Jpan   & Japanese   & Japanese & 0.03553104151 & 3.73                    \\
fas\_Arab   & Farsi    & Arabic   & 0.02056346242 & 2.16                    \\
pol\_Latn   & Polish     & Latin    & 0.02941201046 & 3.08                    \\
por\_Latn   & Portuguese & Latin    & 0.03014131683 & 3.16                    \\
rus\_Cyrl   & Russian    & Cyrillic & 0.05001981574 & 5.24                    \\
spa\_Latn   & Spanish    & Latin    & 0.03872227931 & 4.06                    \\
swe\_Latn   & Swedish    & Latin    & 0.02163597654 & 2.27                    \\
tur\_Latn   & Turkish    & Latin    & 0.02354557948 & 2.47                    \\
vie\_Latn   & Vietnamese & Latin    & 0.01974855103 & 2.07                    \\\midrule
            &            &          &     Total     & 104.86                  \\ \bottomrule
\end{tabular}
\end{table}

\section{Full Results\label{ap:sec:full_results}}

\subsection{Multilingual Addition}\label{ap:sec:full_results:maddition}
\begin{figure}[H]
    \centering
    \includegraphics[width=\linewidth]{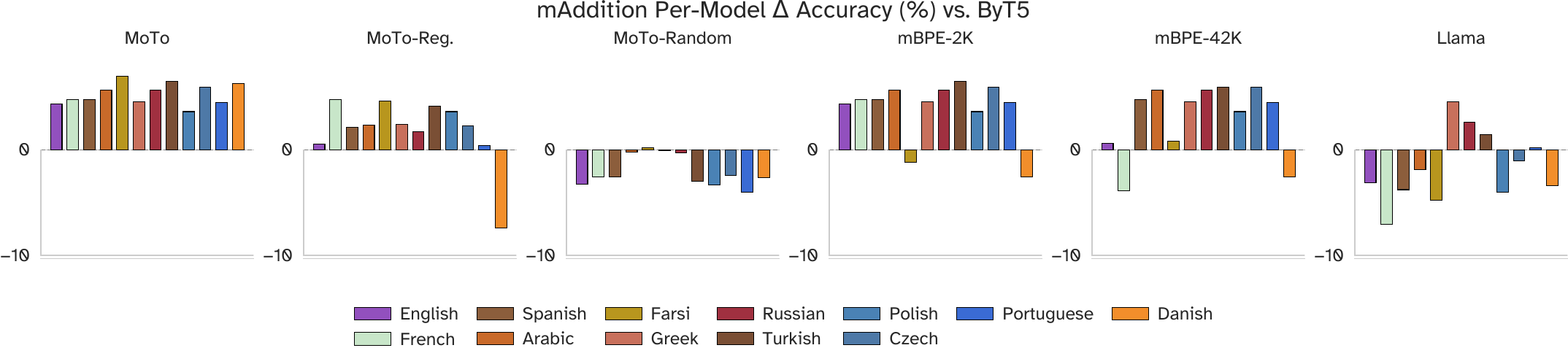}
    \caption{Per-language percentage accuracy delta relative to ByT5 tokenizer on multilingual addition for languages with small performance differences.}
    \label{fig:maddition_small_diff}
\end{figure}

We observe consistent gains on the remaining twelve languages in Figure~\ref{fig:maddition_small_diff} with \moto. In this setting, the disparity between \moto and mBPE is smaller as these languages have simpler numeral morphology than those in Figure~\ref{fig:maddition_big_diff}.

\subsection{Multilingual Pre-training}\label{ap:sec:full_results:pretraining}
We evaluate all models on a diverse set of multilingual and English downstream tasks suitable for base models. 
Not all languages are uniformly available across all benchmarks. Language coverage across all tasks is reported in Table~\ref{ap:tab:eval_coverage}.
For language modeling ability we report bits-per-byte on FLORES+ across all 21 languages and accuracy on MultiBLIMP~\citep{jumelet2026multiblimp} across 17 languages.
For \textbf{commonsense reasoning}, we use HellaSwag~\citep{zellers2019hellaswag}, PIQA~\citep{bisk2020piqa}, BoolQ~\citep{clark2019boolq}, ARC Easy and Challenge~\citep{clark2018arc}, and CommonsenseQA~\citep{talmor2019commonsenseqa} in English, alongside the machine-translated multilingual HellaSwag~\citep{lai2023okapi} across 13 languages, XCOPA~\citep{ponti2020xcopa,roemmele2011copa1} across 5 languages, and XStoryCloze~\citep{lin2022XGLM-XStoryClose} across 6 languages. 
For \textbf{natural language inference}, we evaluate on XNLI~\citep{conneau2018xnli} across 10 languages, PAWS-X~\citep{yang2019pawsx} across 6 languages, and XWinograd~\citep{muennighoff2022crosslingual,tikhonov2021heads} across 6 languages. 
For \textbf{knowledge and language understanding}, we evaluate on canonical completion tasks of TokSuite~\citep{toksuite} across 5 languages. We also evaluate on INCLUDE-base-44~\citep{romanou2024include}, Belebele~\citep{bandarkar2024belebele}, but find these benchmarks too challenging for 1B-parameter base models; yielding near random chance performance.
We report per-language results across all benchmarks relative to the Llama baseline. Unless otherwise noted, positive values indicate improvement over Llama.

\begin{table}[!htbp]
\centering
\tiny
\caption{Downstream task evaluation coverage per language. \checkmark indicates the language is evaluated on the corresponding benchmark group.\\
$\ast$: While Chinese subset is reported in \cite{lai2023okapi}, it is not available in lm-eval-harness~\citep{lm-eval-harness}.}
\label{ap:tab:eval_coverage}
\begin{tabular}{@{}llllllllll@{}}
\toprule
Language   & M-HellaSwag & XCOPA & XStoryCloze & PAWS-X & XNLI & TokSuite & Wino-X & Multi-BLIMP & Coverage \\ \midrule

Arabic     & $\checkmark$ &              & $\checkmark$ &              & $\checkmark$ &              &              & $\checkmark$ & 4                            \\
Chinese    & *            & $\checkmark$ & $\checkmark$ & $\checkmark$ & $\checkmark$ & $\checkmark$ & $\checkmark$ &              & 6                            \\
Czech      &              &              &              &              &              &              &              & $\checkmark$ & 1                            \\
Danish     & $\checkmark$ &              &              &              &              &              &              & $\checkmark$ & 2                            \\
Dutch      & $\checkmark$ &              &              &              &              &              &              & $\checkmark$ & 2                            \\
English    &              &              & $\checkmark$ & $\checkmark$ & $\checkmark$ & $\checkmark$ & $\checkmark$ & $\checkmark$ & 6                            \\
Farsi      &              &              &              &              &              & $\checkmark$ &              & $\checkmark$ & 2                            \\
French     & $\checkmark$ &              &              & $\checkmark$ & $\checkmark$ &              & $\checkmark$ & $\checkmark$ & 5                            \\
German     & $\checkmark$ &              &              & $\checkmark$ & $\checkmark$ &              &              & $\checkmark$ & 4                            \\
Greek      &              &              &              &              & $\checkmark$ &              &              & $\checkmark$ & 2                            \\
Hungarian  & $\checkmark$ &              &              &              &              &              &              & $\checkmark$ & 2                            \\
Indonesian & $\checkmark$ & $\checkmark$ & $\checkmark$ &              &              &              &              &              & 3                            \\
Italian    & $\checkmark$ & $\checkmark$ &              &              &              & $\checkmark$ &              & $\checkmark$ & 4                            \\
Japanese   &              &              &              & $\checkmark$ &              &              & $\checkmark$ &              & 2                            \\
Polish     &              &              &              &              &              &              &              & $\checkmark$ & 1                            \\
Portuguese & $\checkmark$ &              &              &              &              &              & $\checkmark$ & $\checkmark$ & 3                            \\
Russian    & $\checkmark$ &              & $\checkmark$ &              & $\checkmark$ &              & $\checkmark$ & $\checkmark$ & 5                            \\
Spanish    & $\checkmark$ &              & $\checkmark$ & $\checkmark$ & $\checkmark$ &              &              & $\checkmark$ & 5                            \\
Swedish    & $\checkmark$ &              &              &              &              &              &              & $\checkmark$ & 2                            \\
Turkish    &              & $\checkmark$ &              &              & $\checkmark$ & $\checkmark$ &              & $\checkmark$ & 4                            \\
Vietnamese & $\checkmark$ & $\checkmark$ &              &              & $\checkmark$ &              &              &              & 3                            \\
Total      & 13           & 5            & 6            & 6            & 10           & 5            & 6            & 17           &        
\\ \bottomrule
\end{tabular}
\end{table}

\paragraph{English performance}
\begin{table}
\caption{Relative accuracy difference (\%) on English benchmarks compared to Llama-3 tokenizer model. Higher better.}
\label{ap:tab:english_perf}
\tiny
\begin{tabular}{lllllllll}
\toprule
 & HellaSwag & ARC-Easy & ARC-Challenge & CommonsenseQA & PIQA & BoolQ & XWinograd & Avg \\
Model &  &  &  &  &  &  &  &  \\
\midrule
MoTo & $+0.30 \pm 0.31$ & $+0.13 \pm 0.78$ & $+0.09 \pm 1.19$ & $-3.03 \pm 1.66$ & $+0.87 \pm 0.73$ & $+4.38 \pm 1.02$ & $+1.08 \pm 0.88$ & $+0.54 \pm 0.38$ \\
MoTo-GPT4 & $+0.06 \pm 0.31$ & $-0.80 \pm 0.74$ & $+1.37 \pm 1.15$ & $-2.86 \pm 1.67$ & $-0.16 \pm 0.74$ & $+2.45 \pm 1.05$ & $+0.22 \pm 0.89$ & $+0.04 \pm 0.38$ \\
mBPE-GPT4 & $+1.33 \pm 0.31$ & $+0.72 \pm 0.74$ & $+0.17 \pm 1.19$ & $-3.68 \pm 1.25$ & $-0.16 \pm 0.75$ & $+0.80 \pm 0.87$ & $+1.60 \pm 0.86$ & $+0.11 \pm 0.34$ \\
mBPE-Boundless & $+1.34 \pm 0.31$ & $+2.23 \pm 0.77$ & $+1.88 \pm 1.18$ & $-3.11 \pm 1.64$ & $+0.11 \pm 0.73$ & $+1.29 \pm 0.99$ & $+2.33 \pm 0.84$ & $+0.86 \pm 0.38$ \\
Llama-BPE-Drop. & $-1.03 \pm 0.30$ & $+0.55 \pm 0.75$ & $+0.68 \pm 1.15$ & $-3.11 \pm 1.60$ & $-0.71 \pm 0.71$ & $+1.93 \pm 1.00$ & $-0.60 \pm 0.87$ & $-0.33 \pm 0.37$ \\
Parity-Aware BPE & $+1.39 \pm 0.31$ & $-0.29 \pm 0.74$ & $+0.68 \pm 1.18$ & $-1.63 \pm 1.61$ & $-0.33 \pm 0.76$ & $+2.24 \pm 0.97$ & $+0.52 \pm 0.88$ & $+0.37 \pm 0.38$ \\
\bottomrule
\end{tabular}
\end{table}

\begin{figure}[H]
    \centering
    \includegraphics[width=1.1\linewidth]{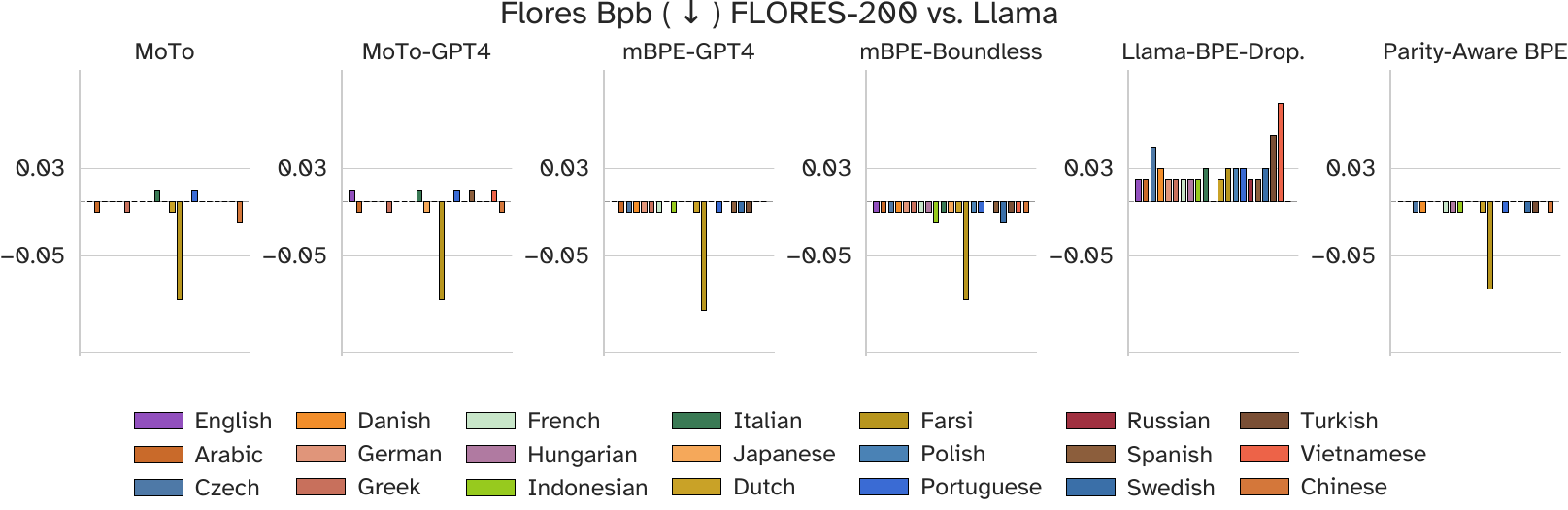}
    \includegraphics[width=1.1\linewidth]{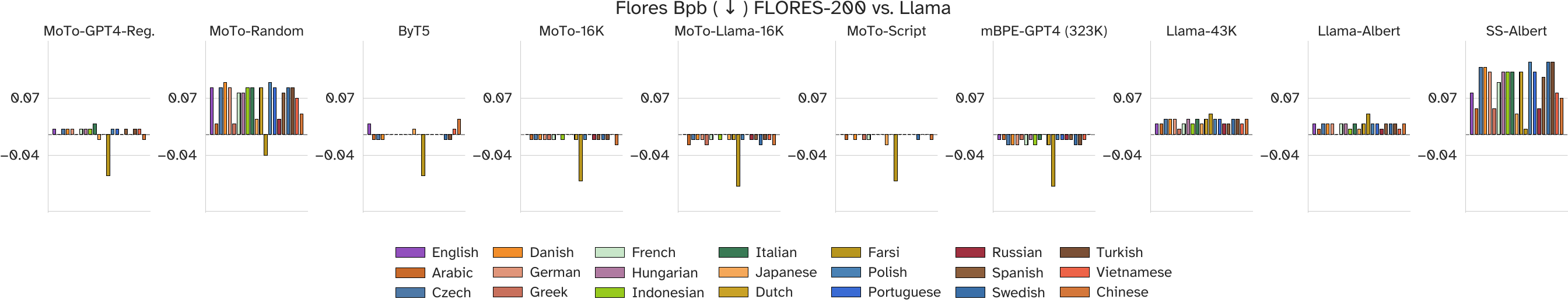}
    \caption{Per-language bits-per-byte (BPB) on FLORES+ relative to the Llama baseline (top: main configurations; bottom: ablations). Lower is better.}
    \label{fig:flores_main_bpb}
\end{figure}
\begin{figure}[h]
    \centering
    \includegraphics[width=1\linewidth]{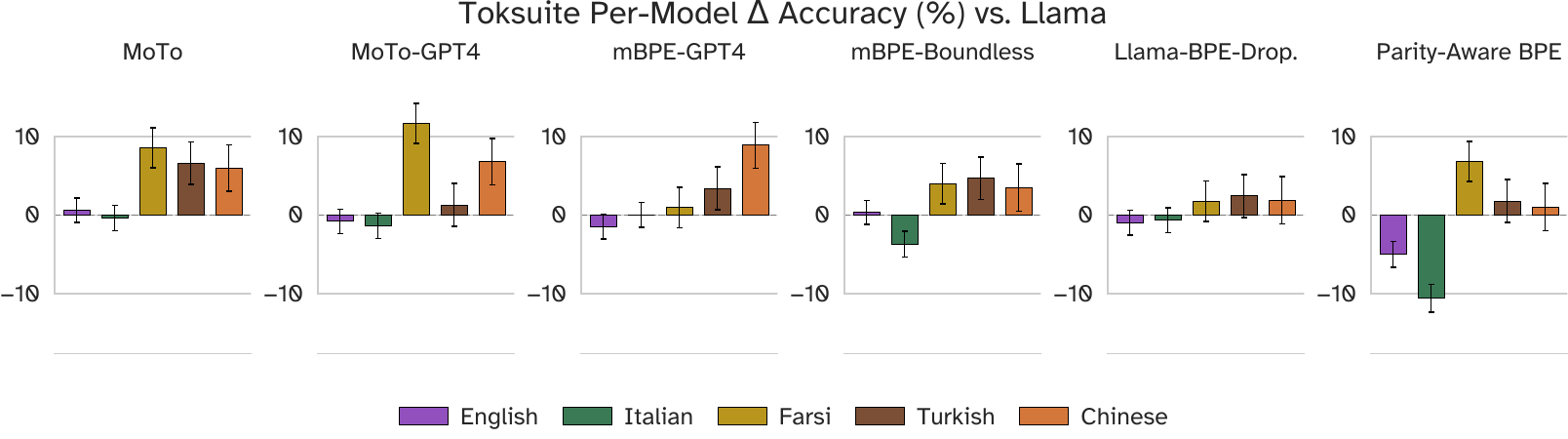}
    \caption{Ablation results on \toksuite completion tasks relative to the Llama baseline. 
    }
    \label{fig:toksuite}
\end{figure}
\begin{figure}[h]
    \centering
    \includegraphics[width=1\linewidth]{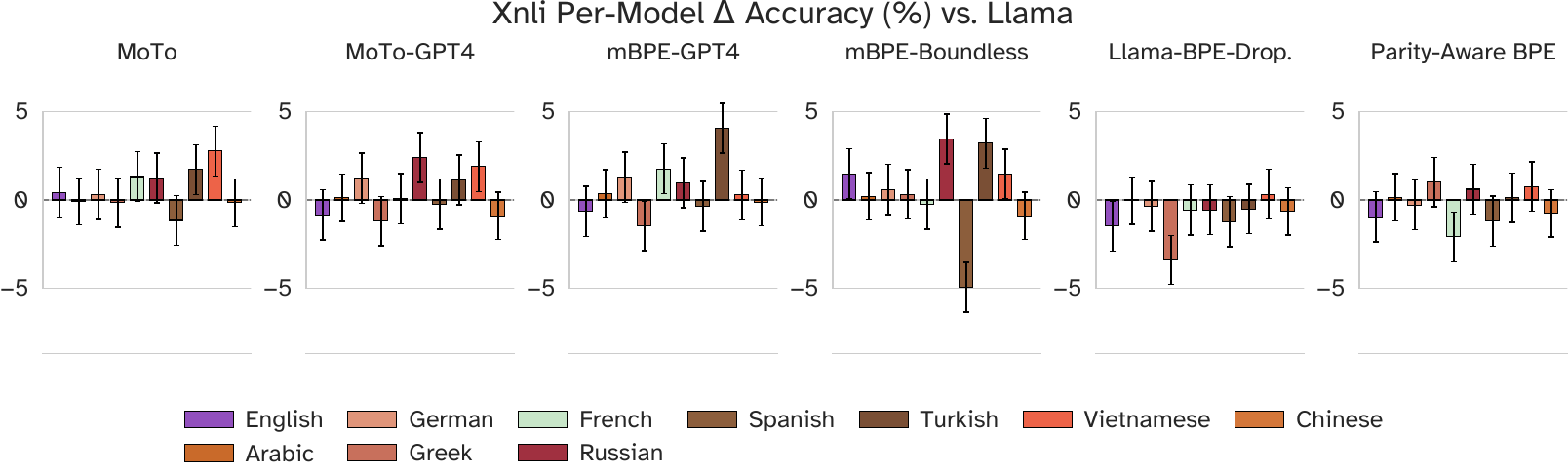}
    \caption{Ablation results on XNLI natural language inference (bottom), relative to the Llama baseline. 
    }
    \label{fig:xnli}
\end{figure}

\paragraph{Language modeling}
\emph{(FLORES+ bits-per-byte (BPB), 
Figure~\ref{fig:flores_main_bpb}).} In terms of bits-per-bytes, \moto settings and mBPE settings perform comparably to the Llama baselines with significant improvement in Farsi. We observe similar gains in Farsi in \toksuite (Figures~\ref{fig:toksuite} and~\ref{fig:xnli}) suggesting that Farsi is under-served by Llama tokenizer. With the distinction of BPE-Dropout baseline performing worse than Llama, all settings perform similarly.

\begin{figure}[H]
    \centering
    \includegraphics[width=1\linewidth]{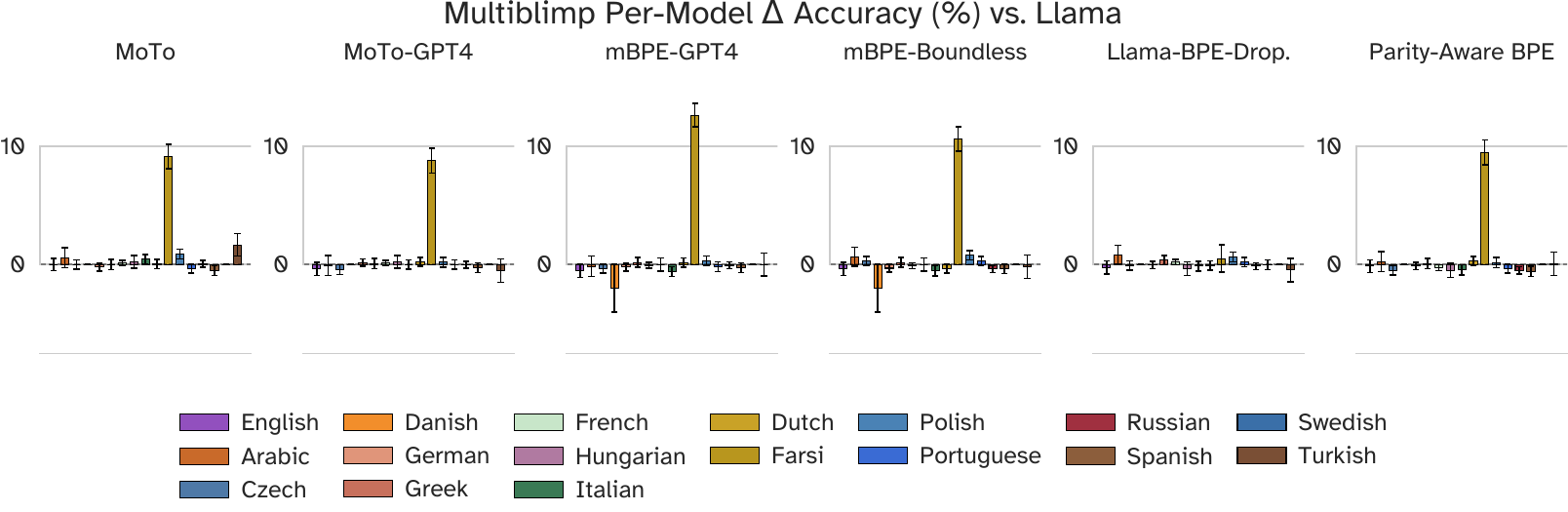}
    \caption{Per-language accuracy difference on MultiBLIMP~\citep{jumelet2026multiblimp} relative to the Llama baseline. 
    }
    \label{fig:multiblimp}
\end{figure}

\paragraph{Grammatical agreement (MultiBLiMP, Figure~\ref{fig:multiblimp}).}
MultiBLiMP~\citep{jumelet2026multiblimp} tests subject--verb agreement in person, number and gender with minimal pairs. A model is correct when it assigns the grammatical sentence a higher likelihood than its ungrammatical twin. We evaluate the 17 of our 21 languages that it covers. In general the absolute accuracies are saturated for MultiBLIMP.

\begin{figure}[H]
    \centering
    \includegraphics[width=1\linewidth]{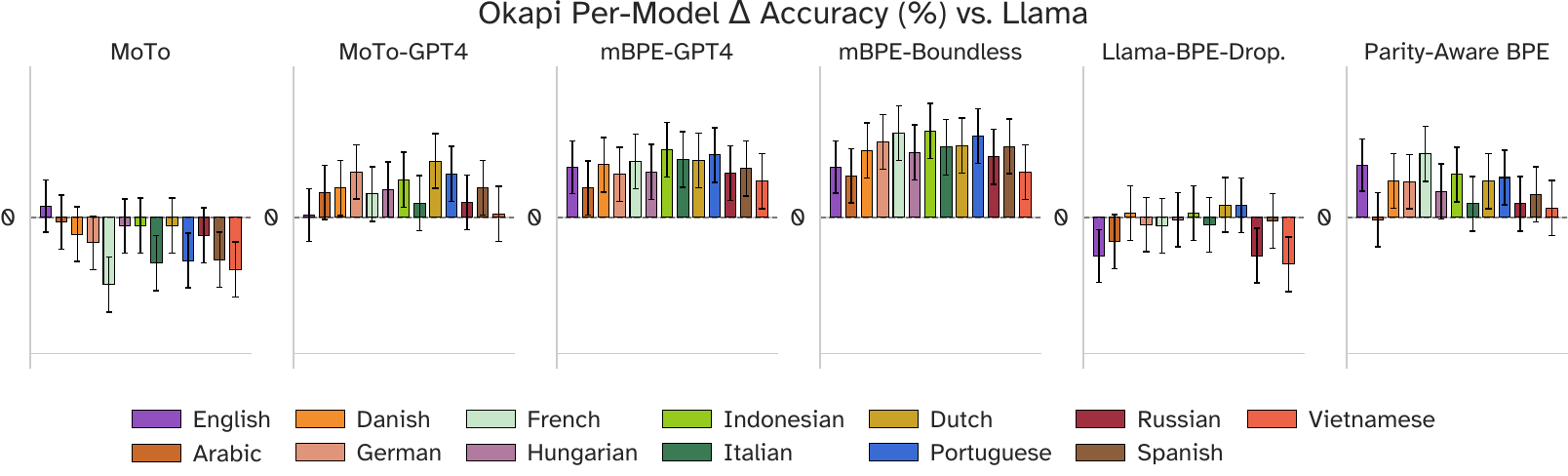}
    \caption{Per-language accuracy difference on multilingual HellaSwag (Okapi)~\citep{lai2023okapi} relative to the Llama baseline.
    }
    \label{fig:okapi}
\end{figure}

\paragraph{Multilingual commonsense reasoning 
(Okapi, Figure~\ref{fig:okapi}).} \moto-GPT4 shows gains across the board for all thirteen languages. Multilingual baselines perform consistently better with Boundless-BPE  pre-tokenization overperforming GPT4 especially in French, Indonesian, and Portuguese. 

\begin{figure}[H]
    \centering
    \includegraphics[width=1\linewidth]{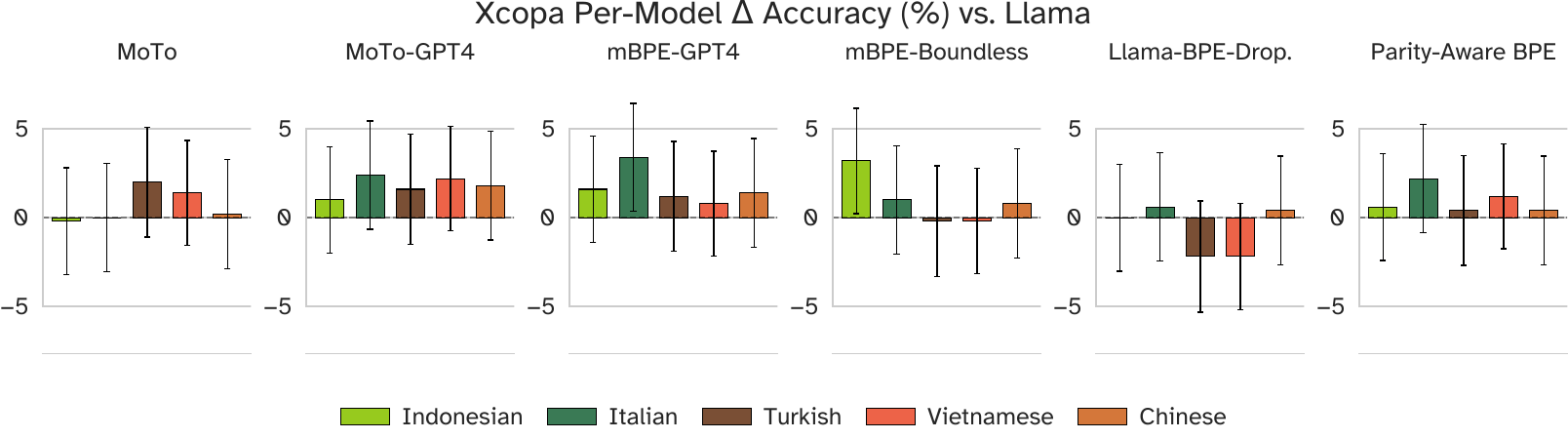}
    \caption{Per-language accuracy difference on XCOPA~\citep{ponti2020xcopa} causal commonsense reasoning benchmark relative to the Llama baseline.}
    \label{fig:xcopa}
\end{figure}

\paragraph{Causal commonsense reasoning 
(XCOPA, Figure~\ref{fig:xcopa}).} \moto variants outperform the Llama baseline and mBPE across all tasks, with mBPE showing larger improvements on Indonesian.

\begin{figure}[H]
    \centering
    \includegraphics[width=1\linewidth]{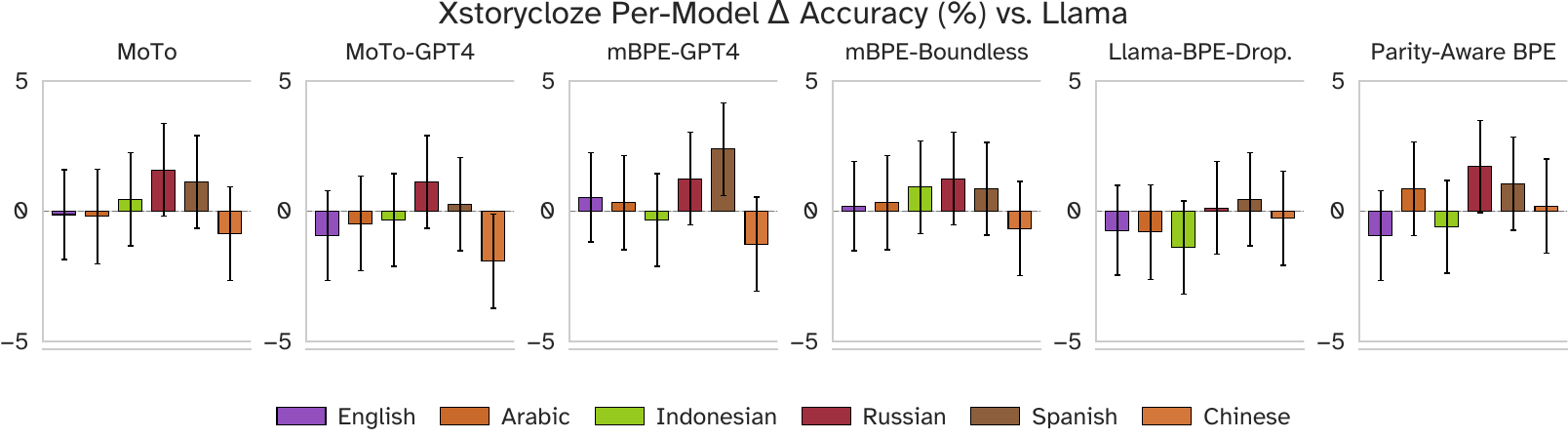}
    \caption{Per-language accuracy difference on XStoryCloze~\citep{lin2022XGLM-XStoryClose} relative to the Llama baseline. XStoryCloze tests narrative commonsense reasoning for story understanding.}
    \label{fig:xstorycloze}
\end{figure}

\paragraph{Narrative commonsense reasoning 
(XStoryCloze, Figure~\ref{fig:xstorycloze}).} \moto variants exhibit consistent gains on Russian, Spanish, and Indonesian. mBPE-GPT4 
achieves consistent gains on Spanish. Larger vocabulary variants (\moto-16K, \moto-Llama-16K) show additional benefits, suggesting vocabulary size matters for story understanding tasks.

\begin{figure}[H]
    \centering
    \includegraphics[width=1\linewidth]{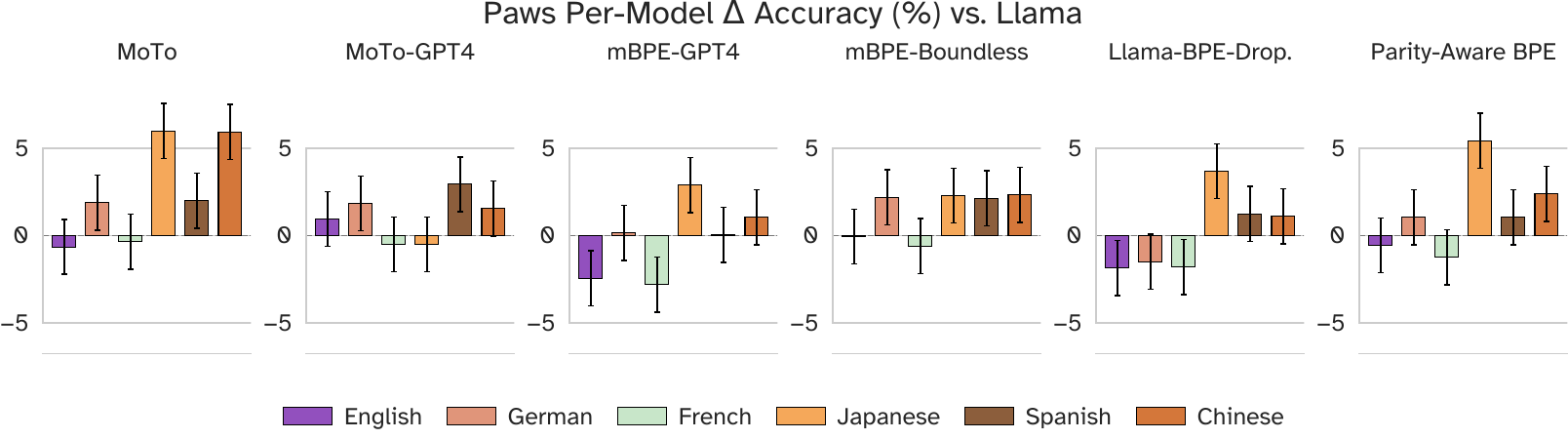}
    \caption{Per-language accuracy difference on PAWS-X~\citep{yang2019pawsx} relative to the Llama baseline. PAWS-X tests paraphrase identification.}
    \label{fig:paws}
\end{figure}

\paragraph{Paraphrase identification 
(PAWS-X, Figure~\ref{fig:paws}).} \moto models show substantial improvements in Japanese and Chinese, with modest gains on Spanish.

\begin{figure}[h]
    \centering
    \includegraphics[width=1\linewidth]{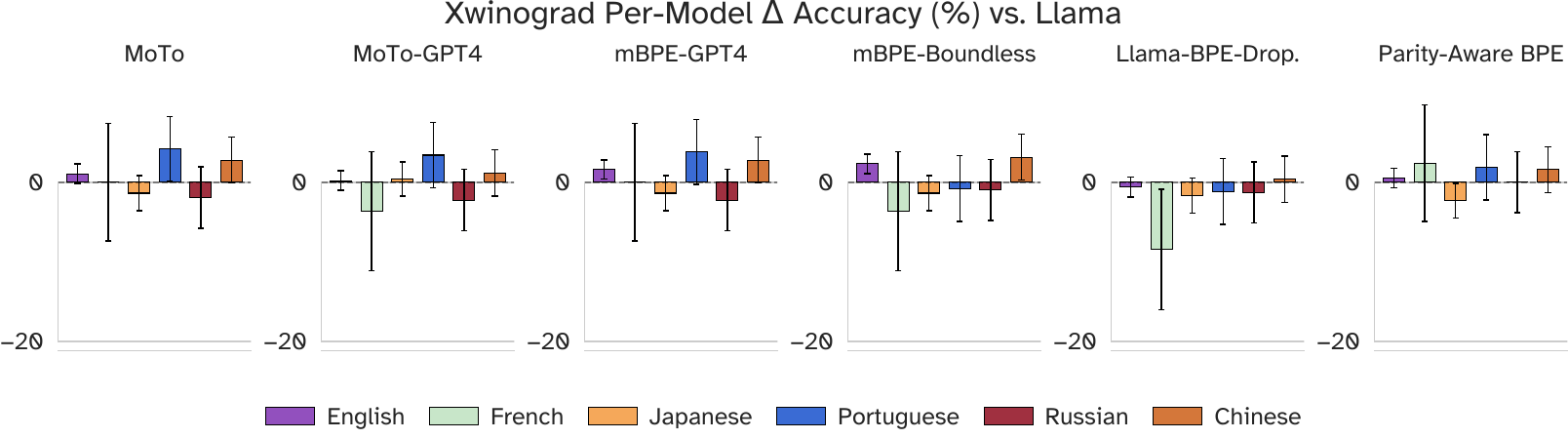}
    \caption{Per-language accuracy difference on XWinograd~\citep{muennighoff2022crosslingual, tikhonov2021heads} relative to the Llama baseline. 
     XWinograd evaluates pronoun resolution and coreference across multiple languages.
     }
    \label{fig:xwinograd}
\end{figure}

\paragraph{Coreference and pronoun resolution 
(XWinograd, Figure~\ref{fig:xwinograd}).} 
\moto{} variants 
exhibit large improvements in Portuguese, an observation we also make in HellaSwag (Figure~\ref{fig:okapi}). Chinese benefits from the modular design across all \moto variants, while improvements in Japanese are modest.
\paragraph{Knowledge probing and NLI 
(\toksuite and XNLI, Figures~\ref{fig:toksuite} and~\ref{fig:xnli}).} 
On \toksuite canonical completion tasks, Farsi shows the most consistent improvement across all \moto variants, reflecting Llama's systematic underservice of Arabic-script 
languages. Chinese and Turkish also show consistent gains, with small losses on Italian. On XNLI, Greek, Turkish, and Russian improve with \moto-16K and 
\moto-Reg.; however, Greek is hurt under script-level grouping (\moto-Script) even though both mixtures utilize the same underlying Greek tokenizer.

\section{Routing\label{ap:sec:routing}}
In \moto each input (a training document or an evaluation prompt) is routed to the tokenizer of its language, which we call LID routing. During pre-training, we take the language from the per-document labels that most multilingual corpora already provide, so routing adds no compute overhead to the training loop. At evaluation time, we use the benchmark's language when it is known, and otherwise predict it with GlotLID~\citep{kargaran2023glotlid}, restricted to the 21 languages in the mixture.
While this is well-defined in the pre-training regime we study, where each tokenizer is trained on a single language and documents come with language labels and each benchmark has language subsets, it raises three questions: what is the additional cost of routing when labels are unavailable, how it handles mixed-language inputs, and what happens when no tokenizer is a clear match for the input.

\begin{figure}[!htbp]
    \centering
    \includegraphics[width=0.8\linewidth]{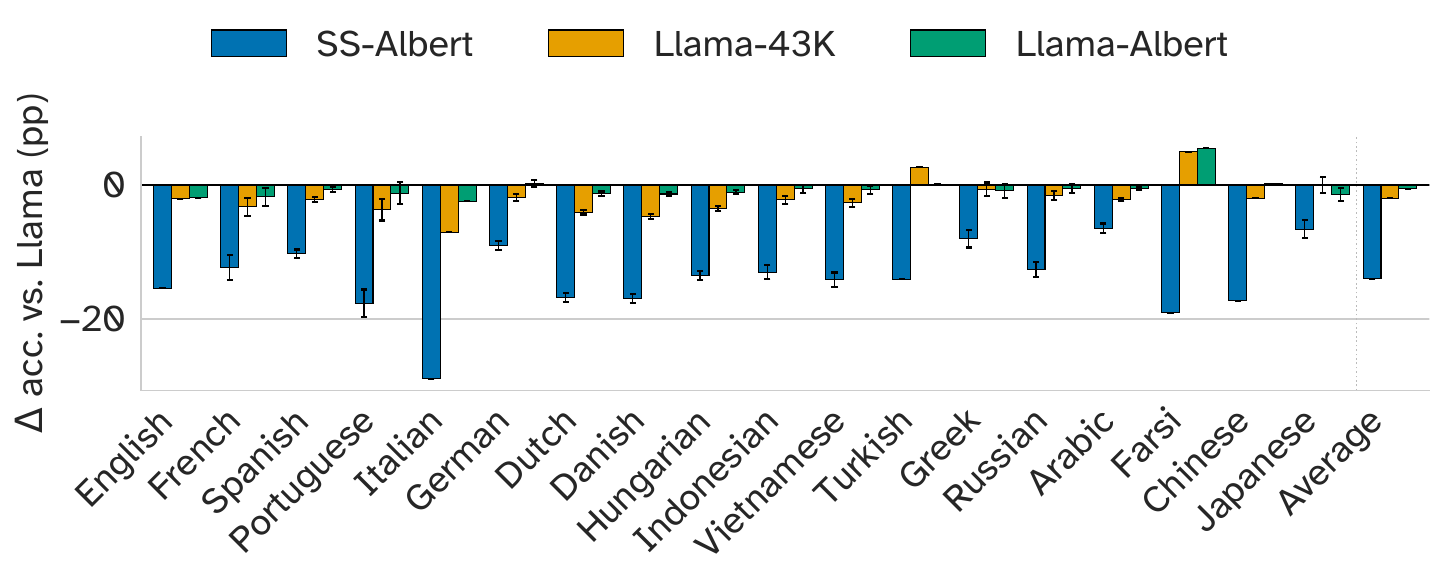}
    \caption{\textbf{Mixing existing tokenizers.} Per-language accuracy difference (pp) from the Llama~3.2 baseline, averaged over each language's tasks; the rightmost group is the average over languages. SS-Albert maps 14 existing tokenizers into one vocabulary with factorized embeddings. Llama-Albert uses the same factorized embeddings with the Llama tokenizer only. Llama-43K is the Llama baseline after 43K of its 100K training steps.}
    \label{fig:industry_per_lang}
\end{figure}

\paragraph{Mixing existing tokenizers.}
\moto{} trains its own per-language tokenizers. A natural alternative is to reuse tokenizers that already exist. Superset-Albert (SS-Albert) maps 14 off-the-shelf tokenizers from \toksuite~\citep{toksuite} into one 851K-entry vocabulary and encodes every training document with one of them, drawn at random: Aya Expanse, BLOOM, Comma, Gemma~2, GPT-2, Llama~3.2, mBERT, Phi-3, Qwen3, XGLM, ByT5, GPT-4o, Tekken and TokenMonster. A full-width table for this vocabulary would dominate the parameter count, so SS-Albert factorizes its input and output embeddings ALBERT-style ($851\text{K} \times 512 \to 2{,}048$). We compare it with two Llama~3.2 controls (Figure~\ref{fig:industry_per_lang}): Llama-Albert, with the same factorized embeddings and the Llama tokenizer alone, which isolates the effect of the tokenizer mixture from that of the factorization; and Llama-43K, the Llama baseline stopped after 43K of its 100K steps, which is the estimated number of steps embeddings for the Llama tokenizer are trained as part of the SS-Albert setting. We observe that all settings result in degradation and discuss implications for SS-Albert further below.

\paragraph{Language-ID (LID) routing has minimal overhead.} For unlabeled corpora, language identification would have to run alongside training, but its cost is negligible compared to that of training. For example, GlotLID is a fastText classifier~\citep{joulin2016fasttext} is a linear model over hashed character and word $n$-gram embeddings that runs on CPU. fastText~\citep{joulin2017bag} classifies half a million sentences among 312K classes in under a minute on a standard multicore CPU, and GlotLID applies the same architecture to over 1,600 languages~\citep{kargaran2023glotlid}. Language identification can therefore run by data-loader workers, in parallel with GPU training and at a negligible fraction of the cost of a forward and backward pass. Restricting predictions to the languages in the mixture, as we do, further simplifies this procedure. 

\begin{figure}[!htbp]
    \centering
    \includegraphics[width=0.8\linewidth]{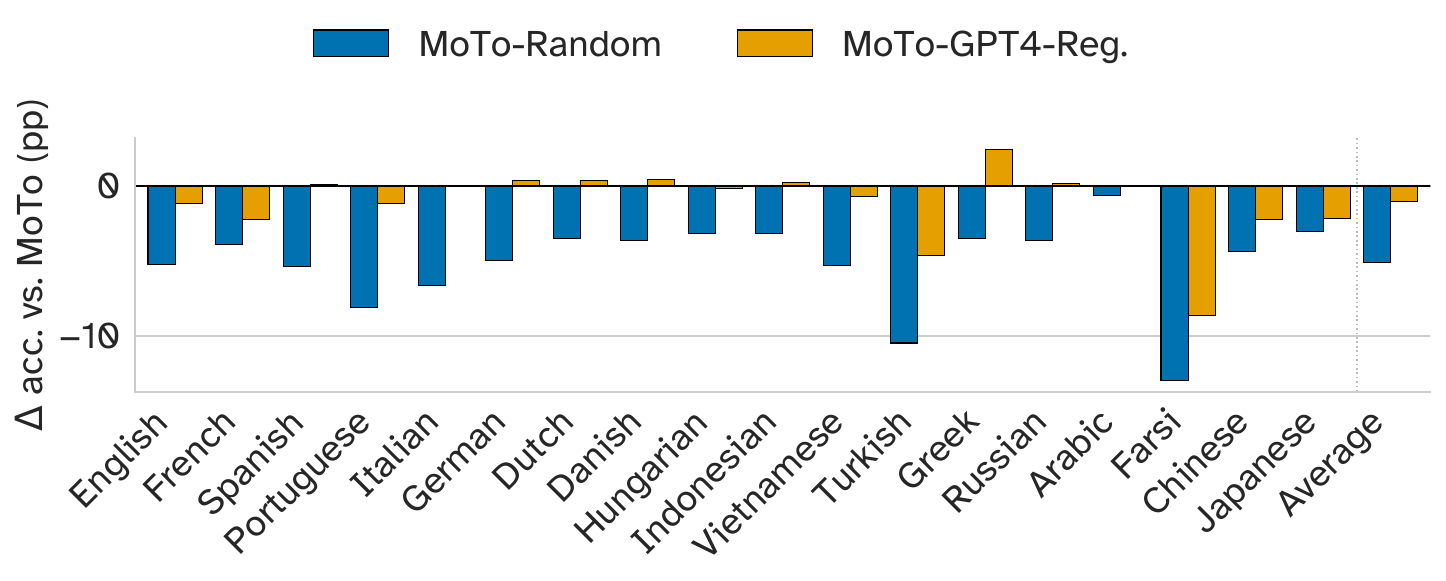}
    \caption{\textbf{Routing ablation.} Per-language accuracy difference (pp) from \moto{}, averaged over each language's tasks; the rightmost group is the average over languages. \moto{}-GPT4-Reg.\ routes each training document to its language's tokenizer with probability 0.9 and to a random component otherwise; \moto{}-Random always routes at random. All models are evaluated with oracle routing.}
    \label{fig:routing_per_lang}
\end{figure}

\paragraph{Regularizing LID routing}
Oracle routing assumes that every document's language is known. To test whether \moto{} depends on this, we train with regularized routing: each document is routed to its language's tokenizer with probability 0.9 and to a uniformly random component otherwise (\moto{}-GPT4-Reg.). We also train a fully random-routing model as a lower bound (\moto{}-Random). We evaluate both settings with oracle routing. Figure~\ref{fig:routing_per_lang} reports each model's accuracy relative to \moto{}. Regularization costs little: 1.1 points on average, and it matches or slightly exceeds \moto{} on 10 of 18 languages. Random routing costs 5.2 points, with the largest drops in Farsi ($-12.9$) and Turkish ($-10.5$), languages whose tokenizers are among the most specialized in the mixture. The model therefore benefits from seeing each language in its own segmentation most of the time, but not from seeing it exclusively.

\paragraph{Limitations of LID routing.} Document-level LID assigns one tokenizer per input, so code-switched or transliterated text could suffer in performance; or short documents could be misrouted to the tokenizer of closely related languages (e.g., the Farsi--Arabic or Spanish--Italian confusions common in short inputs). Languages outside the mixture are routed randomly or to a fallback tokenizer. Routing at a finer granularity or learning a router could address these cases.

\emph{Code-switching.} Assigning one language per input raises the question of how to route mixed-language text. We evaluate two code-switched settings: \toksuite~\citep{toksuite} examples in Chinese, Farsi, Italian, and Turkish with English mixed in, and CodeMixBench~\citep{yang2025codemixbench}, covering Spanish--English sentiment analysis and TruthfulQA with English mixed into Arabic, Chinese, and Spanish. For each input, we compare routing to the non-English tokenizer, routing to the English tokenizer, and GlotLID-based routing to the most probable tokenizer in the mixture. On \toksuite (Table~\ref{ap:tab:codeswitch:toksuite}), GlotLID routing matches or exceeds the better of the two fixed choices for every language.
On CodeMixBench (Table~\ref{ap:tab:codeswitch:codemixbench}), it stays between the two fixed choices, although the differences are small and TruthfulQA accuracy is close to chance for all three. 
\begin{observation}[motoDarkPurple]\label{obs:app:langID}
    Language-ID routing therefore does not degrade on code-switched inputs, but finer-grained (e.g., span-level) routing may help where the two tokenizers diverge more.
\end{observation}

\begin{table}[!htbp]
\centering
\caption{Accuracy (\%) of \moto on code-switched \toksuite examples (non-English with English mixed in) when routing everything to the non-English tokenizer (Tok.~A), to the English tokenizer (Tok.~B), or via GlotLID. Best per row in bold.}
\label{ap:tab:codeswitch:toksuite}
\small
\begin{tabular}{@{}lccc@{}}
\toprule
Lang.~A & Tok.~A & Tok.~B (English) & GlotLID \\
\midrule
Chinese & \textbf{85} & 43 & \textbf{85} \\
Farsi   & 70 & 33 & \textbf{72} \\
Italian & \textbf{87} & 70 & \textbf{87} \\
Turkish & \textbf{72} & 40 & \textbf{72} \\
\bottomrule
\end{tabular}
\end{table}

\begin{table}[!htbp]
\centering
\caption{Accuracy (\%) of \moto on CodeMixBench when routing everything to the non-English tokenizer (Tok.~A), to the English tokenizer (Tok.~B), or via GlotLID. Best per row in bold.}
\label{ap:tab:codeswitch:codemixbench}
\small
\begin{tabular}{@{}llccc@{}}
\toprule
Task & Lang.~A & Tok.~A & Tok.~B (English) & GlotLID \\
\midrule
Sentiment (3-way)      & Spanish & 44 & \textbf{49} & 45 \\
\midrule
TruthfulQA (4--11-way) & Arabic  & \textbf{22} & \textbf{22} & \textbf{22} \\
                       & Chinese & 22 & \textbf{23} & \textbf{23} \\
                       & Spanish & 21 & \textbf{24} & 22 \\
\bottomrule
\end{tabular}
\end{table}

\paragraph{Without a clear target tokenizer.} \label{ap:sec:routing:albert}
Superset-Albert, an illustrative baseline rather than our main method, combines fourteen nominally multilingual off-the-shelf tokenizers with no well-defined language-to-tokenizer mapping, so we train and evaluate it with random routing. 
To estimate the headroom for better routing, Table~\ref{ap:tab:ss_albert:routing}
reports oracle accuracy, where an example counts as correct if any of the fourteen tokenizers yields the correct answer. The gap between random and oracle routing is large on every benchmark, reaching 47.36 points on PAWS-X.

\emph{Deconstructing the oracle gap.} The oracle in Table~\ref{ap:tab:ss_albert:routing} is an optimistic noisy upper bound, because choosing the best of fourteen predictions after the fact also rewards chance agreement. PAWS-X shows this most clearly: individual tokenizers are near chance (51.25\%), and fourteen independent guesses on a binary task would include a correct answer for $1 - 0.5^{14} > 99.9\%$ of examples, close to the observed 97.3\%. To separate signal from noise, we group examples by how many of the fourteen tokenizers answer them correctly. Pooled over all benchmarks, random routing (bootstrapped over 10,000 trials) reaches $38.94 \pm 0.11\%$. $43.15 \pm 0.23\%$ of examples are answered correctly by at least seven tokenizers, and $59.59 \pm 0.22\%$ by at least one. The examples answered correctly by many tokenizers show that routing has real headroom. The rest of the gap up to the oracle includes a large chance component that no realistic router could exploit.

\begin{table}
\caption{Percentage of examples answered correctly by at least $n$ of the 14 tokenizers in Superset-Albert. Random routing reaches $38.94 \pm 0.11\%$ accuracy over all benchmarks combined. $\pm$ denotes 95\% bootstrap confidence intervals over 10,000 resamples.}
\label{ap:tab:ss_albert:routing}
\centering
\footnotesize
\begin{tabular}{llrrrrr}
\toprule
Group & Task & $N$ & $\geq$1 tok.\ & $\geq$7 tok.\ & All toks.\ & No tok.\ \\
\midrule
  English Reasoning    & arc\_easy                    &   2376 &  81.1\% &  63.9\% &  12.2\% &  18.9\% \\
                       & arc\_challenge               &   1172 &  55.3\% &  31.4\% &   5.2\% &  44.7\% \\
                       & hellaswag                   &  10042 &  50.8\% &  38.8\% &  17.3\% &  49.2\% \\
                       & piqa                        &   1838 &  85.9\% &  70.6\% &  34.5\% &  14.1\% \\
                       & boolq                       &   3270 &  94.2\% &  59.3\% &   5.6\% &   5.8\% \\
                       & commonsense\_qa              &   1221 &  55.1\% &  18.2\% &   0.9\% &  44.9\% \\
 \cline{4-7}\noalign{\vskip 2pt}&&&  70.4\% &  47.0\% &  12.6\% &  29.6\% \\[2pt]
\midrule
  XWinograd            & en                          &   2325 &  94.9\% &  73.0\% &  26.8\% &   5.1\% \\
                       & fr                          &     83 &  89.2\% &  60.2\% &  15.7\% &  10.8\% \\
                       & ja                          &    959 &  98.2\% &  65.4\% &   6.7\% &   1.8\% \\
                       & pt                          &    263 &  95.1\% &  64.6\% &  15.2\% &   4.9\% \\
                       & ru                          &    315 &  94.6\% &  61.3\% &  12.1\% &   5.4\% \\
                       & zh                          &    504 &  98.2\% &  72.2\% &   9.3\% &   1.8\% \\
 \cline{4-7}\noalign{\vskip 2pt}&&&  95.0\% &  66.1\% &  14.3\% &   5.0\% \\[2pt]
\midrule
  XCOPA                & id                          &    500 &  85.8\% &  64.0\% &  27.8\% &  14.2\% \\
                       & it                          &    500 &  85.0\% &  59.0\% &  25.2\% &  15.0\% \\
                       & tr                          &    500 &  83.2\% &  58.2\% &  25.4\% &  16.8\% \\
                       & vi                          &    500 &  84.8\% &  64.2\% &  28.8\% &  15.2\% \\
                       & zh                          &    500 &  91.8\% &  59.8\% &  12.8\% &   8.2\% \\
 \cline{4-7}\noalign{\vskip 2pt}&&&  86.1\% &  61.0\% &  24.0\% &  13.9\% \\[2pt]
\midrule
  XStoryCloze          & ar                          &   1511 &  75.8\% &  55.1\% &  26.5\% &  24.2\% \\
                       & en                          &   1511 &  78.8\% &  64.1\% &  31.3\% &  21.2\% \\
                       & es                          &   1511 &  76.8\% &  58.8\% &  28.9\% &  23.2\% \\
                       & id                          &   1511 &  76.3\% &  59.2\% &  29.7\% &  23.7\% \\
                       & ru                          &   1511 &  77.5\% &  58.6\% &  28.8\% &  22.5\% \\
                       & zh                          &   1511 &  86.6\% &  56.3\% &  15.8\% &  13.4\% \\
 \cline{4-7}\noalign{\vskip 2pt}&&&  78.6\% &  58.7\% &  26.8\% &  21.4\% \\[2pt]
\midrule
  PAWS-X               & de                          &   2000 &  96.8\% &  57.2\% &   5.2\% &   3.1\% \\
                       & en                          &   2000 &  97.7\% &  55.6\% &   6.3\% &   2.3\% \\
                       & es                          &   2000 &  90.8\% &  58.6\% &  14.6\% &   9.2\% \\
                       & fr                          &   2000 &  98.5\% &  52.9\% &   2.6\% &   1.4\% \\
                       & ja                          &   2000 & 100.0\% &  50.6\% &   0.0\% &   0.0\% \\
                       & zh                          &   2000 & 100.0\% &  52.4\% &   0.0\% &   0.0\% \\
 \cline{4-7}\noalign{\vskip 2pt}&&&  97.3\% &  54.6\% &   4.8\% &   2.7\% \\[2pt]
\midrule
  HellaSwag-X          & ar                          &   9176 &  46.9\% &  31.6\% &  12.3\% &  53.1\% \\
                       & da                          &   9305 &  49.2\% &  34.3\% &  14.7\% &  50.8\% \\
                       & de                          &   9368 &  48.7\% &  33.6\% &  14.8\% &  51.3\% \\
                       & fr                          &   9338 &  49.0\% &  35.1\% &  15.1\% &  51.0\% \\
                       & hu                          &   9119 &  45.8\% &  31.7\% &  14.2\% &  54.2\% \\
                       & id                          &   9314 &  48.4\% &  33.6\% &  15.2\% &  51.6\% \\
                       & it                          &   9193 &  48.8\% &  34.1\% &  15.1\% &  51.2\% \\
                       & nl                          &   9265 &  49.0\% &  34.0\% &  14.2\% &  51.0\% \\
                       & pt                          &   9229 &  49.8\% &  35.0\% &  14.9\% &  50.2\% \\
                       & ru                          &   9272 &  48.7\% &  33.8\% &  13.4\% &  51.3\% \\
                       & vi                          &   9162 &  46.2\% &  33.8\% &  15.7\% &  53.8\% \\
                       & es                          &   9374 &  50.5\% &  36.4\% &  15.0\% &  49.5\% \\
 \cline{4-7}\noalign{\vskip 2pt}&&&  48.4\% &  33.9\% &  14.5\% &  51.6\% \\[2pt]
\midrule
  XNLI                 & ar                          &   2490 &  51.6\% &  33.7\% &  15.7\% &  48.4\% \\
                       & zh                          &   2490 &  70.1\% &  33.9\% &   1.2\% &  29.9\% \\
                       & de                          &   2490 &  64.0\% &  47.3\% &  13.6\% &  36.0\% \\
                       & el                          &   2490 &  93.2\% &  40.3\% &   0.2\% &   6.8\% \\
                       & en                          &   2490 &  76.3\% &  52.0\% &  10.0\% &  23.7\% \\
                       & es                          &   2490 &  65.2\% &  47.6\% &   6.6\% &  34.8\% \\
                       & fr                          &   2490 &  67.1\% &  48.4\% &   9.8\% &  32.9\% \\
                       & ru                          &   2490 &  66.7\% &  44.5\% &  15.3\% &  33.3\% \\
                       & tr                          &   2490 &  72.5\% &  46.2\% &   4.9\% &  27.5\% \\
                       & vi                          &   2490 &  86.2\% &  46.8\% &   0.2\% &  13.8\% \\
 \cline{4-7}\noalign{\vskip 2pt}&&&  71.3\% &  44.1\% &   7.8\% &  28.7\% \\[2pt]
\bottomrule

\end{tabular}

\end{table}

\paragraph{Learned routing.} Instead of relying on language ID, the tokenizer could be chosen by a router trained end-to-end with the language model. Since this choice is made before tokenization, the router needs its own representation of the raw input, e.g., a lightweight byte-level encoder like those used in byte-level and latent transformers~\citep{pagnoni2025BLT}. In preliminary experiments on the synthetic multilingual addition task (\S\ref{sec:maddition}), such a router proved difficult to train, and we leave learned routing to future work.

\clearpage
\section{Merging Per-Language Tokenizers into a Monolithic BPE Tokenizer}
\label{app:moto-merge}

Given $N$ language-specific BPE tokenizers $\{T_1, \dots, T_N\}$, each defined by a corresponding vocabulary $V_i$, an ordered merge table $M_i$, and a pretokenizer $P_i$, we describe a procedure to produce a single monolithic BPE tokenizer $T^*$ with
vocabulary $V^*$, merge table $M^*$, and pretokenizer $P^*$.

\paragraph{Vocabulary} The superset vocabulary described in Figure~\ref{fig:superset:pipeline} serves directly as the vocabulary of the monolithic tokenizer, where the merge priority is determined heuristically below.

\paragraph{Pre-tokenization}
In general, constituent tokenizers may have divergent pre-tokenization rules, causing the same input string to be split into different pre-tokens across tokenizers
This matters because a merge $(a, b) \to ab$ is only valid if $a$ and $b$ can
appear as adjacent substrings within a single pre-token; a merge that is
reachable under one pre-tokenizer may be unreachable under another.
When pre-tokenizers diverge, one must either (1) select a canonical $P^*$ common to all constituent tokenizers and prune all merges in $M^*$ that are unreachable under $P^*$, propagating pruning through the dependency DAG, 
or (2) retain per-language pre-tokenization via a language-routing layer.

We adopt option (1) and set $P^*$ to the shared canonical configuration.
Pre-tokenization rules help manage the computational complexity of tokenizer training by restricting the space of candidate merges to within-pretoken substrings.
Since our method is post-hoc, we are not concerned by this; the relevant cost instead might arise slightly at inference time, where tokenization greedily applies $M^*$ within each pre-token. Because $P^*$ uses a shared, coarser boundary rule than the script-specific pre-tokenizers of the constituent tokenizers, pre-tokens in $T^*$ can span longer substrings, increasing the number of candidate merges considered per pre-token.

In our setting, all constituent tokenizers share identical
punctuation, whitespace, number, and byte-level pre-tokenization rules, differing only
in script-specific word-boundary patterns.
Since no constituent tokenizer learns merges that cross punctuation or
whitespace boundaries, the script-specific rules can be dropped without
affecting reachability of any merge in $M^*$.

\begin{wrapfigure}{l}{0.55\linewidth}
    \includegraphics[width=\linewidth]{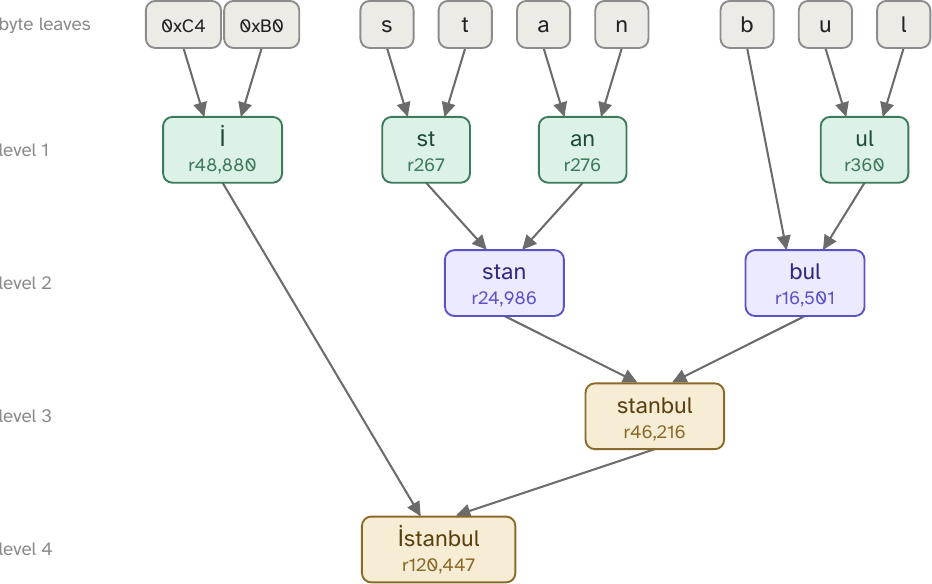}
    \captionof{figure}{BPE merge tree for \.{I}stanbul in the Llama-3.2 tokenizer. Leaves are UTF-8 bytes; each internal node is a merged token annotated with its merge rank (r).}
    \label{fig:istanbul-merges}
\end{wrapfigure}
\paragraph{Merge table construction via DAG union.}
Each merge table $M_i$ induces a dependency DAG $G_i$ whose nodes are tokens.
A merge $(a, b) \to ab$ has incoming edges $a \to ab$ and $b \to ab$; base tokens (UTF-8 bytes in most modern tokenizers) have no incoming edges and form the leaves (Figure~\ref{fig:istanbul-merges}).
The union DAG $G^* = \bigcup_i G_i$ takes the union of all nodes and edges, identifying nodes by their token string.
Because different tables may build the same token from different pairs, a node in $G^*$ can have more than one incoming merge.
 
Conflicts arise when two tokenizers $T_i$ and $T_j$ both contain merges $m_1$ and $m_2$ but disagree on their relative order, introducing a cycle in $G^*$.
We distinguish two conflict types: \emph{dependency conflicts}, where two tokenizers' dependency chains directly contradict one another (one requires $m_1$ before $m_2$ via a chain of structural dependencies, while another  requires the reverse), and \emph{ordering conflicts}, where $m_1$ and $m_2$ are independent in both DAGs but their relative rank is reversed.

\begin{figure}[!htbp]
    \centering
    \begin{subfigure}{0.48\textwidth}
        \includegraphics[width=\linewidth]{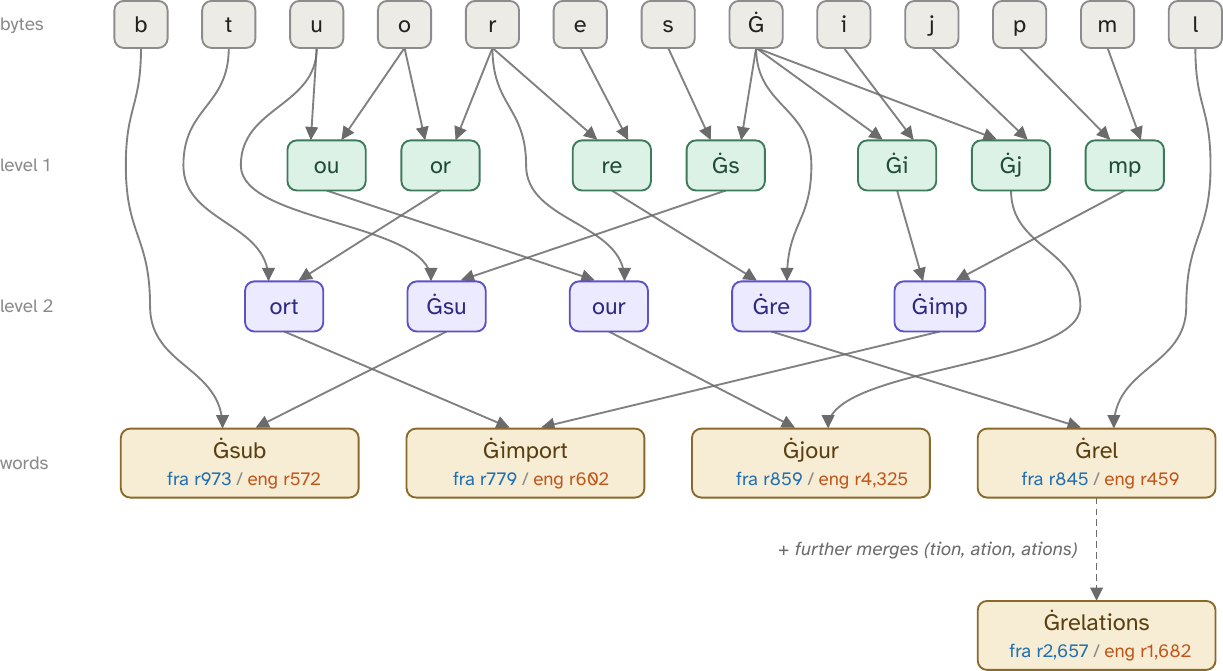}
    \end{subfigure}\hfill
    \begin{subfigure}{0.48\textwidth}
        \includegraphics[width=\linewidth]{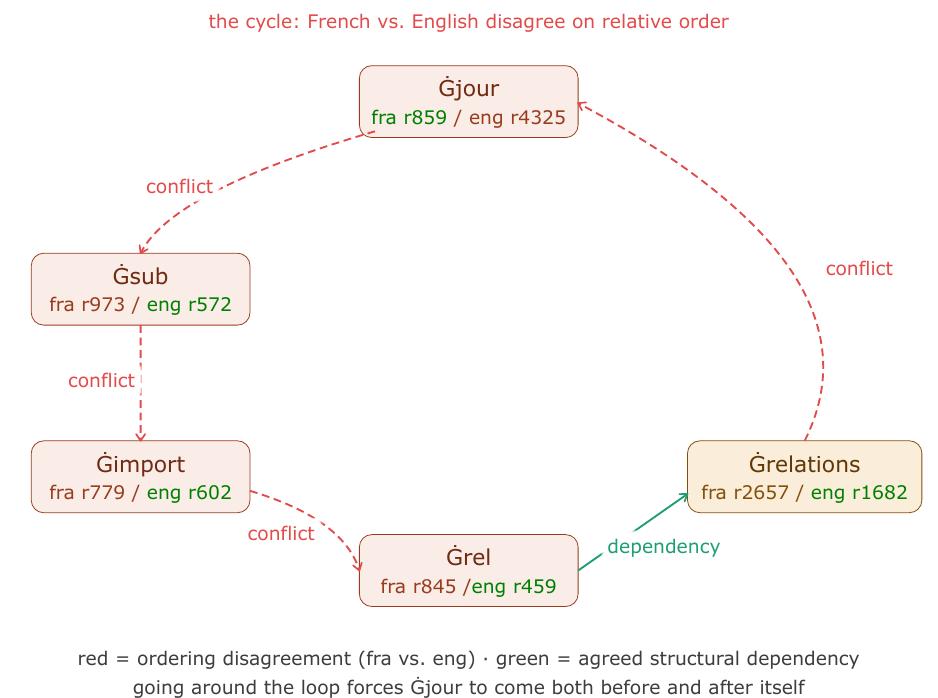}
    \end{subfigure}
    \caption{A real ordering-conflict cycle found between the French and English constituent tokenizers. \textbf{Left:} dependency tree from byte leaves to the five merges involved in the cycle, with both tokenizers' ranks shown at every node (\textit{fra r}\,/\,\textit{eng r}). \textbf{Right:} the cycle isolated. Four edges (red, dashed) are ordering conflicts where French and English disagree on relative rank; one edge (green, solid), $\texttt{Ġrel} \to \texttt{Ġrelations}$, is a structural dependency both tokenizers agree on. No linear order of these five merges satisfies both tokenizers simultaneously, motivating the rank resolution policy.}
    \label{fig:cycle-example}
\end{figure}

Note that dependency conflicts cannot occur for shared merges: if $m_2$ is a
structural ancestor of $m_1$ in $G_i$ (i.e.\ $m_1$'s output string contains
$m_2$'s output string as a constituent produced by an earlier merge), this
ancestry is a property of the token strings themselves, not of which
tokenizer learned it, any tokenizer containing both merges must agree on the
direction, since string concatenation is not reversible.
All cycles in $G^*$ are therefore necessarily caused by ordering conflicts,
which reduce to a rank resolution policy; a resulting cycle may still pass
through an uncontested dependency edge alongside one or more conflicting
edges, as we observe empirically (Figure~\ref{fig:cycle-example}).

Ordering conflicts are resolved by \textbf{majority vote}: for each conflicting pair $(m_1, m_2)$, we retain the ordering agreed upon by the majority of tokenizers containing both merges.
For pairs appearing in exactly two tokenizers, majority vote degenerates and we fall back to \textbf{min-rank}: we retain the ordering from whichever tokenizer assigns the lower absolute rank to either merge.
The resolved $G^*$ is acyclic by construction; we obtain $M^*$ by topological sort, breaking ties by average normalized rank across all tokenizers containing each merge.

\paragraph{A real example of ordering conflicts}

Figure~\ref{fig:cycle-example} (left) shows an ordering conflict found between the French and English constituent tokenizers.
The English tokenizer ranks \texttt{Ġsub}, \texttt{Ġimport}, and \texttt{Ġrel} before \texttt{Ġjour}, while the French tokenizer ranks \texttt{Ġjour} before all three.
Since \texttt{Ġrelations} depends structurally on \texttt{Ġrel} (both tokenizers agree that \texttt{Ġrel} must precede \texttt{Ġrelations}, as the latter is built by appending \texttt{ations}), the four ordering
disagreements together with this one dependency edge close a five-node cycle (Figure~\ref{fig:cycle-example}, right):
$\texttt{Ġjour} \to \texttt{Ġsub} \to \texttt{Ġimport} \to \texttt{Ġrel} \to
\texttt{Ġrelations} \to \texttt{Ġjour}$.
No single linear order of these five merges is consistent with both tokenizers, illustrating why a direct concatenation or interleaving of merge tables is insufficient and motivating the rank resolution policy.

\paragraph{Vocabulary-composition effects.}
\label{par:vocab-composition}
The DAG-union procedure above resolves conflicts between merges that are \emph{shared} across constituent tokenizers. A distinct failure mode arises from merges that are \emph{not} shared: two tokenizers may segment the same substring in entirely different, non-overlapping ways because their training corpora favored different decompositions, with neither tokenizer's intermediate merges appearing in the other's table at all. 
This does not produce any cycles, there is no rank to resolve, since the competing merges never co-occur, but it can still yield segmentations that violate morphological boundaries once both decompositions coexist in $M^*$ and greedy BPE is applied.
 
For example, tokenizing the English word \texttt{rescheduled} with the French and English constituent tokenizers yields \texttt{res|che|du|le|d} and \texttt{res|ched|ul|ed} respectively: neither contains the other's merge (\tokenbox{motoPurpleBg}{c},\tokenbox{motoBlueBg}{he}) vs.\ (\tokenbox{motoPinkBg}{ch},\tokenbox{motoOrangeBg}{e}) or \tokenbox{motoTealBg}{c},\tokenbox{motoGreenBg}{hed}, so the two decompositions are independent rather than conflicting. 
Critically, the French tokenizer's split reuses two of the most frequent French function-word fragments, \tokenbox{motoYellowBg}{du} (``of the'') and \tokenbox{motoOrangeBg}{le} (``the''), purely because they are high-rank merges in French and happen to match this English substring by coincidence. Similarly, \texttt{unexpected} is split as
\tokenbox{motoTealBg}{une}\tokenbox{motoGreenBg}{x}\tokenbox{motoTealBg}{pec}\tokenbox{motoGreenBg}{ted} 
by the French tokenizer, reusing the French feminine indefinite article \texttt{une} (``a''), versus the more morphologically sensible 
\tokenbox{motoTealBg}{un}\tokenbox{motoGreenBg}{exp}\tokenbox{motoTealBg}{ected}
 from the English tokenizer.
 
Because these cases do not produce cycles, they are invisible to the rank resolution policy and instead emerge from the standard greedy longest-match behavior of BPE once both decompositions are present in $M^*$. We view this as a limitation of the present method: while DAG-union resolution guarantees a well-defined, acyclic merge table, it does not guarantee that the resulting segmentations are linguistically coherent for languages whose high-frequency fragments coincide with substrings of unrelated words in other languages. We leave a systematic audit of how frequently such cross-lingual fragment intrusion occurs, and its downstream effect on tokenization quality, to future work.

\begin{figure}[!htbp]
    \centering
    \includegraphics[width=0.8\linewidth]{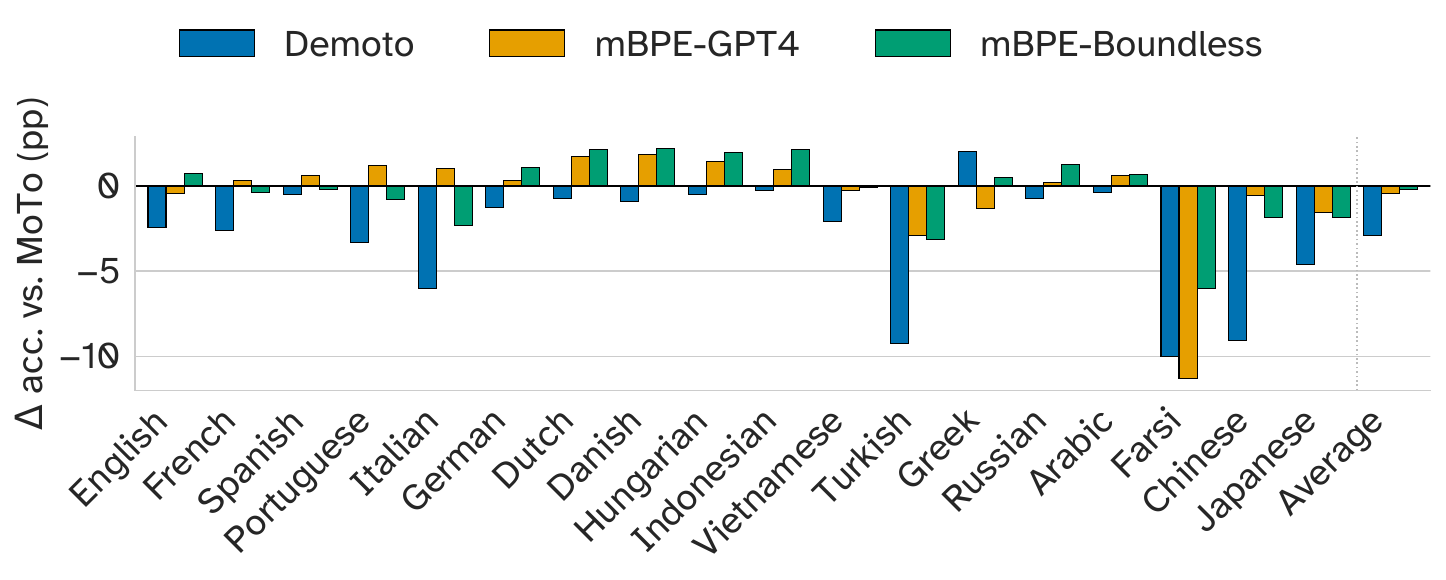}
    \caption{\textbf{\moto\ against single-vocabulary alternatives.} Per-language accuracy difference (\%) from \moto, averaged over each language's tasks. Bars below zero mean the model is worse than \moto. Demoto uses \moto's merged vocabulary as one monolithic tokenizer, so every language is segmented by the same merges but potentially different ordering; mBPE-GPT4 and mBPE-Boundless are multilingual BPE tokenizers trained on the same data mixture, with GPT-4 and Boundless pre-tokenization respectively. The rightmost group is the average over languages.}
    \label{fig:demoto_mbpe_per_lang}
\end{figure}

\paragraph{Constructed monolithic tokenizer (De-\moto{}) against \moto{}}
We see that De-\moto{} performs worse than \moto{} and mBPE alternatives (Figure~\ref{fig:demoto_mbpe_per_lang}).

\clearpage
\section{Tokenizer Quality \& Compression\label{ap:sec:intrinsics}}
We report tokenizer quality metrics for all configurations computed on the FLORES+~\citep{flores1-nllb2022} dataset using \texttt{TokEval}~\citep{meisterTokEvalTokenizerAnalysis2026} library.

\begin{figure}
    \centering
    \includegraphics[width=0.48\linewidth]{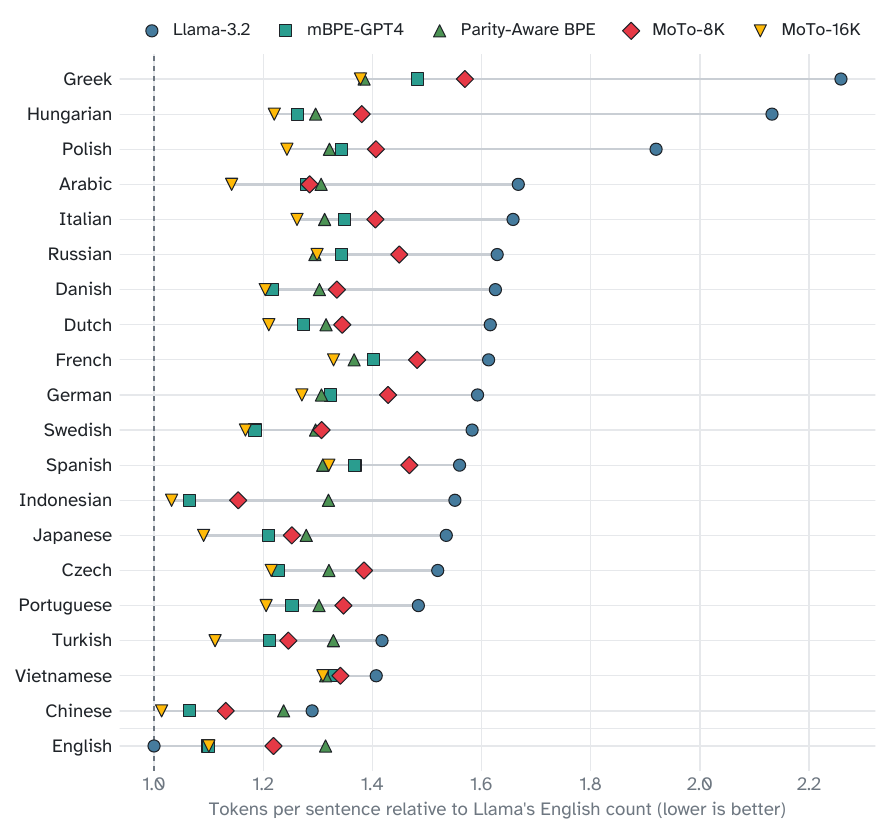}
    \includegraphics[width=0.48\linewidth]{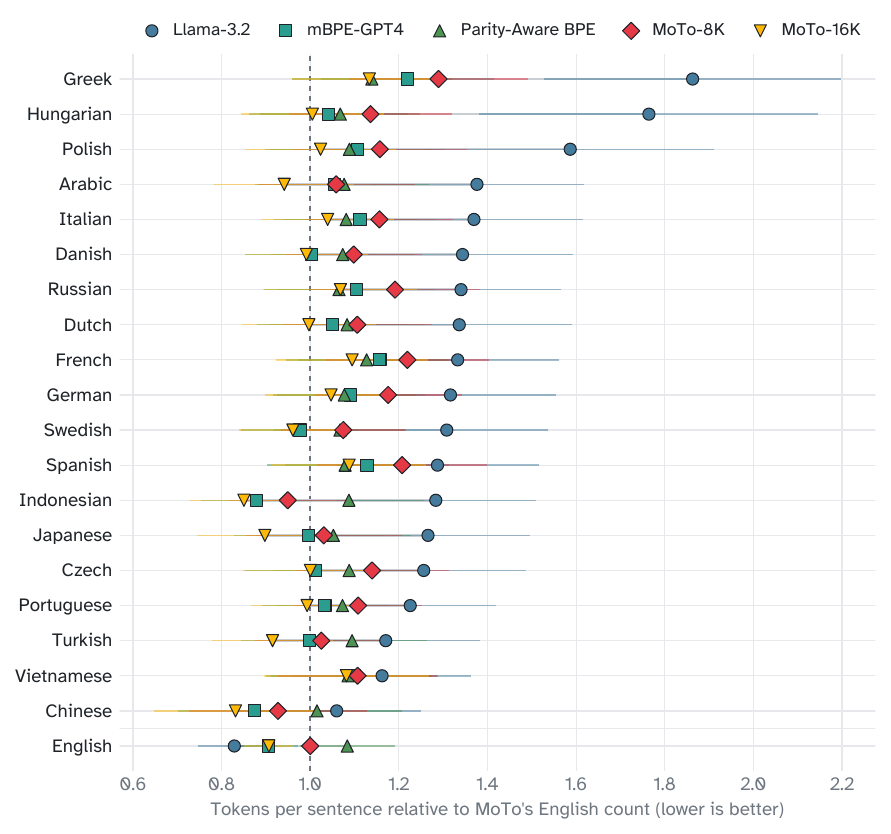}
    \caption{\textbf{Relative parity per language} (lower is more fair). Tokens per FLORES+ sentence divided by the English token count of Llama-3.2 (left) or \moto{} 8K (right), so the dashed line at 1 is the reference tokenizer on English. Each marker is one tokenizer setting, and languages are sorted by Llama's parity with English last. \moto{} (16K) is the cheapest in most. Full numbers are in Tables~\ref{tab:parity-against-llama} and~\ref{tab:parity-against-moto}.}
    \label{ap:fig:relative_parity}
\end{figure}

\paragraph{Relative parity.}
To further examine the relative token cost of each setting, we compare the relative cost of tokenizing the same sentence in that language against English for Llama-3.2 (Figure~\ref{ap:fig:relative_parity}, left and Table~\ref{tab:parity-against-llama}) and MoTo (Figure~\ref{ap:fig:relative_parity}, right and Table~\ref{tab:parity-against-moto}). We tokenize the same 1000 parallel FLORES+ devtest sentences in every language and divide each tokenizer's token count by
Llama-3.2's count for the English version of that sentence (Table~\ref{tab:parity-against-llama}), where 1 means ``as many tokens as Llama needs for the English
sentence''; lower is better parity, where the languages cost similarly across tokenizers. Llama pays the most outside English, from 1.3 on Chinese to 2.3 on Greek (1.60 on average). \moto{} (8K) needs fewer
tokens than Llama on every non-English language and averages 1.35; \moto{} (16K) averages 1.21, the lowest of all tokenizers in the table, but note that this setting has naturally more capacity than others. 
Naturally, \moto{} (8K) needs 22\% more tokens than Llama-3 on English (1.22, against 1.10 for mBPE-GPT4), and Parity-Aware BPE needs 31\% more.

\begin{table*}[t]
  \centering
  \footnotesize
  \setlength{\tabcolsep}{4pt}
  \caption{Parity against Llama-3 tokenizer: tokens per sentence relative to Llama's English count (mean $\pm$ std over sentences; lower is better).}
  \label{tab:parity-against-llama}
  \begin{tabular}{lrrrrrrr}
    \toprule
    \multicolumn{1}{l}{} & \multicolumn{4}{c}{\textit{Baselines}} & \multicolumn{3}{c}{\textit{Per-language}} \\
    \cmidrule(lr){2-5}\cmidrule(lr){6-8}
    \textbf{Language} & \textbf{Llama-3.2} & \textbf{mBPE-GPT4} & \textbf{mBPE-Boundless} & \textbf{Parity-Aware BPE} & \textbf{\moto-GPT4} & \textbf{\moto} & \textbf{\moto\ (16K)} \\
    \midrule
    Arabic & 1.67 $\pm$ 0.29 & 1.28 $\pm$ 0.22 & 1.27 $\pm$ 0.22 & 1.31 $\pm$ 0.24 & 1.29 $\pm$ 0.23 & 1.29 $\pm$ 0.22 & \textbf{1.14 $\pm$ 0.20} \\
    Czech & 1.52 $\pm$ 0.26 & 1.23 $\pm$ 0.19 & 1.23 $\pm$ 0.19 & 1.32 $\pm$ 0.22 & 1.39 $\pm$ 0.23 & 1.38 $\pm$ 0.23 & \textbf{1.21 $\pm$ 0.20} \\
    Chinese & 1.29 $\pm$ 0.26 & 1.06 $\pm$ 0.24 & 1.09 $\pm$ 0.24 & 1.24 $\pm$ 0.27 & 1.53 $\pm$ 0.30 & 1.13 $\pm$ 0.28 & \textbf{1.01 $\pm$ 0.26} \\
    Danish & 1.63 $\pm$ 0.28 & 1.22 $\pm$ 0.18 & 1.22 $\pm$ 0.18 & 1.30 $\pm$ 0.21 & 1.33 $\pm$ 0.21 & 1.33 $\pm$ 0.21 & \textbf{1.20 $\pm$ 0.18} \\
    German & 1.59 $\pm$ 0.27 & 1.32 $\pm$ 0.20 & 1.33 $\pm$ 0.20 & 1.31 $\pm$ 0.20 & 1.42 $\pm$ 0.22 & 1.43 $\pm$ 0.22 & \textbf{1.27 $\pm$ 0.19} \\
    Greek & 2.26 $\pm$ 0.40 & 1.48 $\pm$ 0.26 & 1.48 $\pm$ 0.26 & \textbf{1.38 $\pm$ 0.24} & 1.56 $\pm$ 0.28 & 1.57 $\pm$ 0.28 & \textbf{1.38 $\pm$ 0.24} \\
    French & 1.61 $\pm$ 0.26 & 1.40 $\pm$ 0.22 & 1.39 $\pm$ 0.22 & 1.37 $\pm$ 0.22 & 1.52 $\pm$ 0.25 & 1.48 $\pm$ 0.25 & \textbf{1.33 $\pm$ 0.22} \\
    Hungarian & 2.13 $\pm$ 0.43 & 1.26 $\pm$ 0.22 & 1.27 $\pm$ 0.22 & 1.30 $\pm$ 0.22 & 1.38 $\pm$ 0.24 & 1.38 $\pm$ 0.24 & \textbf{1.22 $\pm$ 0.21} \\
    Indonesian & 1.55 $\pm$ 0.24 & 1.06 $\pm$ 0.15 & 1.07 $\pm$ 0.15 & 1.32 $\pm$ 0.21 & 1.15 $\pm$ 0.18 & 1.15 $\pm$ 0.18 & \textbf{1.03 $\pm$ 0.15} \\
    Italian & 1.66 $\pm$ 0.28 & 1.35 $\pm$ 0.21 & 1.35 $\pm$ 0.21 & 1.31 $\pm$ 0.21 & 1.42 $\pm$ 0.23 & 1.41 $\pm$ 0.23 & \textbf{1.26 $\pm$ 0.19} \\
    Japanese & 1.54 $\pm$ 0.28 & 1.21 $\pm$ 0.22 & 1.23 $\pm$ 0.22 & 1.28 $\pm$ 0.23 & 2.10 $\pm$ 0.40 & 1.25 $\pm$ 0.23 & \textbf{1.09 $\pm$ 0.20} \\
    Dutch & 1.62 $\pm$ 0.28 & 1.27 $\pm$ 0.20 & 1.28 $\pm$ 0.20 & 1.32 $\pm$ 0.22 & 1.34 $\pm$ 0.23 & 1.34 $\pm$ 0.23 & \textbf{1.21 $\pm$ 0.19} \\
    Polish & 1.92 $\pm$ 0.37 & 1.34 $\pm$ 0.24 & 1.34 $\pm$ 0.24 & 1.32 $\pm$ 0.24 & 1.41 $\pm$ 0.26 & 1.41 $\pm$ 0.26 & \textbf{1.24 $\pm$ 0.22} \\
    Portuguese & 1.48 $\pm$ 0.22 & 1.25 $\pm$ 0.17 & 1.26 $\pm$ 0.17 & 1.30 $\pm$ 0.19 & 1.35 $\pm$ 0.20 & 1.35 $\pm$ 0.20 & \textbf{1.21 $\pm$ 0.16} \\
    Russian & 1.63 $\pm$ 0.29 & 1.34 $\pm$ 0.24 & 1.35 $\pm$ 0.24 & \textbf{1.29 $\pm$ 0.23} & 1.48 $\pm$ 0.27 & 1.45 $\pm$ 0.26 & 1.30 $\pm$ 0.23 \\
    Spanish & 1.56 $\pm$ 0.27 & 1.37 $\pm$ 0.23 & 1.37 $\pm$ 0.23 & \textbf{1.31 $\pm$ 0.22} & 1.47 $\pm$ 0.25 & 1.47 $\pm$ 0.25 & 1.32 $\pm$ 0.22 \\
    Swedish & 1.58 $\pm$ 0.26 & 1.19 $\pm$ 0.16 & 1.19 $\pm$ 0.16 & 1.30 $\pm$ 0.19 & 1.31 $\pm$ 0.20 & 1.31 $\pm$ 0.20 & \textbf{1.17 $\pm$ 0.17} \\
    Turkish & 1.42 $\pm$ 0.25 & 1.21 $\pm$ 0.19 & 1.21 $\pm$ 0.19 & 1.33 $\pm$ 0.22 & 1.25 $\pm$ 0.20 & 1.25 $\pm$ 0.20 & \textbf{1.11 $\pm$ 0.18} \\
    Vietnamese & 1.41 $\pm$ 0.23 & 1.33 $\pm$ 0.21 & 1.33 $\pm$ 0.21 & \textbf{1.31 $\pm$ 0.21} & 1.34 $\pm$ 0.21 & 1.34 $\pm$ 0.21 & \textbf{1.31 $\pm$ 0.21} \\
    English & \textbf{1.00 $\pm$ 0.00} & 1.10 $\pm$ 0.07 & 1.10 $\pm$ 0.08 & 1.31 $\pm$ 0.13 & 1.22 $\pm$ 0.13 & 1.22 $\pm$ 0.13 & 1.10 $\pm$ 0.08 \\
    \midrule
    \textbf{Average} & 1.60 & 1.26 & 1.27 & 1.31 & 1.41 & 1.35 & \textbf{1.21} \\
    \bottomrule
  \end{tabular}
\end{table*}

\begin{table*}[t]
  \centering
  \footnotesize
  \setlength{\tabcolsep}{4pt}
  \caption{Parity against MoTo: tokens per sentence relative to MoTo's English count (mean $\pm$ std over sentences; lower is better).}
  \label{tab:parity-against-moto}
  \begin{tabular}{lrrrrrrr}
    \toprule
    \multicolumn{1}{l}{} & \multicolumn{4}{c}{\textit{Baselines}} & \multicolumn{3}{c}{\textit{Per-language}} \\
    \cmidrule(lr){2-5}\cmidrule(lr){6-8}
    \textbf{Language} & \textbf{Llama-3.2} & \textbf{mBPE-GPT4} & \textbf{mBPE-Boundless} & \textbf{Parity-Aware BPE} & \textbf{\moto-GPT4} & \textbf{\moto} & \textbf{\moto\ (16K)} \\
    \midrule
    Arabic & 1.38 $\pm$ 0.24 & 1.05 $\pm$ 0.18 & 1.04 $\pm$ 0.17 & 1.08 $\pm$ 0.19 & 1.06 $\pm$ 0.18 & 1.06 $\pm$ 0.18 & \textbf{0.94 $\pm$ 0.16} \\
    Czech & 1.26 $\pm$ 0.23 & 1.01 $\pm$ 0.16 & 1.01 $\pm$ 0.16 & 1.09 $\pm$ 0.17 & 1.14 $\pm$ 0.17 & 1.14 $\pm$ 0.17 & \textbf{1.00 $\pm$ 0.15} \\
    Chinese & 1.06 $\pm$ 0.19 & 0.87 $\pm$ 0.17 & 0.90 $\pm$ 0.17 & 1.02 $\pm$ 0.19 & 1.26 $\pm$ 0.24 & 0.93 $\pm$ 0.20 & \textbf{0.83 $\pm$ 0.18} \\
    Danish & 1.34 $\pm$ 0.25 & 1.00 $\pm$ 0.15 & 1.00 $\pm$ 0.15 & 1.07 $\pm$ 0.16 & 1.10 $\pm$ 0.15 & 1.10 $\pm$ 0.15 & \textbf{0.99 $\pm$ 0.14} \\
    German & 1.32 $\pm$ 0.24 & 1.09 $\pm$ 0.17 & 1.09 $\pm$ 0.17 & 1.08 $\pm$ 0.16 & 1.17 $\pm$ 0.16 & 1.18 $\pm$ 0.17 & \textbf{1.05 $\pm$ 0.15} \\
    Greek & 1.86 $\pm$ 0.33 & 1.22 $\pm$ 0.20 & 1.22 $\pm$ 0.19 & 1.14 $\pm$ 0.18 & 1.29 $\pm$ 0.20 & 1.29 $\pm$ 0.20 & \textbf{1.13 $\pm$ 0.17} \\
    French & 1.33 $\pm$ 0.23 & 1.16 $\pm$ 0.18 & 1.15 $\pm$ 0.18 & 1.13 $\pm$ 0.18 & 1.25 $\pm$ 0.19 & 1.22 $\pm$ 0.18 & \textbf{1.09 $\pm$ 0.17} \\
    Hungarian & 1.76 $\pm$ 0.38 & 1.04 $\pm$ 0.18 & 1.05 $\pm$ 0.18 & 1.07 $\pm$ 0.18 & 1.13 $\pm$ 0.18 & 1.14 $\pm$ 0.18 & \textbf{1.00 $\pm$ 0.16} \\
    Indonesian & 1.28 $\pm$ 0.23 & 0.88 $\pm$ 0.12 & 0.88 $\pm$ 0.13 & 1.09 $\pm$ 0.17 & 0.95 $\pm$ 0.13 & 0.95 $\pm$ 0.13 & \textbf{0.85 $\pm$ 0.12} \\
    Italian & 1.37 $\pm$ 0.24 & 1.11 $\pm$ 0.17 & 1.11 $\pm$ 0.17 & 1.08 $\pm$ 0.16 & 1.17 $\pm$ 0.17 & 1.16 $\pm$ 0.17 & \textbf{1.04 $\pm$ 0.15} \\
    Japanese & 1.27 $\pm$ 0.23 & 1.00 $\pm$ 0.17 & 1.01 $\pm$ 0.17 & 1.05 $\pm$ 0.18 & 1.74 $\pm$ 0.35 & 1.03 $\pm$ 0.18 & \textbf{0.90 $\pm$ 0.15} \\
    Dutch & 1.34 $\pm$ 0.25 & 1.05 $\pm$ 0.17 & 1.05 $\pm$ 0.17 & 1.08 $\pm$ 0.17 & 1.11 $\pm$ 0.17 & 1.11 $\pm$ 0.17 & \textbf{1.00 $\pm$ 0.15} \\
    Polish & 1.59 $\pm$ 0.32 & 1.11 $\pm$ 0.20 & 1.11 $\pm$ 0.20 & 1.09 $\pm$ 0.19 & 1.16 $\pm$ 0.20 & 1.16 $\pm$ 0.20 & \textbf{1.02 $\pm$ 0.17} \\
    Portuguese & 1.23 $\pm$ 0.19 & 1.03 $\pm$ 0.14 & 1.04 $\pm$ 0.14 & 1.07 $\pm$ 0.15 & 1.11 $\pm$ 0.14 & 1.11 $\pm$ 0.14 & \textbf{0.99 $\pm$ 0.13} \\
    Russian & 1.34 $\pm$ 0.22 & 1.11 $\pm$ 0.18 & 1.11 $\pm$ 0.18 & \textbf{1.07 $\pm$ 0.17} & 1.22 $\pm$ 0.20 & 1.19 $\pm$ 0.19 & \textbf{1.07 $\pm$ 0.17} \\
    Spanish & 1.29 $\pm$ 0.23 & 1.13 $\pm$ 0.18 & 1.13 $\pm$ 0.18 & \textbf{1.08 $\pm$ 0.18} & 1.21 $\pm$ 0.19 & 1.21 $\pm$ 0.19 & 1.09 $\pm$ 0.17 \\
    Swedish & 1.31 $\pm$ 0.23 & 0.98 $\pm$ 0.13 & 0.98 $\pm$ 0.13 & 1.07 $\pm$ 0.15 & 1.07 $\pm$ 0.14 & 1.07 $\pm$ 0.14 & \textbf{0.96 $\pm$ 0.12} \\
    Turkish & 1.17 $\pm$ 0.21 & 1.00 $\pm$ 0.16 & 0.99 $\pm$ 0.16 & 1.09 $\pm$ 0.17 & 1.03 $\pm$ 0.15 & 1.02 $\pm$ 0.15 & \textbf{0.92 $\pm$ 0.14} \\
    Vietnamese & 1.16 $\pm$ 0.20 & 1.10 $\pm$ 0.19 & 1.10 $\pm$ 0.19 & 1.09 $\pm$ 0.19 & 1.11 $\pm$ 0.18 & 1.11 $\pm$ 0.18 & \textbf{1.08 $\pm$ 0.19} \\
    English & \textbf{0.83 $\pm$ 0.08} & 0.91 $\pm$ 0.07 & 0.91 $\pm$ 0.07 & 1.08 $\pm$ 0.11 & 1.00 $\pm$ 0.02 & 1.00 $\pm$ 0.00 & 0.91 $\pm$ 0.06 \\
    \midrule
    \textbf{Average} & 1.32 & 1.04 & 1.04 & 1.08 & 1.16 & 1.11 & \textbf{0.99} \\
    \bottomrule
  \end{tabular}
\end{table*}

\paragraph{Cross-language Gini (Table~\ref{tab:tokeval-gini}).}
While parity measures cost; the Gini coefficient~\citep{meisterTokEvalTokenizerAnalysis2026} measures how evenly it is spread. For each tokenizer we take the cost of a language, either tokens per FLORES+ sentence or tokens per UTF-8 byte, and compute the Gini coefficient across languages:
0 means every language costs the same, and larger values mean the cost is concentrated in a few languages. Per sentence,
\moto{} (8K) scores 0.045, the same as the jointly trained mBPE-GPT4 and mBPE-Boundless (0.047, 0.045) and far below Llama (0.082). Parity-Aware BPE is the most even (0.011), as it is trained to equalize parity. Per byte the ranking is
similar, except that the byte-level baseline ByT5 scores 0 by construction. We report the per-sentence column and the tokens per sentence next to it, since a low Gini does not reveal how many tokens are spent.
\begin{table}[!htbp]
\centering
\footnotesize
\caption{Cross-language Gini coefficient of tokenization cost over 20 languages (excluding fa): cost is tokens per FLORES+ sentence (parallel text) or per UTF-8 byte; 0 means every language costs the same. A byte-level tokenizer is 0 per byte by construction, so read the Gini next to the tokens per sentence.}
\label{tab:tokeval-gini}
\begin{tabular}{lrrrr}
\toprule
 & Gini (sentence) & Gini (byte) & Tokens/sentence & Max/min cost \\
\midrule
Llama-3.2 & 0.082 & 0.102 & 42.6 & 2.23 \\
mBPE-GPT4 & 0.047 & 0.071 & 33.7 & 1.39 \\
mBPE-Boundless & 0.045 & 0.072 & 33.8 & 1.38 \\
mBPE-GPT4 (323K) & 0.050 & 0.073 & 29.8 & 1.40 \\
Parity-Aware BPE & \textbf{0.011} & 0.092 & 34.9 & 1.12 \\
ByT5 & 0.116 & \textbf{0.000} & 163.2 & 2.38 \\
MoTo-GPT4 & 0.062 & 0.103 & 37.7 & 1.80 \\
MoTo & 0.045 & 0.080 & 35.9 & 1.38 \\
MoTo-16K & 0.046 & 0.078 & 32.2 & 1.36 \\
MoTo-16K + Llama & 0.049 & 0.079 & 32.1 & 1.37 \\
\bottomrule
\end{tabular}
\end{table}

\end{document}